\pdfoutput=1
\documentclass{article}

    \usepackage[preprint]{neurips_2023}

\usepackage[utf8]{inputenc} 
\usepackage[T1]{fontenc}    
\usepackage{hyperref}       
\usepackage{url}            
\usepackage{booktabs}       
\usepackage{amsfonts}       
\usepackage{nicefrac}       
\usepackage{microtype}      
\usepackage[dvipsnames]{xcolor}
\usepackage{graphicx}
\usepackage{pifont}
\usepackage{amssymb}
\newcommand{\xmark}{\ding{55}}%
\usepackage{multirow}
\usepackage{placeins}
\usepackage{subcaption}

\newcommand\blue[1]{\textcolor{blue}{#1}}
\newcommand\red[1]{\textcolor{red}{#1}}
\newcommand\green[1]{\textcolor{ForestGreen}{#1}}

\title{Enhancing Visual Domain Adaptation with Source Preparation}

\author{%
  Anirudha Ramesh 
  \quad
  Anurag Ghosh 
  \quad
Christoph Mertz 
\quad
Jeff Schneider
\vspace{5pt}\\
Carnegie Mellon University\\
\texttt{\{aramesh3, anuraggh, cmertz, jeff4\}}\texttt{@andrew.cmu.edu} \\ 
}

\begin{document}

\maketitle

\begin{abstract}
Robotic Perception in diverse domains such as low-light scenarios, where new modalities like thermal imaging and specialized night-vision sensors are increasingly employed, remains a challenge. Largely, this is due to the limited availability of labeled data. Existing Domain Adaptation (\texttt{DA}) techniques, while promising to leverage labels from existing well-lit RGB images, fail to consider the characteristics of the source domain itself. We holistically account for this factor by proposing Source Preparation (\texttt{SP}), a method to mitigate source domain biases.

Our Almost Unsupervised Domain Adaptation (\texttt{AUDA}) framework, a label-efficient semi-supervised approach for robotic scenarios -- employs Source Preparation (\texttt{SP}), Unsupervised Domain Adaptation (\texttt{UDA}) and Supervised Alignment (\texttt{SA}) from limited labeled data. We introduce CityIntensified, a novel dataset comprising temporally aligned image pairs captured from a high-sensitivity camera and an intensifier camera for semantic segmentation and object detection in low-light settings. We demonstrate the effectiveness of our method in semantic segmentation, with experiments showing that \texttt{SP} enhances \texttt{UDA} across a range of visual domains, with improvements up to \green{40.64\%} in mIoU over baseline, while making target models more robust to real-world shifts within the target domain. We show that \texttt{AUDA} is a label-efficient framework for effective \texttt{DA}, significantly improving target domain performance with only tens of labeled samples from the target domain.

\end{abstract}

\section{Introduction}

Visual perception in diverse environments and domains such as low-light is challenging. Animals are adept at perception in such situations, due to structural adaptations in their perception mechanism~\cite{blake1979visual} or novel sensing mechanisms that lets them sense radiant heat beyond the visible spectrum~\cite{newman1981integration}. 
Can we bestow such capabilities to our robots by employing emerging sensing and imaging modalities like thermal and specialized night-vision sensors?

\begin{figure}[t]
    \centering
    \includegraphics[scale=0.148]{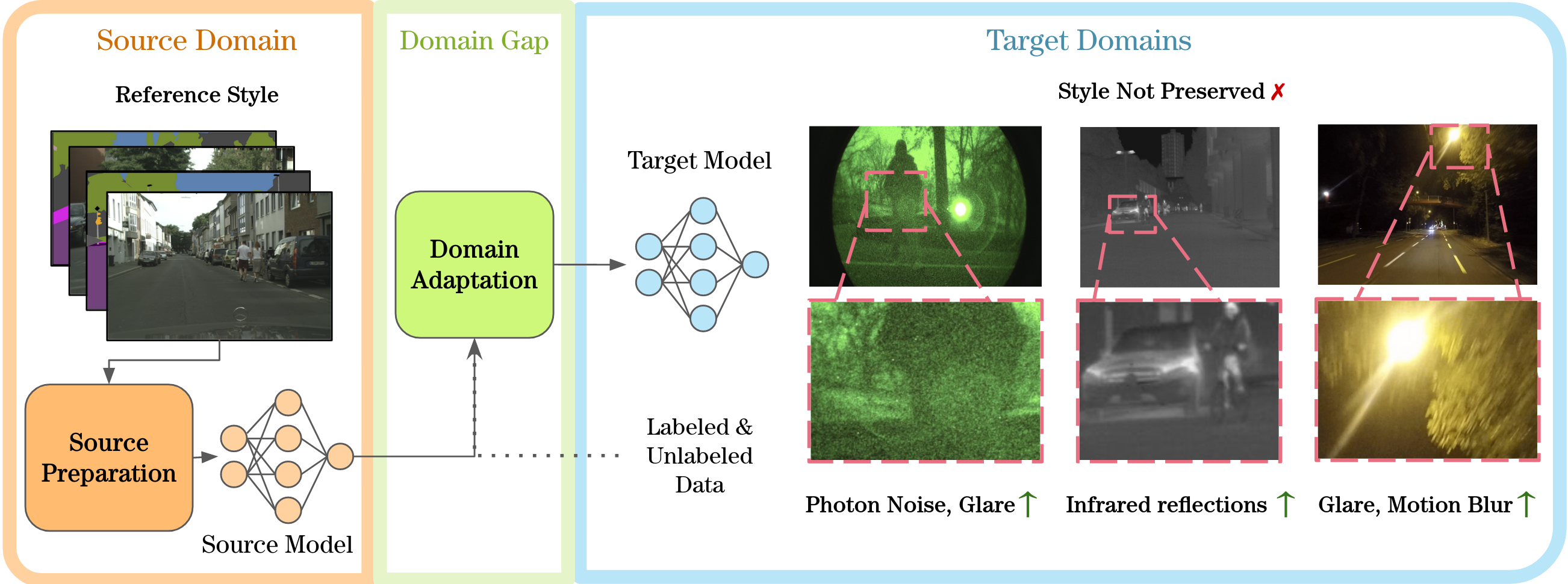}
    \caption{Target domains exhibit characteristics distinct from the source domain, such as high photon noise in intensifier images and infrared reflections in thermal camera images. Similarly, source domain specific characteristics exist, and a source model overfitting to such characteristics can hinder Domain Adaptation. To mitigate this, we propose Source Preparation as an alternative to conventional source model training. Source Preparation enhances domain adaptation by minimizing overfitting in the source domain while implicitly encouraging the learning of features relevant to the target domain.}
    \label{fig:motivation}
    \vspace{-10pt}
\end{figure}

Challenges in robotics in low-light scenarios (such as Figure~\ref{fig:motivation}) can be addressed by employing such sensors and adapting models to operate on these new modalities. However, labeled data in these new domains is limited and developing robotic systems with multimodal capabilities is difficult. Domain adaptation~\cite{Ben-David2010} promises the best of both worlds -- allowing us to leverage similarities across domains without having access to many hard-to-obtain labels while also relying on existing labelled data available in mainstrean visual domains (such as RGB images taken in daytime). In many of these scenarios (such as off-road autonomous driving, and zone exploration at night) it is realistic to assume availability of limited labeled data in addition to unlabelled target data from the target domain. 

Such scenarios aren't captured appropriately by existing approaches to Domain Adaptation in label scarce scenarios, such as Unsupervised Domain Adaptation (\texttt{UDA}), ~\cite{ganin2015unsupervised, long2016unsupervised, saito2018maximum}, Semi-Supervised Domain Adaptation (\texttt{SSDA})~\cite{chen2021semi,  Chen2022SSDA-3, singh2021cldaSSDA, SSDAyoon2021semisupervised}, and  Few-Shot Supervised Domain Adaptation methods (\texttt{FSSDA})\cite{Motiian2017FewShotAdversarialDomainAdaptation, Zhao2021DAFSL, Tavera_2022_WACV_PixDA, FSSDA}. \texttt{UDA} methods attempt to adapt models without utilizing any labeled target domain data, while \texttt{SSDA} methods require hundreds of labeled target samples for complex tasks like semantic segmentation, and existing approaches to \texttt{FSSDA} are generally designed to adapt across small domain gaps.

Moreover, while recent domain adaptation techniques~\cite{bruggemann2023refign, wu2021dannet, DarkZurich, guan2023iterativeActiveDA, hoyer2023mic} adapt models trained on labeled data in source domain to perform well in a different target domain, they fail to consider the characteristics of the source domain itself, that the source model becomes biased towards. Based on this observation, we form the hypothesis, \textit{can we assume that all the features learned by the model trained on the source domain be adapted to other domains?}

To address these issues, we take a holistic view of Domain Adaptation and propose a label-efficient three-stage Semi-Supervised framework called Almost Unsupervised Domain Adaptation (\texttt{AUDA}). Firstly, we propose Source Preparation (\texttt{SP}) as an alternative to conventional source model training, to improve the adaptability of source models (Figure \ref{fig:motivation}). With \texttt{SP}, we test our hypothesis and attempt to mitigate biases towards source domain-specific characteristics by minimizing overfitting in the source
domain while implicitly encouraging the learning of features relevant to the target domain. Then, we employ Unsupervised Domain Adaptation (\texttt{UDA}) to exploit available unlabelled target domain images. Lastly, we exploit the few labeled target images ($\approx$20-50) available to us to perform a limited Supervised Alignment (\texttt{SA}) to the target domain. 

As \texttt{AUDA} employs a far lower number of labeled samples and operates in a different label regime compared to existing \texttt{SSDA} approaches for semantic segmentation~\cite{chen2021semi, Chen2022SSDA-3}, it can be applied to label-scarce domains, while still being able to adapt across larger domain gaps than \texttt{FSSDA} approaches~\cite{Tavera_2022_WACV_PixDA}, by exploiting unlabeled target data more effectively with \texttt{SP} and \texttt{UDA}. 


To rigorously evaluate \texttt{AUDA} and understand the implications of \texttt{SP}, we introduce CityIntensified\footnote{Name changed to preserve anonymity.}, a first-of-its-kind dataset comprising temporally aligned image pairs captured from a high-sensitivity camera and an intensifier camera, with semantic and instance labels, in various low-light scenarios (Section \ref{IntensifierDataset}). While thermal sensors can be used even when it's completely dark, low-light scenarios often have some light to be exploited which regular RGB cameras cannot sufficiently do. We address this gap in existing public datasets for low-light vision tasks and provide paired High-Sensitivity RGB and Intensifier images to enable \texttt{DA} to images captured by an intensifier camera.

Our results show that \texttt{AUDA} and critically, \texttt{SP} improves model performance in various target domains (See Section \ref{Effect of Source Preparation for Domain Adaptation}), while also enhancing robustness to realistic shifts within the target domain (Section \ref{Effects of Source Preparation on Robustness}). 
Our experiments also confirm the efficacy of \texttt{AUDA} for label-efficient \texttt{DA} across challenging domains, with access to as few as 20-50 labeled target samples (Section \ref{Improving Supervised Alignment with Source Preparation}, \ref{Label Efficient Learning}). 
We also provide design principles for selecting or developing \texttt{SP} methods for new target domains. 


\section{Related Work}

\subsection{Domain Adaptation with Limited Supervision}
\label{related:ssda-fssda}

Semi-Supervised Domain Adaptation (\texttt{SSDA}) \cite{chen2021semi,  Chen2022SSDA-3, singh2021cldaSSDA, SSDAyoon2021semisupervised} and Few-Shot Supervised Domain Adaptation (\texttt{FSSDA}) \cite{Motiian2017FewShotAdversarialDomainAdaptation, Zhao2021DAFSL, Tavera_2022_WACV_PixDA, FSSDA} are two lines of work that assume limited availability of labeled samples from the target domain, similar to \texttt{AUDA}. While most \texttt{SSDA} algorithms are proposed for image classification, few proposed for segmentation operate in different label regime, requiring hundreds of labeled target domain instances~\cite{chen2021semi, Chen2022SSDA-3}, compared to tens used by \texttt{AUDA}. On the other hand, \texttt{FSSDA} techniques aim to adapt using \textit{only} a limited number (1-5) of labeled samples from the target domain, and generally do not leverage unlabeled target domain data. This makes it difficult to adapt across large domain gaps, with these methods usually focusing on adaptation across smaller gaps like adapting across cities in CityScapes~\cite{Tavera_2022_WACV_PixDA}(See Section~\ref{Label Efficient Learning}).

In contrast, \texttt{AUDA} leverages all unlabeled data alongside limited labeled target data, thereby combining the strengths of both \texttt{SSDA} and \texttt{FSSDA}, enabling label-efficient adaptation across large domain gaps.

\subsection{Unsupervised Domain Adaptation and Domain Generalization}
\label{related:uda-dg}

In Unsupervised Domain Adaptation (\texttt{UDA})\cite{wilson2020surveyUDA}, data from a labeled source domain and an unlabeled target domain are available. These algorithms employ labeled source data for task supervision, and target data to assist alignment~\cite{wu2021dannet,bruggemann2023refign,hoyer2023mic, DarkZurich}. Generally, they employ an adversarial framework~\cite{hoffman2016fcns,hoffman2017cycada, tsai2018learning, vu2019advent} based on~\cite{ganin2016domainadversarialDANN} and/or propose self-training~\cite{li2019bidirectional, kim2020learning, Yang2018, wang2020classes} approaches which generate and use pseudo-labels~\cite{DongHyun2013} for the target domain. These works focus on improving \texttt{UDA} given source data and a model trained on it. We take a holistic view of the problem, and enhance Domain Adaptation by focusing on creating more adaptable models through Source Preparation. Our proposal is agnostic to specific algorithms and improves these \texttt{UDA} methods.

Another class of methods, Domain Generalization \cite{DomainGeneralizationBlanchard} assume that target domain is unknown, and aim to perform well under arbitrary domain shifts. However, in most robotic scenarios, target domain is known, and utilizing this as a signal can help maximize performance, especially with large domain gaps. Thus we do not explore this class of methods. Reducing source domain-specific overfitting has inspired some recent works in domain generalization~\cite{jin2022SNR, yuan2022domainspecificBiasFiltering}. However, these methods do not connect this idea with preparing more suitable source models for domain adaptation. 

\subsection{Robot Vision in Low Light}
\label{related:low-light}

Vision in low light can be tackled using active or passive sensors and every such sensor represents a new target domain. Active sensors (like LiDAR) are often not applicable due to cost and operating constraints, necessitating the use of passive sensors. Unfortunately, regular cameras are not sensitive enough. While many datasets contain night-time images captured with standard cameras~\cite{DarkZurich, nighttimedriving, bdd100k} in structured environments (roads with street lights). However, they are unable to capture darker environments important for many robotic tasks such as off-road driving. Our dataset, CityIntensified, addresses this gap, and to the best of our knowledge, is the first to capture images from high-sensitivity and intensifier cameras in structured and off-road scenarios.

While most aforementioned methods in Section~\ref{related:ssda-fssda} and ~\ref{related:uda-dg} focus on adaptation from synthetic-to-real images, or across different conditions with a regular RGB camera, we demonstrate our performance on a wider, more challenging variety of domains across changes in time and lighting~\cite{DarkZurich}, and modalities like thermal~\cite{MFNet-thermal} and Intensifier Cameras via CityIntensified dataset.

\section{Methodology}\label{Methodology}

\begin{figure}[t]
    \centering
    \begin{minipage}[t]{0.9\textwidth}
    \raggedright
    \emph{Existing Solutions to \texttt{UDA}} \quad \quad \quad \quad \quad \quad \quad \quad
    \emph{Proposed \texttt{AUDA} Framework}
    \end{minipage}
    \includegraphics[scale=0.12]{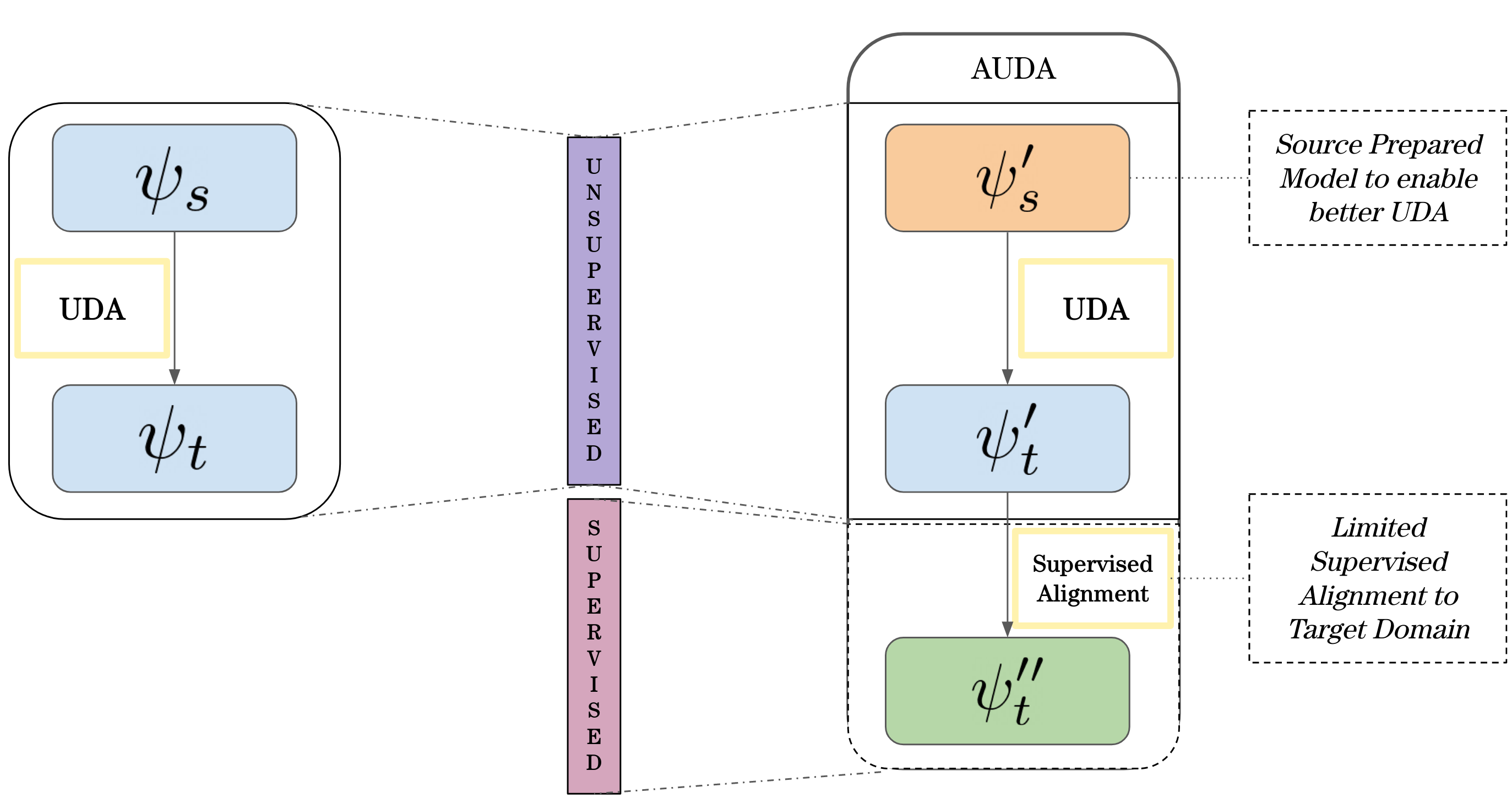}
    \caption{This figure illustrates our proposed  framework, \texttt{AUDA}, for realistic robotic scenarios where some labeled target samples can be obtained. In contrast to traditional \texttt{UDA}, our approach includes Source Preparation (\texttt{SP}) to create a more `adaptable' model for \texttt{UDA}, and Supervised Alignment (\texttt{SA}) to leverage the limited labeled data available in the target domain.}
    \label{fig:method}
\end{figure}

\subsection{Problem Setup}

Domain adaptation involves a source domain \emph{S} abundant in labels, and a target domain \emph{T} with limited to no labels. Given this setup, our goal is to create a model that performs well on \emph{T}. To show the efficacy of our proposals, we focus on improving semantic segmentation for realistic robotic scenarios with existing \texttt{UDA} approaches, though our work can be extended to other tasks, and forms of \texttt{DA}. \\
Let $D_s = {\{x_s^{(i)}}\}_{i=1}^{N_s}$ be the set of images from the source domain, where $x_s^{(i)} \in \mathbb{R}^{H_s \times W_s \times 3}$. 
Let  $L_s = \{y_s^{(i)}\}_{i=1}^{N_s}$ be the set of corresponding one-hot labels for the source domain images, where $y_s^{(i)} \in \{0,1\}^{H_s \times W_s \times C}$, and C is the number of classes. $D_t$ is defined similarly for the target domain. \\
Let $f$ signify the process of training a segmentation model, $\psi_s$, on \emph{S}, where $f$ includes both input data processing and network architecture. Let $g$ be the method performing \texttt{UDA} that aims to adapt $\psi_s$ to the \emph{T} to obtain $\psi_t$. Traditionally,  $\psi_s$ is trained on the labeled source domain data $(D_s, L_s)$ with $f$, while $\psi_t$ is obtained by applying the method performing \texttt{UDA}, $g$, to $\psi_s$ using the source domain data $(D_s, L_s)$ and the unlabeled target domain data $D_t$. For simplicity and ease of understanding, we represent these steps from here on out as $\psi_s = f(D_s, L_s)$, $\psi_t = g(\psi_s, D_s, L_s, D_t)$.

\subsection{Overview of Proposed \texttt{AUDA} Framework}

Our proposed framework for label-efficient \texttt{DA} to \emph{T} given \emph{S} can be separated into 3-stages as follows:-
\begin{itemize}
    \item \textbf{Source Model Preparation} for Domain Adaptation using only $D_s$ and $L_s$.
    \item \textbf{Unsupervised Domain Adaptation} from \emph{S} to \emph{T}, using $D_s$, $L_s$, and $D_t$.
    \item \textbf{Supervised Alignment} with limited labeled data in \emph{T} to improve final performance in \emph{T}. 
\end{itemize}

Concretely, our Source Preparation step introduces $f'$ in place of $f$ in the original problem setup. The newly formulated setup now looks like $\psi_s' = f'(D_s, L_s)$,   $\psi_t' = g(\psi_s', D_s, L_s, D_t)$. In Section \ref{Source Preparation Method} we detail how we design $f'$, but it's key to note that we do not propose adding any additional parameters or significantly changing the network architecture. Our final step, Supervised Alignment, performs the following update to obtain our final target model $\psi_t'' = h(\psi_t', D'_t, L'_t)$, assuming we have a labeled target set $\{D'_t, L'_t\}$ where $\vert D'_t \vert \ll \vert D_t \vert$ and $L'_t$ is the set of labels corresponding to $D'_t$, defined similar to $L_s$.
Our framework is illustrated in Figure \ref{fig:method}, wherein we highlight the modifications made to build \texttt{AUDA} atop existing \texttt{UDA} approaches.  

\subsection{Source Preparation}\label{Source Preparation Method}

Our Source Preparation (\texttt{SP}) step aims to create source models with features more suitable for domain adaptation. We do so by trying to reduce biases in the source model towards source domain-specific characteristics by addressing overfitting in the source domain. We propose and evaluate the efficacy of 3 schemes across different approaches : (1) explicitly targeting style-based biases in subsection \ref{MixStyle}, (2) regularization in subsection \ref{mixup}, and (3) high-frequency detail reduction in subsection \ref{Blur}. These schemes however do not form an exhaustive set of all possible approaches to \texttt{SP}, rather just aim to demonstrate the promise of \texttt{SP}. We explain the motivations behind specific \texttt{SP} schemes detailed in subsections \ref{MixStyle}-\ref{Blur} below. These are elaborated further in the supplemental.

Prior works like \cite{mixstyle_zhou2021domain} suggest that visual domain is closely related to image style. We hypothesize that a source model not overfit to style should be easier to transfer to new domains with varying styles. With this, we develop a scheme detailed in \ref{MixStyle}. 

We hypothesize that increased regularization during the source model's training can help us learn features more robust to domain-specific noise. 
We propose a \texttt{SP} scheme from this intuition which is further detailed in \ref{mixup}.

In domains such as low-light environments, we attribute high-freq noise to be a key component of domain-specific noise. Low-light photon noise and glare are examples of domain-specific noise with high-frequency components. While rough shapes are usually preserved across domains such as regular day-night images, and thermal images, the details often vary. In \ref{Blur}, we target this directly.

We use \texttt{SegFormer} \cite{xie2021segformer} (\texttt{MiT-B5}) as our segmentation model, and explain any modifications made to this network during \texttt{SP} below. 
Note that, we don't add any learnable parameters in any of these modifications, and the unmodified original network architecture is used in subsequent steps. While some of the methods we use in \texttt{SP} have been proposed in other settings, we contextualize them in the \texttt{AUDA} paradigm and intend to exploit their properties to aid the creation of more `adaptable' source models for DA.


\begin{figure}
    \centering
    \includegraphics[scale=0.15]{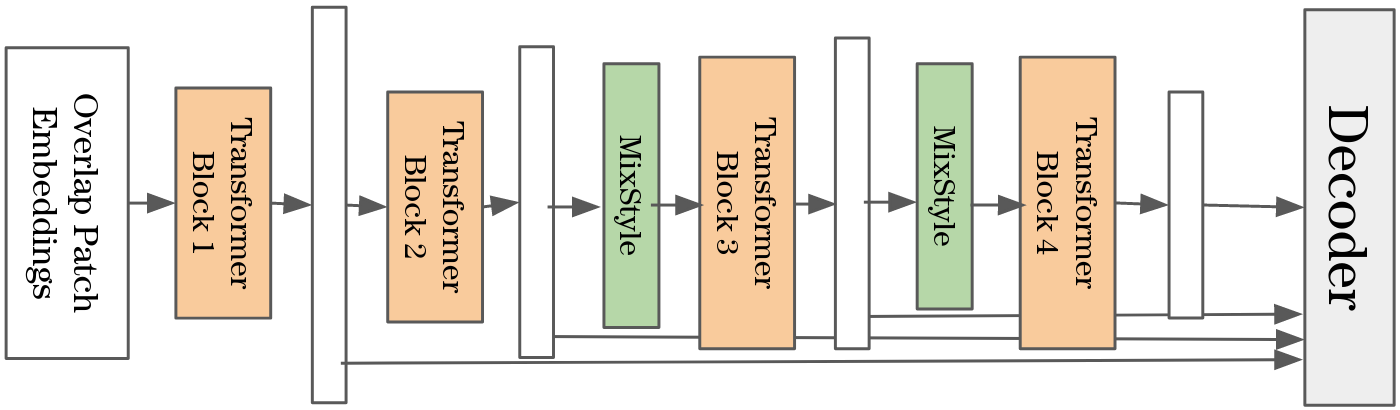}
    \caption{\texttt{MixStyle} is used for Source Preparation (\texttt{SP}) by making the \green{highlighted} modification in \texttt{SegFormer's} encoder \cite{xie2021segformer}. These modifications are used only for \texttt{SP}, and not \texttt{UDA} or \texttt{SA}.}
    \label{fig:SP-encoder-modification}
\end{figure}

\subsubsection{MixStyle} \label{MixStyle}
Prior works show that instance-level feature statistics like mean and variance capture style in neural networks \cite{huang2017arbitrary_adain, Gatys2016StyleTransfer, li2017demystifying}, including transformers for vision \cite{kim2022instaformer}. \texttt{MixStyle} \cite{mixstyle_zhou2021domain} is a \texttt{DG} approach based on probabilistically mixing these statistics of training samples from source domain(s), to learn features more robust to variations in image-style. We include \texttt{MixStyle} after block-2 and block-3 in \texttt{SegFormer's} encoder (\texttt{MiT-B5}) to train our source prepared model, $\psi_s'$, as illustrated in Figure \ref{fig:SP-encoder-modification}.  


\subsubsection{Mixup}  \label{mixup}

\texttt{mixup} \cite{zhang2018mixup} is a data-augmentation technique that regularizes neural networks to favor simple linear behavior in-between training examples by training it on convex combinations of pairs of samples and their labels. We choose our mixup parameters based on \cite{pinto2023regmixup}.

\subsubsection{Blur} \label{Blur} 
GaussianBlur and other kernel-based blurring methods are commonly employed in data augmentation to enhance the robustness of neural networks against variations in high-frequency details and noise \cite{Shorten2019SurveyOnDataAugmentation}. We propose using a strong blurring scheme during source model training where we blur out images with a 50\% chance, using a Gaussian kernel of size uniformly sampled from (5, 5) to (19, 19).


\subsection{Unsupervised Domain Adaptation} \label{UnsupervisedDomainAdaptation}

Unsupervised Domain Adaptation (\texttt{UDA}) aims to align $\psi_s'$ to \emph{T} to create $\psi_t'$, usually with task supervision from $\{D_s, L_s\}$ and supervision for alignment from $D_t$. 
While we demonstrate our results with \texttt{Refign} \cite{bruggemann2023refign}, further detailed in Section \ref{Experiments}, our framework is independent of specific \texttt{UDA} methods. 
During \texttt{UDA}, our model receives some supervisory signal from both \emph{S} and \emph{T} leaving it less likely to be biased towards characteristics specific to the source domain as compared to the prior step, where only supervisory signal from \emph{S} is available. This means that we don't require the application of \texttt{SP} techniques as much during UDA. 
Our framework allows us to stop at this step if we don't have any labels in \emph{T}, remaining fully unsupervised, while still obtaining the benefits of \texttt{SP}.
\subsection{Supervised Alignment}
Supervised Alignment (\texttt{SA}) aims to account for realistic robotic scenarios where a small amount (20-50 samples) of labeled data in \emph{T} can be obtained. While we align the model we obtain after \texttt{UDA} with supervision using finetuning, other methods, such as linear probing, can be used here. 
\texttt{SA}, as a part of \texttt{AUDA}, is more label-efficient than \texttt{SSDA} approaches \cite{chen2021semi, Chen2022SSDA-3}, which typically use 100s of labeled images for from the target domain for tasks like semantic segmentation, while still leveraging all unlabeled data in the first two steps, unlike \texttt{FSSDA} approaches \cite{FSSDA, Tavera_2022_WACV_PixDA}, to perform better in the target domain (Section \ref{Label Efficient Learning}).

\section{CityIntensified Dataset} \label{IntensifierDataset}

\begin{figure}[t]
    \centering
    \includegraphics[scale=0.09]{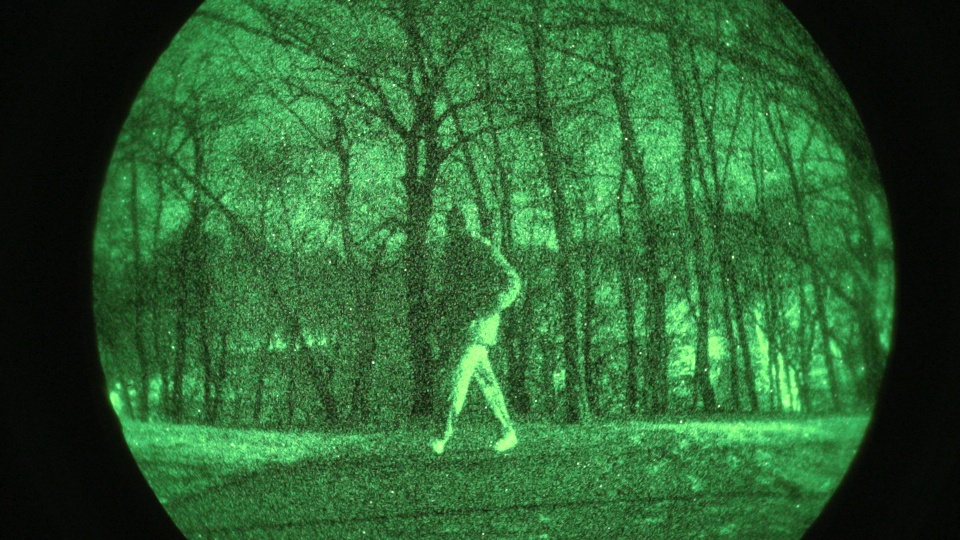}
    \includegraphics[scale=0.09]{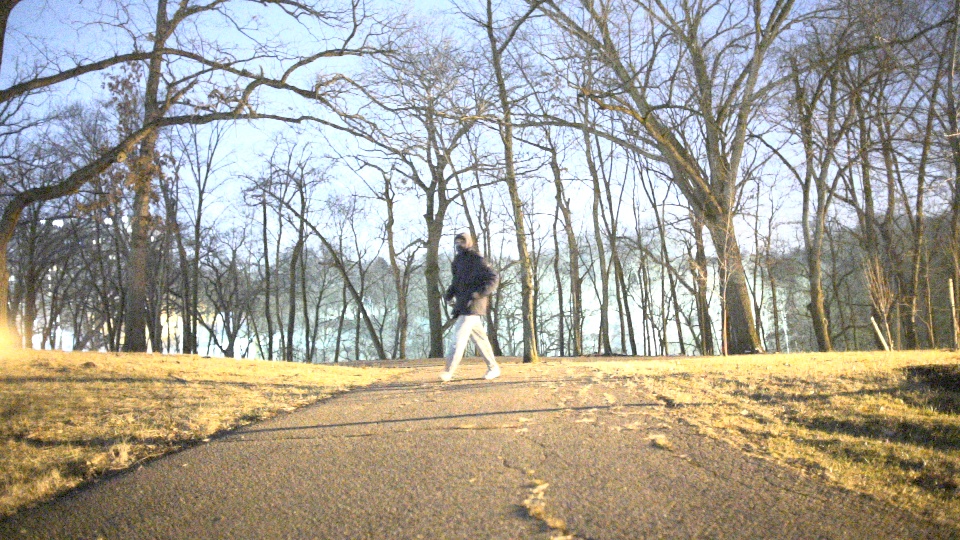}
    \includegraphics[scale=0.09]{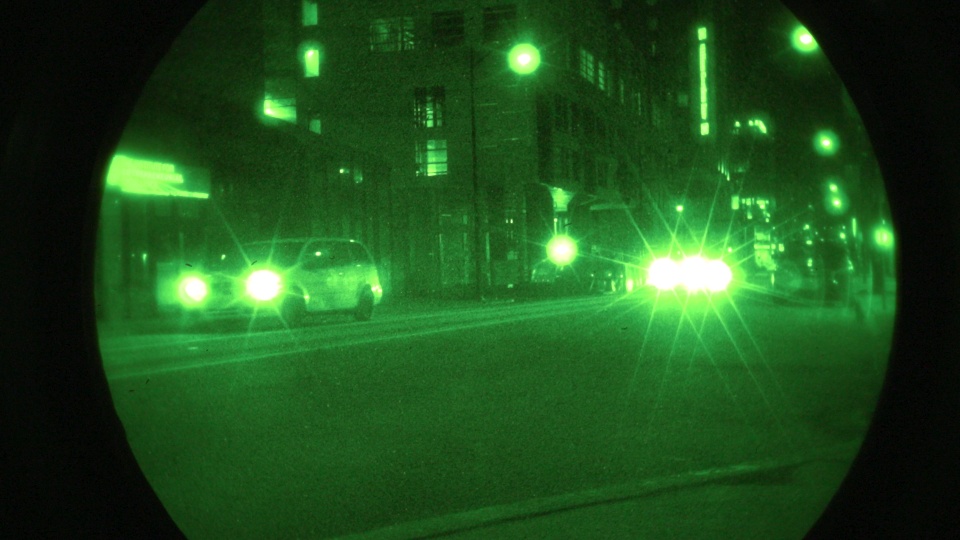}
    \includegraphics[scale=0.09]{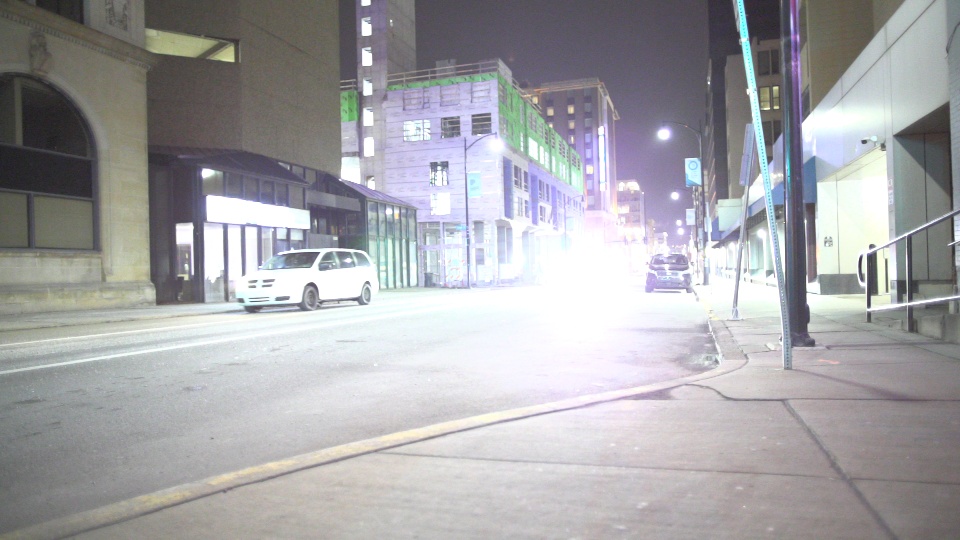}
    \includegraphics[scale=0.09]{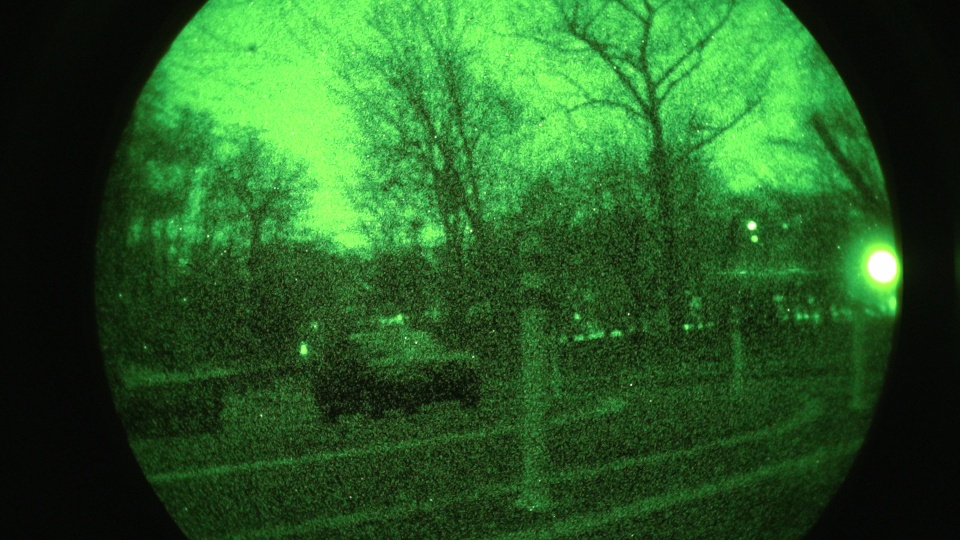}
    \includegraphics[scale=0.09]{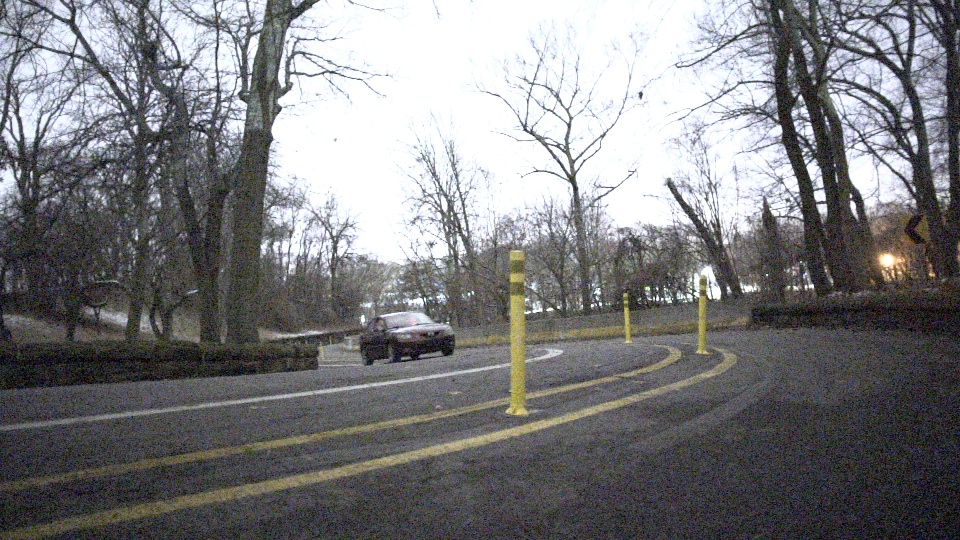}
    \includegraphics[scale=0.09]{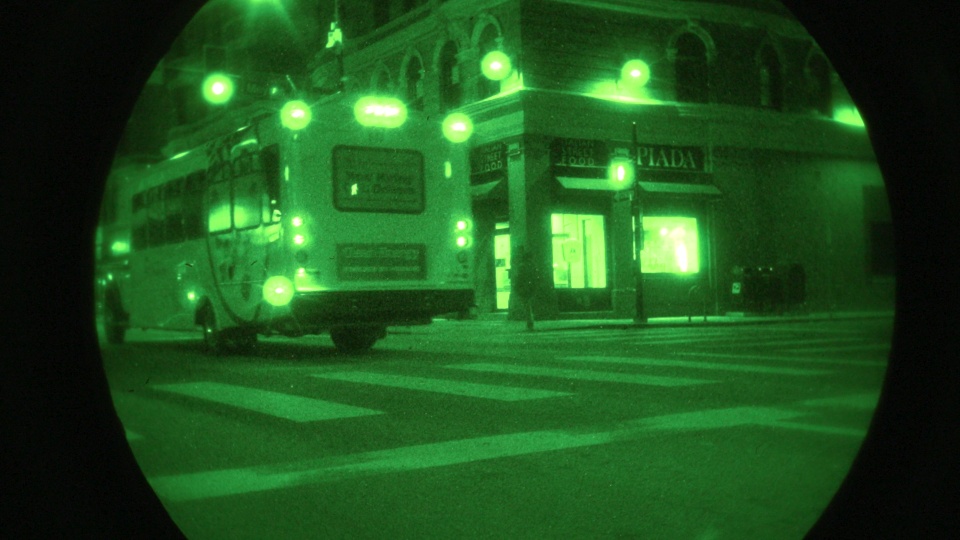}
    \includegraphics[scale=0.09]{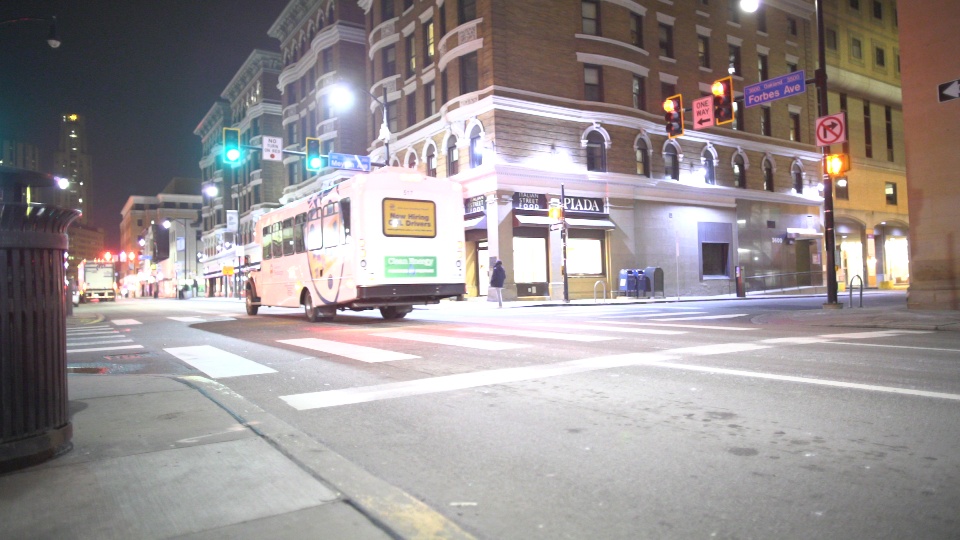}
    \caption{Representative examples of paired images from CityIntensified. Note that, despite the appearance in the high sensitivity camera, these images have been taken at night, in the dark in both structured and unstructured environments.}\label{fig:CityIntensifiedSamples}
\end{figure}

Introducing CityIntensified, a new dataset designed for low-light robotic scenarios, where the utilization of light-sensitive sensors allows for maximally exploiting available light. To the best of our knowledge, no such dataset exists in the public domain. By employing an intensifier as a sensor for low-light vision, CityIntensified bridges the gap between regular RGB cameras, which lack the required sensitivity for such scenarios, and thermal cameras, which operate in a different wavelength range. The dataset comprises 4792 image pairs captured at night, featuring a high-sensitivity RGB camera (\href{https://www.usa.canon.com/shop/p/me20f-sh}{Canon ME20F-SH}) and an intensifier camera (Canon ME20F-SH with \href{https://www.lynred-usa.com/products/night-vision2/astroscope-canon-eos.html}{AstroScope 9350-EOS-PRO Gen 3}), in diverse low-light settings encompassing public streets and parks.

We provide semantic and instance-level labels, obtained by manually correcting labels generated with SegmentAnything~\cite{kirillov2023segany}, for people and vehicles in 393 intensifier images. This is split into a validation set consisting of 293 images, and a train set for limited Supervised Alignment with 100 images. With our paired dataset, we hope to facilitate building a bridge between RGB images, which are captured by the high-sensitivity camera, to images from the intensifier. We provide illustrative samples in Figure \ref{fig:CityIntensifiedSamples} and additional details in our supplementary materials. 

\section{Experiments and Results}\label{Experiments}

\textbf{Task.} We situate this work on the illustrative task of semantic segmentation. However, our approach can however be used for other tasks like panoptic segmentation.

\textbf{Dataset.} We test our proposals across different target domains illustrated in Figure \ref{fig:motivation}. With Cityscapes \cite{cordts2016cityscapes} (CS) as our source domain, we adapt to target domains across time and lighting to DarkZurich \cite{DarkZurich} \emph{(CS$\rightarrow$DZ)}, across modalities to MFNetThermal \cite{MFNet-thermal} \emph{(CS$\rightarrow$MFNT)} and CityIntensified (Section \ref{IntensifierDataset}) \emph{(CS$\rightarrow$CI)}. We evaluate our solutions on labels common to both source and target domains. While \emph{CI} and \emph{MFNT} have train sets for \texttt{SA}, \emph{DZ} does not.

\textbf{Implementation Details.} We choose \texttt{Refign-HRDA*} \cite{bruggemann2023refign} and \texttt{SegFormer} (\texttt{MiT-B5}) \cite{xie2021segformer} as our \texttt{UDA} method, and segmentation network respectively. We train \texttt{Refign} with a scheme similar to the original paper, in exception to increasing iterations $1.5\times$, and \texttt{SegFormer} as the original paper does. In \texttt{SA}, we finetune Segformer for 4000 iterations, scaling down warm-up iterations of the `poly' scheduler to 150. Our approach is independent of specific \texttt{UDA} methods and should extend to others. 

\subsection{Effect of Source Preparation} \label{Effect of Source Preparation for Domain Adaptation}

\begin{table}
  \caption{Comparison (mIoU) on respective validation sets after \texttt{UDA} from Cityscapes to DarkZurich, MFNet Thermal, CityIntensified with different Source Preparation techniques. In each case we can improve the potency of \texttt{UDA} with the right kind of Source Preparation.}
  \label{DA-Refign-cs-target}
  \centering
  \begin{tabular}{lllll}
    \toprule
    \texttt{DA}-Method     & Source Preparation Method     & {CS$\rightarrow$DZ} & {CS$\rightarrow$MFNT} & {CS$\rightarrow$CI} \\
    \midrule
    None & None & $29.30$ & $55.30$ &  $4.57$ \\ 
    \texttt{Refign} & None  & $48.97$  & \underline{$63.45$} & $32.50$   \\
    \texttt{Refign} & \texttt{MixStyle} & $\textbf{49.50}$ & $\textbf{65.00}$ (\green{+1.55}) &  $50.83$\\
    \texttt{Refign} & \texttt{mixup} & $47.39$ & $62.65$ & $\underline{71.87}$ (\green{+39.37})\\
    \texttt{Refign} & Blur & \underline{$49.41$} & $60.40$ & $\textbf{73.14}$ (\green{+40.64}) \\
    \bottomrule
  \end{tabular}
\end{table}

In this section, we compare different \texttt{SP} methods and their impact on target domain performance after \texttt{UDA}. We also show that \texttt{SP} can make the models we obtain after \texttt{UDA} more robust in \ref{Effects of Source Preparation on Robustness}, and that it can improve the models we obtain after Supervised Alignment in \ref{Improving Supervised Alignment with Source Preparation}.

In Table \ref{DA-Refign-cs-target}, we show the performance of models obtained after \texttt{UDA} and different \texttt{SP} schemes we had proposed. We analyze and explain our results based on target domain characteristics below. 

\textbf{MixStyle.}  Regularizing over styles with \texttt{MixStyle} improves performances in both cross-modal and cross-time tasks, with the highest improvement of all tested \texttt{SP} schemes in \emph{CS$\rightarrow$MFNT} (\green{+1.55\%} mIoU) and \emph{CS$\rightarrow$DZ} (\green{+0.53\%} mIoU). It also significantly improves \emph{CS$\rightarrow$CI} by \green{+18.33\%} mIoU.

\textbf{Mixup.} Regularization from \texttt{mixup}  boosts the cross-modal task, \emph{CS$\rightarrow$CI}, with \green{+39.37\%}mIoU, which is \green{2.2$\times$} what we obtain without \texttt{SP}. We hypothesize that regularizing during \texttt{SP} helps prevent the model from biasing toward photon noise and glare in low-light images captured by an intensifier.

\textbf{Blur.} We improve performance across all our source-target pairs with low-light noise. In \emph{CS$\rightarrow$CI} we obtain a boost of \green{+40.64\%}mIoU, which is \green{$2.25\times$} what we obtain with no \texttt{SP}. This indicates that overfitting to high-frequency detail in the source domain can lead to the target model being biased towards similar features, which corresponds mainly to noise in the target domain. 


CityIntensified proved to be very challenging for the baseline source model prior to \texttt{UDA}, at $4.57\%$ mIoU, indicating that of the features learnt by the source model, few were relevant across these domains, i.e. a lot of learnt features were source domain-specific. This resulted in limited improves with \texttt{UDA}. We hypothesize that our \texttt{SP} techniques greatly enhanced UDA because our source models are more adaptable.
Since our \texttt{SP} mainly focuses on mitigating overfitting in the source domain, all results above validate our hypothesis that a source model less biased toward different kinds of source domain-specific characteristics is more suitable for adaptation.

\textbf{Selecting the right \texttt{SP} method.} From the trends we observe, we can extract guidelines for selecting or designing the right \texttt{SP} method for a specific \emph{T}. If \emph{T} has a lot of high-frequency noise, techniques that aim to reduce sensitivity to such noise, like blurring and regularization with \texttt{mixup}, might be appropriate. If there is a significant difference in style between the source and target domains, regularizing over style, as with \texttt{MixStyle}, is an effective \texttt{SP} technique. 

\subsubsection{Effect of Source Preparation on Robustness} \label{Effects of Source Preparation on Robustness}
\begin{figure}[h]
\centering
\includegraphics[scale=0.14]{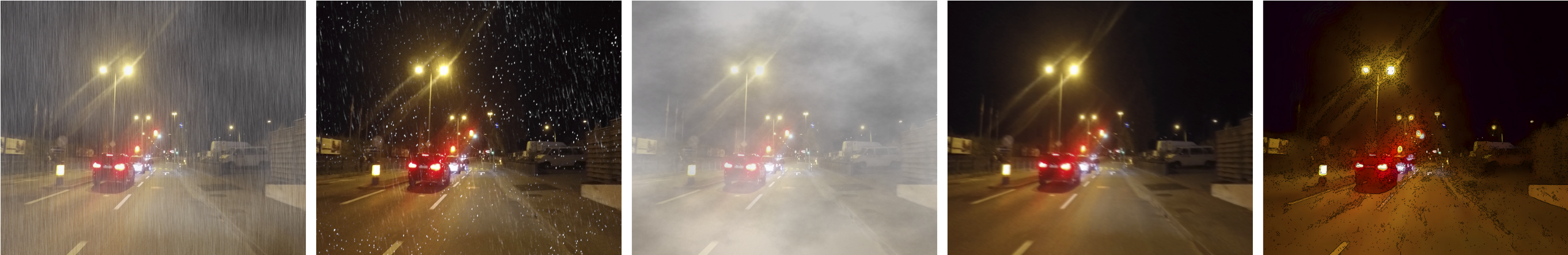}
\caption{Examples from DarkZurich with rain, snow, fog, increased motion blur, and cartoonification.}\label{fig:realistic-robustness-test}
\end{figure}

\texttt{SP} not only enhances performance in the target domain but also increases the robustness of the adapted model to possible real-world changes in the target domain. Our results are detailed in Table \ref{DA-Refign-cs-dz-robustness}, and examples of augmentations, generated using imgaug \cite{imgaug}, are shown in Figure \ref{fig:realistic-robustness-test}. We modify the images in the DarkZurich Val (DZv) set to add rain, fog, snow, and increased motion blur. We also `cartoonify' the images to test across another stylistic variation. In all cases, models obtained after \texttt{UDA} with \texttt{SP}   
beat models obtained without \texttt{SP}, with \green{+5.45\%}mIoU in DZv-rainy, \green{+1.55\%}mIoU in DZv-snowy being examples. We attribute this to reduced sensitivity to noise and stylistic variations.

\subsubsection{Improving Supervised Alignment with Source Preparation} \label{Improving Supervised Alignment with Source Preparation}

We evaluate the performance of models obtained after \texttt{UDA}, with and without \texttt{SP}, after performing \texttt{SA} in the form of finetuning with a very limited number of labeled samples from \emph{T}. We show our results in Table \ref{finetune-cs-mfnt} on \emph{CS$\rightarrow$MFNT} and \emph{CS$\rightarrow$CI} across labeled target train sets of different sizes. In each case, barring \emph{CI} with 100 (comprising its entire train set), we perform four rounds of finetuning on the same randomly subsampled portions of the train set for each method, and subsequently average the results. \texttt{SP} improves performance in \emph{T} after \texttt{SA}, particularly in cases with very few labeled samples from \emph{T}, such as \green{+4.46} mIoU on \emph{MFNT} with 20 samples, \green{+4.66} mIoU on \emph{CI} with 50 samples as compared to finetuning the model without \texttt{SP}, while also generally giving results with less variation. 


\begin{table}
  \caption{Comparison (mIoU) on DarkZurich Val under various potential real-world shifts (and another style-shift) in the target domain after \texttt{UDA} from Cityscapes with different Source Preparation (\texttt{SP}) techniques. \texttt{SP} performs better across all shifts, indicating increased robustness in the target model. }
  \label{DA-Refign-cs-dz-robustness}
  \centering
  \begin{tabular}{llllll}
    \toprule
\texttt{SP} Method     & Rainy & Snowy & Foggy & Motion Blur & Cartoonified\\
    \midrule
    None  &  28.81 & 35.21 & 35.98 & \underline{42.38} & 18.25   \\
    \texttt{MixStyle} & 26.66 & 34.67 & \textbf{36.32} & 40.72 & 18.62 \\

    \texttt{mixup} & \underline{33.04} (\green{+4.23}) & 34.82 & \underline{36.16} & 41.48 & 18.01 \\

    Blur & \textbf{34.26} (\green{+5.45}) & \textbf{36.76} (\green{+1.55}) & 35.66 & \textbf{42.47} & \textbf{20.05} (\green{+1.80})\\

    \bottomrule
  \end{tabular}
\end{table}

\begin{table}[!h]
  \caption{Comparison (mIoU) on respective validation sets after limited Supervised Alignment (SA) of the models with and without Source Preparation (\texttt{SP}), and \texttt{UDA} from Cityscapes. Incorporating \texttt{SA} after both \texttt{SP} and \texttt{UDA} yields the best-performing models in the target domain, particularly when labeled target samples are scarce.}
  \label{finetune-cs-mfnt}
  \centering
  \begin{tabular}{llllll}
    \toprule
    Dataset & \texttt{SP}? & \texttt{UDA}? & \multicolumn{3}{c}{Number of labels for \texttt{SA}}   \\ 
    & & & 20 & 50 & 100 \\
    \midrule
    \midrule
    \multirow{3}{*}{MFNetThermal} & \xmark & \xmark & 66.30 $\pm 1.4$  & 77.25 $\pm 1.2$ & 79.15 $\pm 4.3$ \\ 
    & \xmark & $\checkmark$ & 74.93 $\pm 8.7$  & 84.19 $\pm 3$ & 85.41$\pm 1.3$  \\
    & $\checkmark$ & $\checkmark$ & \textbf{79.39} (\green{+4.46}) $\pm 2.3$ & \textbf{84.46}$\pm 2.1$ & \textbf{85.67} $\pm 0.8$ \\ 
    \midrule
    \multirow{3}{*}{CityIntensified} & \xmark & \xmark & 49.67 $\pm 7.3$  & 58.41  $\pm 3.1$ & 69.82 \\ 
    & \xmark & $\checkmark$ & 73.43 $\pm 1.7$ & 77.03 $\pm 2$  & 80.76\\
    & $\checkmark$ & $\checkmark$ & \textbf{74.29} $\pm 0.8$ & \textbf{81.69} (\green{+4.66})$\pm 1.5$ & \textbf{81.89}  \\ 
    \bottomrule
  \end{tabular}
\end{table}

\subsection{AUDA for Effective Label Efficient Domain Adaptation across large Domain Gaps} \label{Label Efficient Learning}


\textbf{Stage-wise contributions in \texttt{AUDA}.} We present experimental results of our proposed framework, \texttt{AUDA}, detailing the contributions of each step in Table \ref{Ablations-3stage} , demonstrating their positive impact. Across different source-target pairs, different stages are most effective. In \emph{CS$\rightarrow$DZ}, \texttt{UDA} improves target performance the most, at \green{+19.67\%} mIoU, while \texttt{SA} does so in \emph{CS$\rightarrow$MFNT}, with \green{+17.92\%} mIoU, and \texttt{SP} in \emph{CS$\rightarrow$CI}, \texttt{SP} increases target domain performance by \green{+40.64\%} mIoU.

\begin{table}
  \caption{Analysing the contribution of each stage of \texttt{AUDA}, with 50 labeled target samples of \texttt{SA}, with their improvements (mIoU) \green{highlighted}. Results shown over respective validation sets. }
  \label{Ablations-3stage}
  \centering
  \begin{tabular}{llll}
    \toprule
    Method & DarkZurich & MFNetThermal & CityIntensified \\
    \midrule
    Baseline & 29.30 &  55.36 & 4.57 \\
    \texttt{UDA} &  48.97 (\green{+19.67}) & 63.45 (\green{+8.09}) & 32.50 (\green{+27.93}) \\
    \texttt{SP} + \texttt{UDA} & 49.50 (\green{+0.53}) & 65.00 (\green{+1.55}) & 73.14 (\green{+40.64}) \\ 
    \texttt{SP} + \texttt{UDA} + \texttt{SA} & N/A & 82.92 (\green{+17.92}) & 82.61 (\green{+9.47})\\
    \bottomrule
  \end{tabular}
\end{table}

\textbf{Necessity of \texttt{SP}.} We compare \texttt{SP} techniques applied directly during \texttt{UDA} and in a separate \texttt{SP} step to investigate the necessity of a preparatory step. Results in Table \ref{Ablations-SP-cs-target} show that a separate \texttt{SP} step consistently yields superior outcomes, supporting our hypothesis that source models need to be made more adaptable before \texttt{UDA}.
\begin{table}[t]
  \caption{Comparison (mIoU) on respective validation sets with the best performing \texttt{SP} technique for each dataset applied as a separate \texttt{SP} step before \texttt{UDA} or together with \texttt{UDA}. Results indicate that a separate \texttt{SP} generally yields superior target models. }
\label{Ablations-SP-cs-target}
\centering
\begin{tabular}{cclll}
\toprule
\texttt{SP}?     & \texttt{SP}  modification during \texttt{UDA}?   &  DarkZurich & MFNetThermal & CityIntensified \\
\midrule
\xmark & \xmark  & 48.97 & 63.45 & 32.50\\
\xmark & $\checkmark$ & 48.39 & 65.27 & 39.97 \\
$\checkmark$ & \xmark & \textbf{49.50} & 65.00 & \textbf{73.14}\\ 
$\checkmark$ & $\checkmark$ & 47.55 & \textbf{65.45} & 51.94 \\ 
\bottomrule
\end{tabular}
\end{table}

\textbf{Comparisons with \texttt{FSSDA}.} In Table \ref{ComparisonWithFSSDA}, we compare \texttt{AUDA} with an instantiation of \texttt{FSSDA}, \texttt{PixDA} \cite{Tavera_2022_WACV_PixDA}, on \emph{CS$\rightarrow$CI} and \emph{CS$\rightarrow$MFNT}. We provide both approaches access to the same set of labeled data from the target domain (20 labeled samples in \emph{CS$\rightarrow$MFNT}, 50 in \emph{CS$\rightarrow$CI}) during training, and report the best of 1-shot and 5-shot performance during evaluation. The backbone segmentation networks are however different with \texttt{PixDA} using \texttt{DeepLabv2}, and \texttt{AUDA} using \texttt{SegFormer(MiTB5)}. 
\texttt{AUDA} performs significantly better in both these cases, indicating having a greater ability to adapt across larger domain gaps, which we attribute to \texttt{SP}, and exploitation of unlabeled target samples. \texttt{SSDA} approaches typically use hundreds of labeled target samples, and cannot be utilized in these scenarios. 

\begin{table}[h]
  \caption{Comparison (mIoU) between \texttt{AUDA} and \texttt{PixDA}. Results shown for validation sets of respective datasets. These indicate that \texttt{AUDA} can adapt more effectively across larger domain gaps.}
  \label{ComparisonWithFSSDA}
  \centering
  \begin{tabular}{lll}
    \toprule
    Method & MFNetThermal & CityIntensified \\
    \midrule
    \texttt{PixDA} & 17.14  & 16.49   \\
    \texttt{AUDA} & 75.46 &   82.61  \\
    \bottomrule
  \end{tabular}
\end{table}



\section{Conclusion}

In this work, we introduce Source Preparation, a method to account for source domain specific characteristics, and enhance Domain Adaptation by preparing an `adaptable' source model. Source Preparation improves performance of models across diverse domains, while also improving robustness to real-world shifts within each domain. Our label-efficient Domain Adaptation framework, Almost Unsupervised Domain Adaptation further accounts for robotic scenarios through Supervised Alignment, such as off-road environments in our CityIntensified dataset, where limited labeled target data can be obtained. 


\textbf{Limitations and Future Work.} While we propose some design principles for designing new Source Preparation techniques, automatically learning or selecting the optimal source preparation technique from data itself remains an open challenge.

\textbf{Societal Impact.} Our method has implications for extending the functionality and operating range of robots. However, they can also do the same for surveillance systems. 


\clearpage

\appendix

\section{AUDA for Label Efficient Domain Adaptation: A Comparison Against SSDA Approaches}


While we previously stated that \texttt{SSDA} approaches typically use hundreds of labeled target domain samples for domain adaptation, in this section we show that their performance degrades rapidly in the limited label scenarios we are working with. 

We compare \texttt{AUDA} with two different \texttt{SSDA} approaches. First, we modify \texttt{Refign-HRDA*} \cite{bruggemann2023refign} such that we add target image label pairs along with the source image label pairs as a part of the task-supervision, in addition to providing unlabeled target samples for alignment just as before. We term this \texttt{Refign-SS} and train it with the same scheme and hyper-parameters we use to train \texttt{Refign} as a \texttt{UDA} method in \texttt{AUDA}. For the second, we modify \texttt{USSS} \cite{kalluri2019universal} by replacing \texttt{DRNet} \cite{Yu2017DRN} (as used in the original paper) with \texttt{SegFormer} \cite{xie2021segformer} to ensure a fair comparison, and improve it by doing so. While \texttt{USSS} assumes partial annotations in both (source and target) domains, we provide all available source domain annotation, in addition to limited target annotations, and all unlabeled samples in both domains. We train \texttt{USSS} with its official code release, using all associated hyper-parameters. In all our comparisons we provide access to the same randomly selected sets of labeled target samples of different sizes. MFNetThermal \cite{MFNet-thermal} (\emph{MFNT}) with 1567, and  CityIntensified (\emph{CI}) with 100 target samples correspond to utilizing the entirety of their respective train sets.

Our results over the validation set of each dataset, shown in Table \ref{supplemental:comparison-ssda}, clearly demonstrate that \texttt{AUDA} performs much better in label scarce scenarios than other approaches in our comparison, with up to \green{+34.75\%} mIoU in \emph{MFNT} with 20 target labels. Moreover, it is important to note that with an increase in target label scarcity, the degradation in performance is far steeper in existing \texttt{SSDA} approaches as compared to \texttt{AUDA}. In \emph{MFNT}, \texttt{AUDA} reaches \green{$92.5\%$} of its performance with the full-training set, with just 20 labeled target domain samples, as compared with \texttt{Refign-SS} reaching just $63.4\%$ of its performance under full-supervision. Similarly, in \emph{CI}, \texttt{AUDA} reaches \green{$89.2\%$} of its performance under full-supervision as compared to \texttt{Refign-SS}'s $82.1\%$, both with 20 labeled target samples.
This is in addition to \texttt{AUDA}'s ability to be used in a target label-free scenario as well, with only the use of the first two of the three steps, i.e. \texttt{SP} and \texttt{UDA}. This is important for environments and tasks where iterative development is critical, as this provides us with the ability to first train a model in the new target domain without any supervision, deploy it, and iteratively improve it with \texttt{SA}, without having to retrain it entirely.  

\begin{table}[!h]
  \caption{Comparison (mIoU) to showcase label-efficiency of \texttt{AUDA} vs other \texttt{SSDA} approaches. \texttt{AUDA} not only performs better under label scarcity, but the degradation in performance as we approach label scarcity is also reduced.}
  \label{supplemental:comparison-ssda}
  \centering
  \begin{tabular}{llcccc}
    \toprule
    Dataset & Method & \multicolumn{4}{c}{Number of labels for \texttt{SA}}   \\ 
    & & 20 & 50 & 100 & 1567 \\
    \midrule
    \midrule
    \multirow{3}{*}{MFNetThermal} & \texttt{USSS} & 29.10 & 35.30 & 43.15 & 59.10 \\
    & \texttt{Refign-SS} & 46.05 & 61.44 & 68.32 & 72.55 \\ 
     & \texttt{AUDA}  & \textbf{80.80}  (\green{+34.75\%}) & \textbf{84.10} (\green{+22.66\%})  & \textbf{85.04} & \textbf{87.34} \\ 
    \midrule
    \multirow{3}{*}{CityIntensified} 
    & \texttt{USSS} & 67.51 & 71.04 & 73.94 & - \\
    & \texttt{Refign-SS} & 69.40 & 80.69 & \textbf{84.47} & - \\ 
    & \texttt{AUDA}  & \textbf{74.25} & \textbf{82.61}  & 83.20 & -  \\ 
    \bottomrule
  \end{tabular}
\end{table}

\section{SP with an Alternate Domain Adaptation Approach}

\begin{table}[!h]
  \caption{Comparison (mIoU) to showcase the effect of \texttt{SP} on a different Domain Adaptation technique, \texttt{USSS}. \texttt{SP} shows that it can boost performance across both datasets and different levels of label scarcity. }
  \label{supplemental:SP-alternate-network}
  \centering
  \begin{tabular}{llcccc}
    \toprule
    Dataset & \texttt{SP}? & \multicolumn{4}{c}{Number of labels for \texttt{SA}}   \\ 
    & & 20 & 50 & 100 & 1567 \\
    \midrule
    \midrule
    \multirow{2}{*}{MFNetThermal}  
    & \xmark  & \textbf{29.10}  & \textbf{35.30} & 43.15 & 59.10 \\ 
    & $\checkmark$ & 26.58 & 30.87 & \textbf{49.98} (\green{+6.83\%}) & \textbf{67.43} (\green{+8.33\%}) \\
    \midrule
    \multirow{2}{*}{CityIntensified} 
    & \xmark & 66.91 & \textbf{71.02} & 73.67 & -  \\ 
    & $\checkmark$ & \textbf{67.61} & 70.25 & \textbf{76.29} (\green{+2.62\%}) & - \\ 
    \bottomrule
  \end{tabular}
\end{table}

To test the ability of \texttt{SP} in improving other techniques and approaches to domain adaptation, we run the modified \texttt{USSS} algorithm from above (an approach to \texttt{SSDA}), with and without a source prepared source model trained on Cityscapes \cite{cordts2016cityscapes} (\emph{CS}). We report the outcome of these experiments in Table \ref{supplemental:SP-alternate-network}. Our results indicate that \texttt{SP} can significantly improve performance across these semi-supervised domain adaptation techniques as well, with \green{+8.33\%} and \green{+6.83\%} in mIoU in \emph{CS $\rightarrow$ MFNT}  with 1567 and 100 labeled target samples respectively, and \green{+2.62\%} mIoU in \emph{CS $\rightarrow$ CI} with 100 labels.

\section{Effect of Naive Stacking of Different SP Schemes.}

While approaching source preparation with the intention to reduce different forms of source domain biases at the same time may be effective, naively performing all of our \texttt{SP} schemes together, i.e. naively stacking our \texttt{SP} schemes, does not work very well. We show the results of our experiments in Table \ref{supplemental:SP-stacking}, in which we compare the performances of the model obtained after the best single \texttt{SP} method for each dataset with \texttt{SP-stacked} after \texttt{UDA}. In each case, we can see that our performance degrades upon naively stacking \texttt{SP} methods, indicating that some consideration is necessary while designing new \texttt{SP} schemes. 

\begin{table}[!h]
  \caption{Comparison (mIoU) to showcase the effect of chaining our \texttt{SP} schemes vs best individual \texttt{SP} scheme for each dataset after \texttt{UDA}.}
  \label{supplemental:SP-stacking}
  \centering
  \begin{tabular}{lll}
    \toprule
    Dataset & \texttt{SP-Single} & \texttt{SP-Stacked}   \\ 
    \midrule
    DarkZurich  & 49.50 & 43.54 \\
    MFNetThermal  & 65.00 & 60.94 \\
    CityIntensified & 73.14 & 49.71 \\
    \bottomrule
  \end{tabular}
\end{table}

\section{Analyzing Qualitative Results}

\begin{figure}[h]
    \centering
    \begin{subfigure}[b]{0.3\textwidth}
        \includegraphics[width=\linewidth]{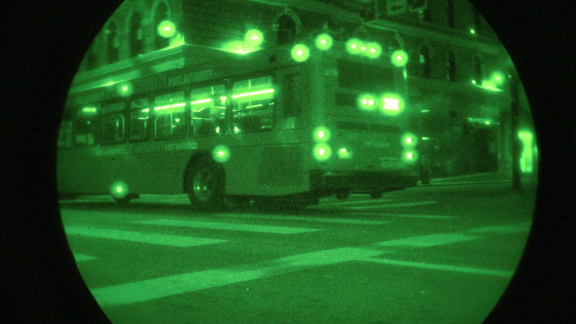}
    \end{subfigure}
    \hfill
    \begin{subfigure}[b]{0.3\textwidth}\includegraphics[width=\linewidth]{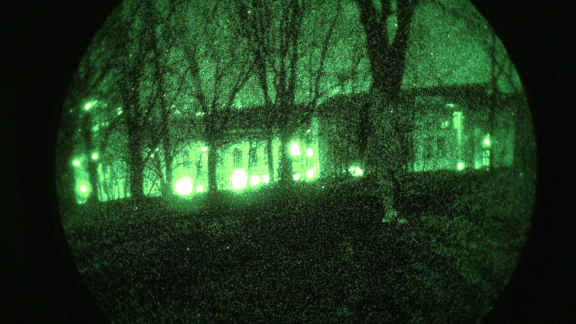}
    \end{subfigure} 
    \hfill
    \begin{subfigure}[b]{0.3\textwidth}
        \includegraphics[width=\linewidth]{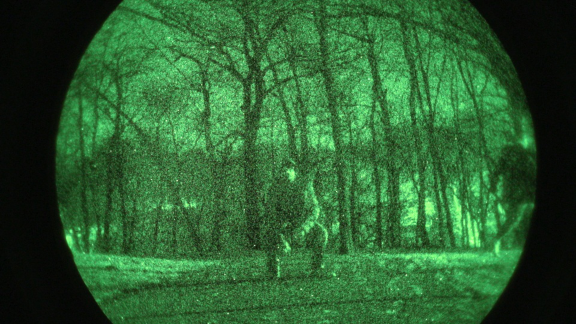}
    \end{subfigure}
    \begin{subfigure}[b]{0.3\textwidth}
        \includegraphics[width=\linewidth]{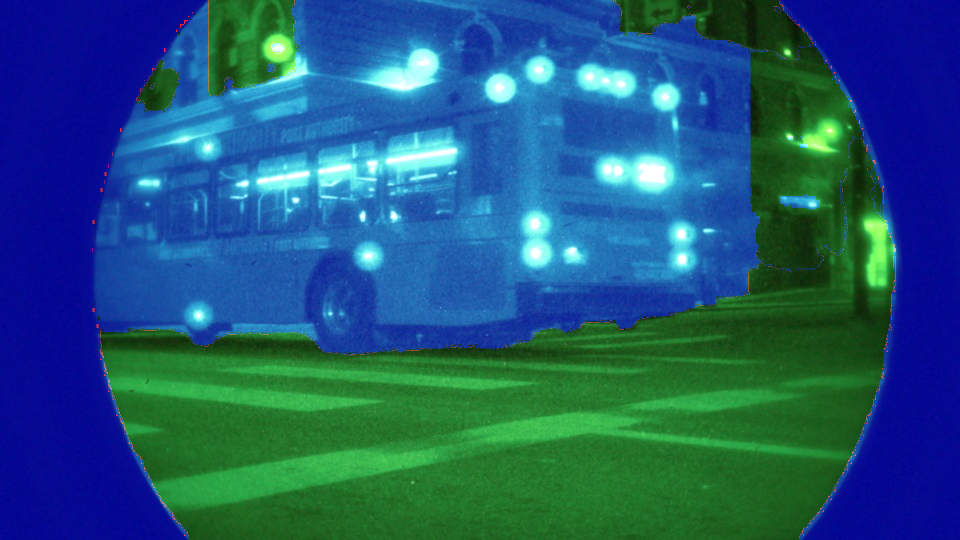}
    \end{subfigure}
    \hfill
    \begin{subfigure}[b]{0.3\textwidth}\includegraphics[width=\linewidth]{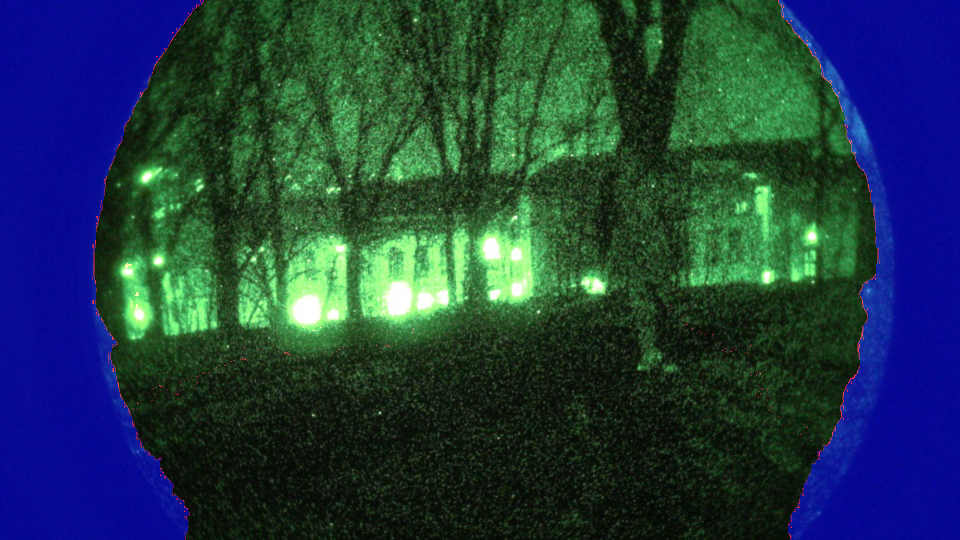}
    \end{subfigure} 
    \hfill
    \begin{subfigure}[b]{0.3\textwidth}
        \includegraphics[width=\linewidth]{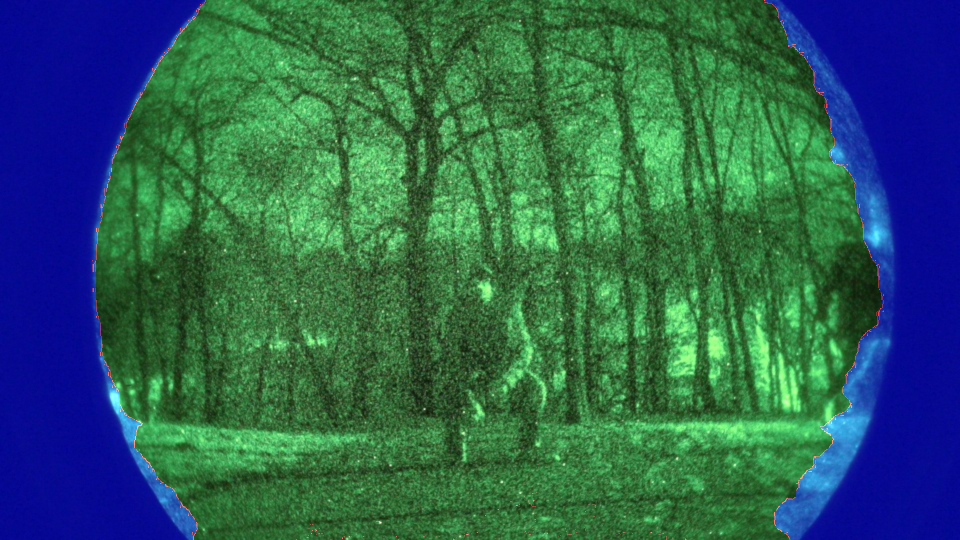}
    \end{subfigure}

    \begin{subfigure}[b]{0.3\textwidth}
        \includegraphics[width=\linewidth]{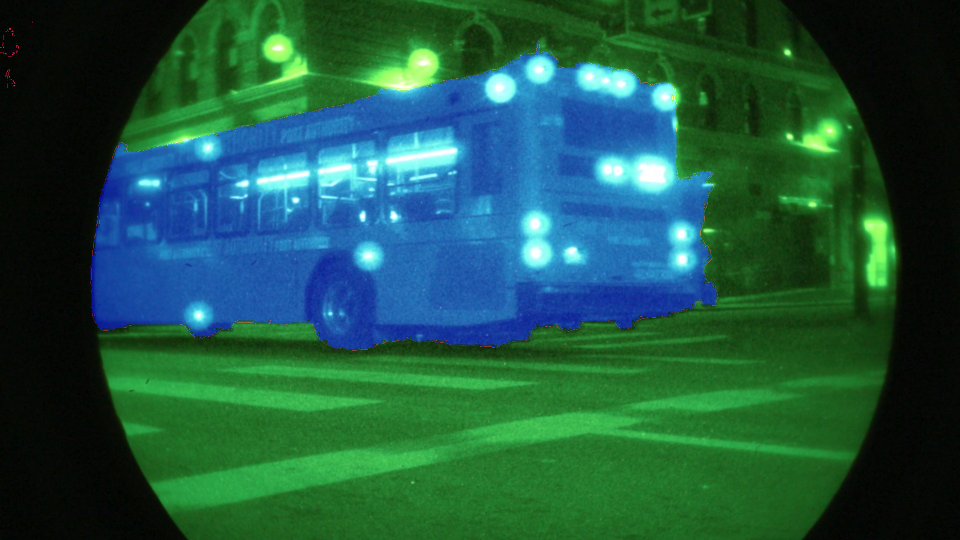}
    \end{subfigure}
    \hfill
    \begin{subfigure}[b]{0.3\textwidth}\includegraphics[width=\linewidth]{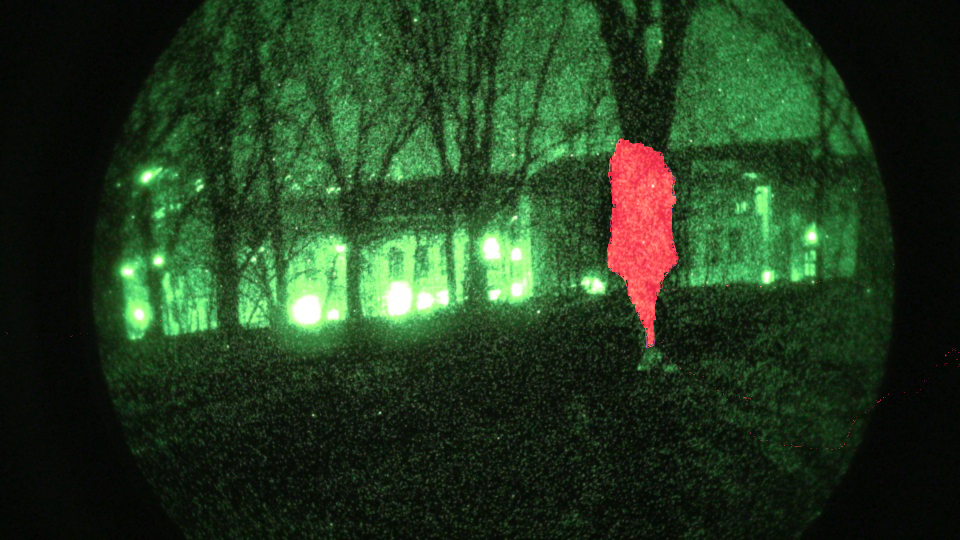}
    \end{subfigure} 
    \hfill
    \begin{subfigure}[b]{0.3\textwidth}
        \includegraphics[width=\linewidth]{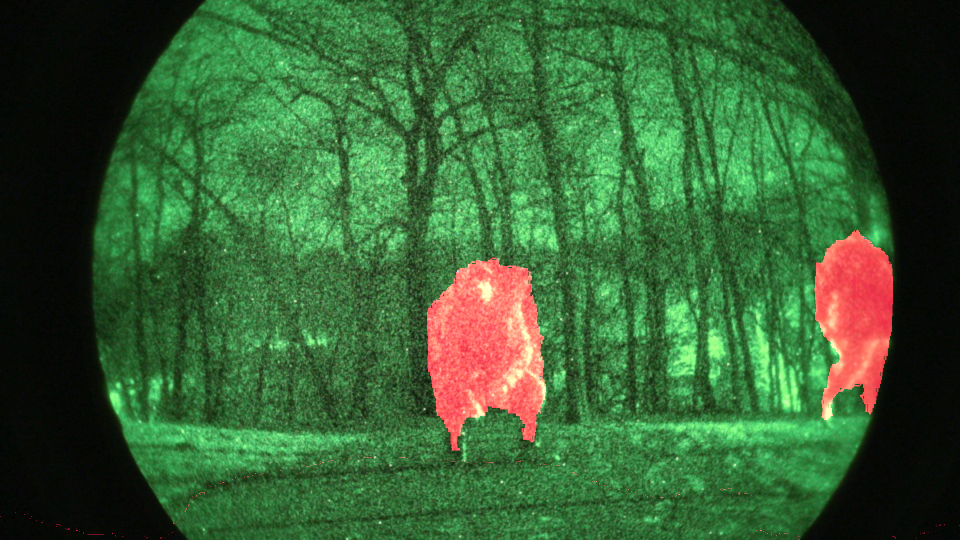}
    \end{subfigure}

    \begin{subfigure}[b]{0.3\textwidth}
        \includegraphics[width=\linewidth]{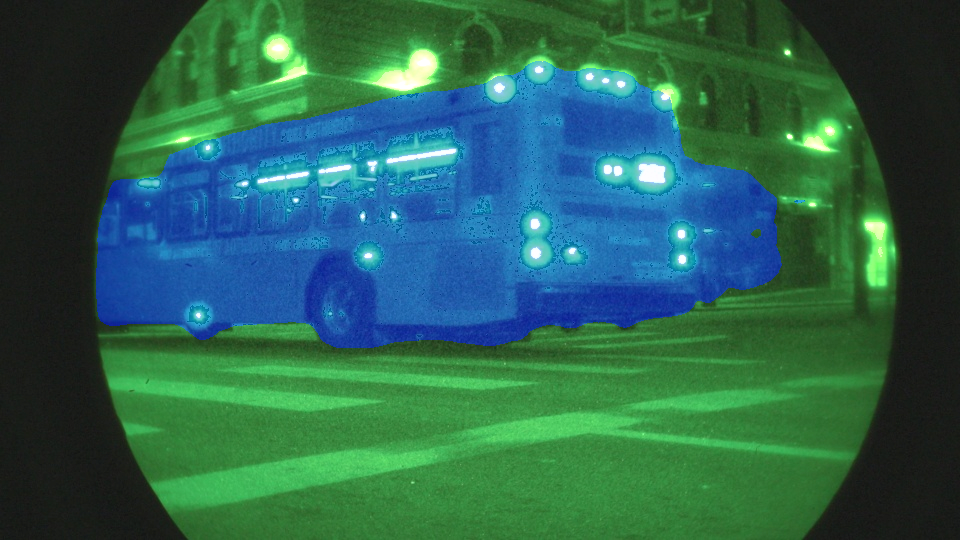}
    \end{subfigure}
    \hfill
    \begin{subfigure}[b]{0.3\textwidth}\includegraphics[width=\linewidth]{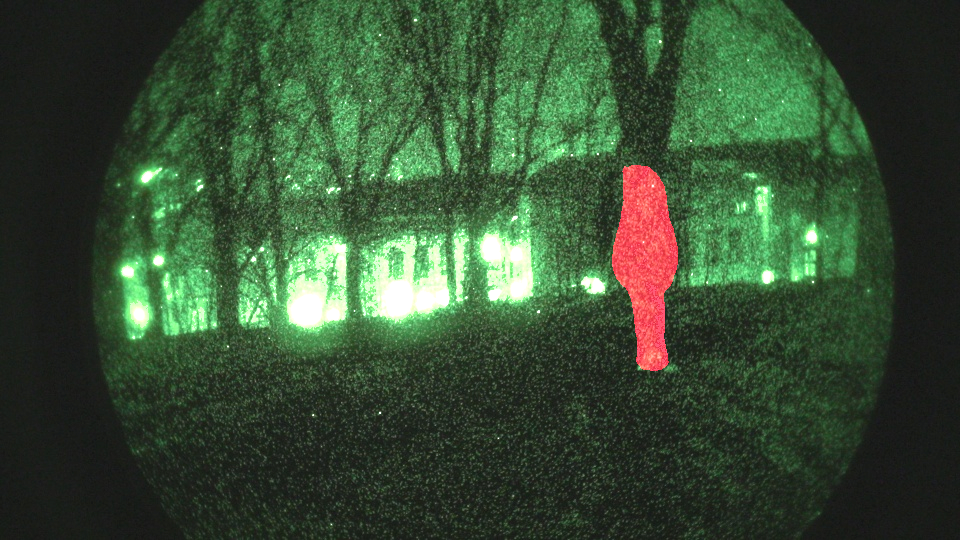}
    \end{subfigure} 
    \hfill
    \begin{subfigure}[b]{0.3\textwidth}
        \includegraphics[width=\linewidth]{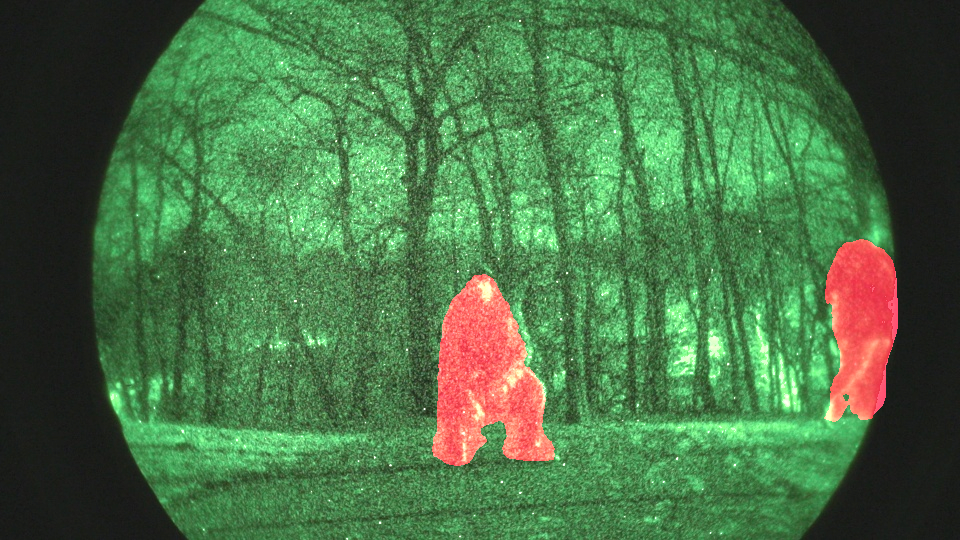}
    \end{subfigure}

    \begin{subfigure}[b]{0.3\textwidth}
        \includegraphics[width=\linewidth]{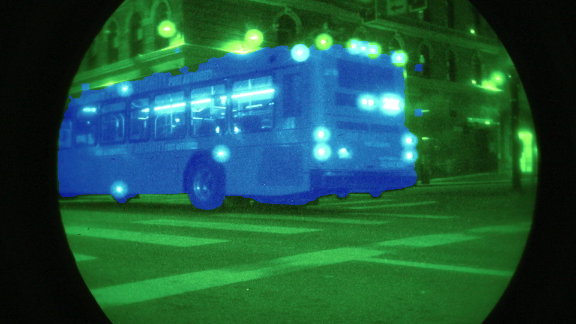}
    \end{subfigure}
    \hfill
    \begin{subfigure}[b]{0.3\textwidth}\includegraphics[width=\linewidth]{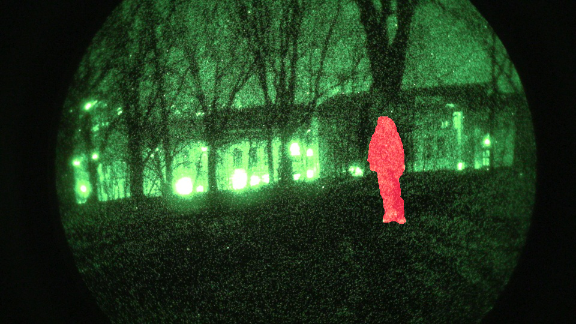}
    \end{subfigure} 
    \hfill
    \begin{subfigure}[b]{0.3\textwidth}
        \includegraphics[width=\linewidth]{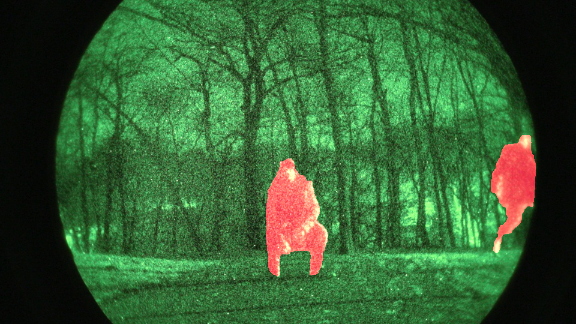}
    \end{subfigure}
    
    \caption{Stage-wise qualitative results, with first row corresponding to query image, second to baseline \texttt{UDA} results, third to \texttt{UDA} with \texttt{SP}, fourth with \texttt{UDA}, \texttt{SP}, and \texttt{SA}, i.e. \texttt{AUDA}, and the last corresponding to ground truth labeling. It is clear that each step of \texttt{AUDA}, critically \texttt{SP}, improves our target domain outputs.}
    \label{fig:qualitative-results1}
\end{figure}
\begin{figure}[h]
    \centering
    \begin{subfigure}[b]{0.3\textwidth}
        \includegraphics[width=\linewidth]{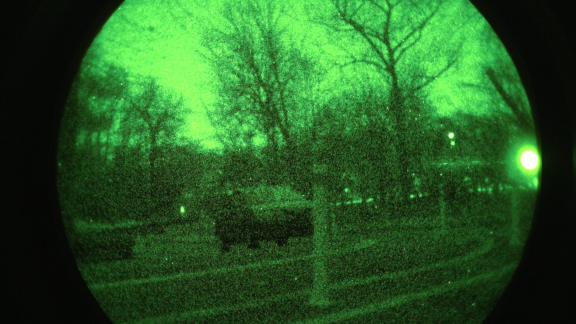}
    \end{subfigure}
    \hfill
    \begin{subfigure}[b]{0.3\textwidth}\includegraphics[width=\linewidth]{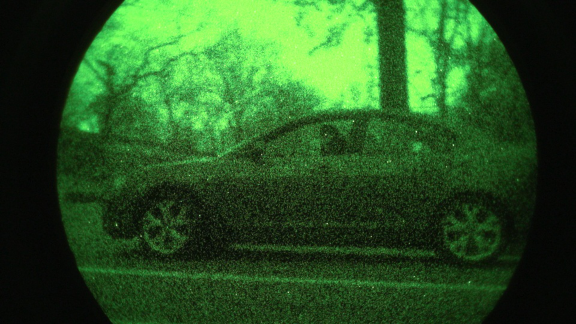}
    \end{subfigure} 
    \hfill
    \begin{subfigure}[b]{0.3\textwidth}
        \includegraphics[width=\linewidth]{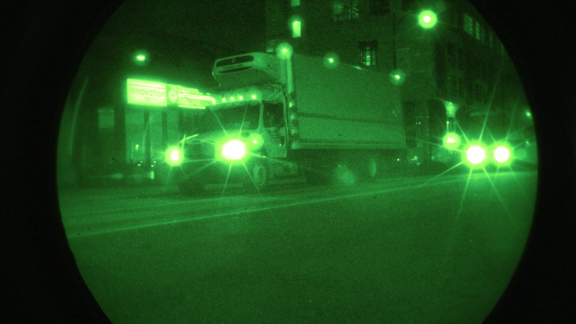}
    \end{subfigure}
    \begin{subfigure}[b]{0.3\textwidth}
        \includegraphics[width=\linewidth]{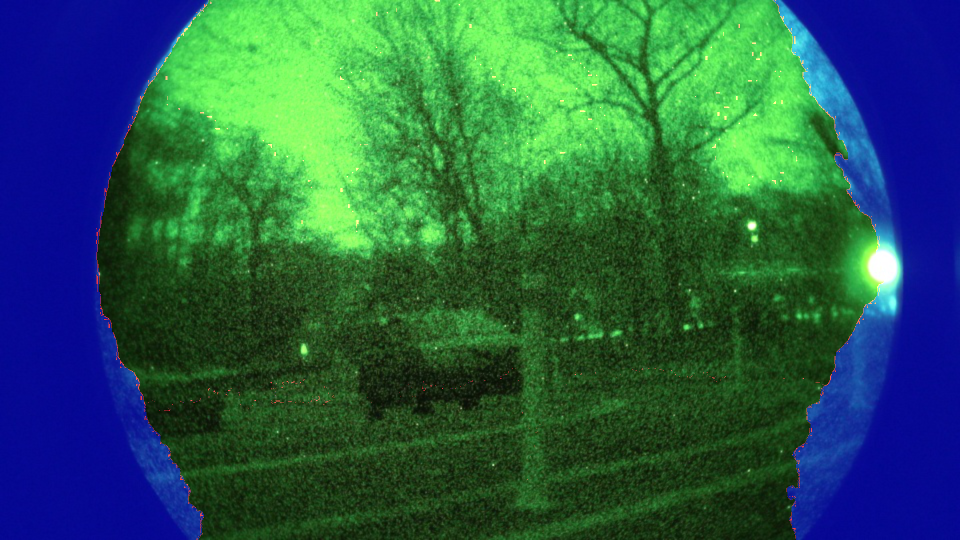}
    \end{subfigure}
    \hfill
    \begin{subfigure}[b]{0.3\textwidth}\includegraphics[width=\linewidth]{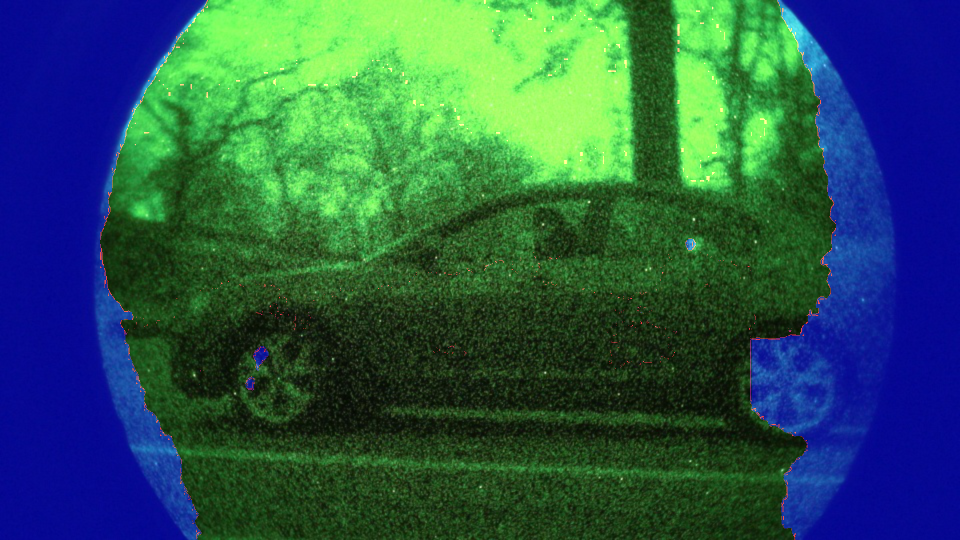}
    \end{subfigure} 
    \hfill
    \begin{subfigure}[b]{0.3\textwidth}
        \includegraphics[width=\linewidth]{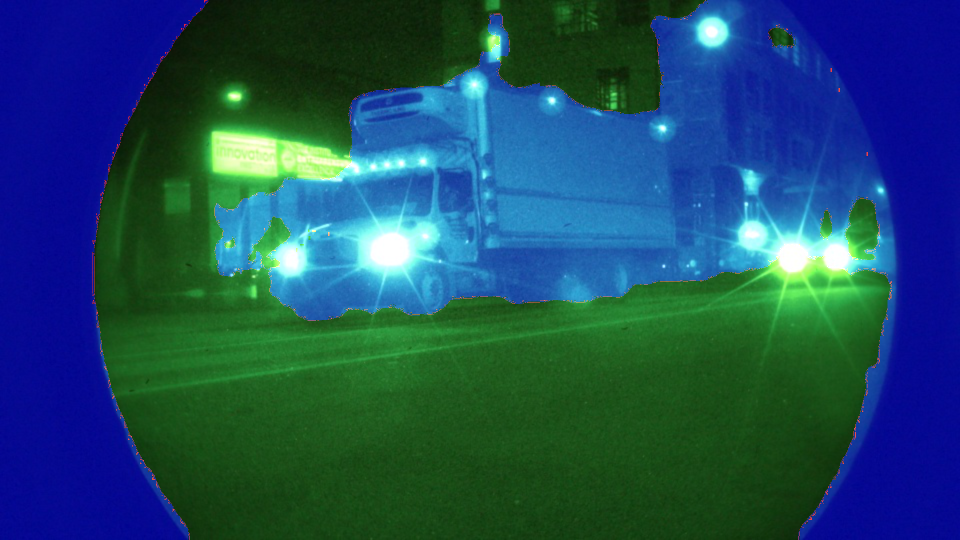}
    \end{subfigure}

    \begin{subfigure}[b]{0.3\textwidth}
        \includegraphics[width=\linewidth]{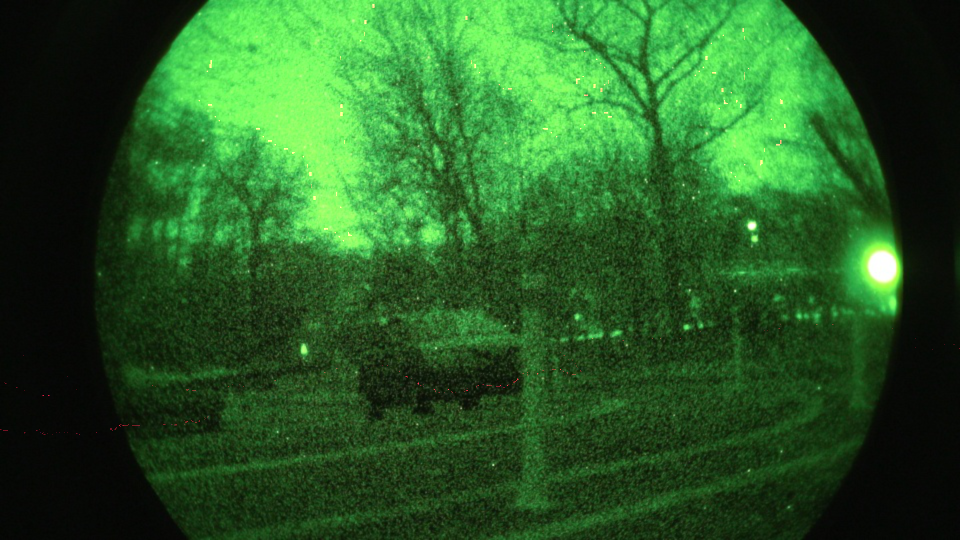}
    \end{subfigure}
    \hfill
    \begin{subfigure}[b]{0.3\textwidth}\includegraphics[width=\linewidth]{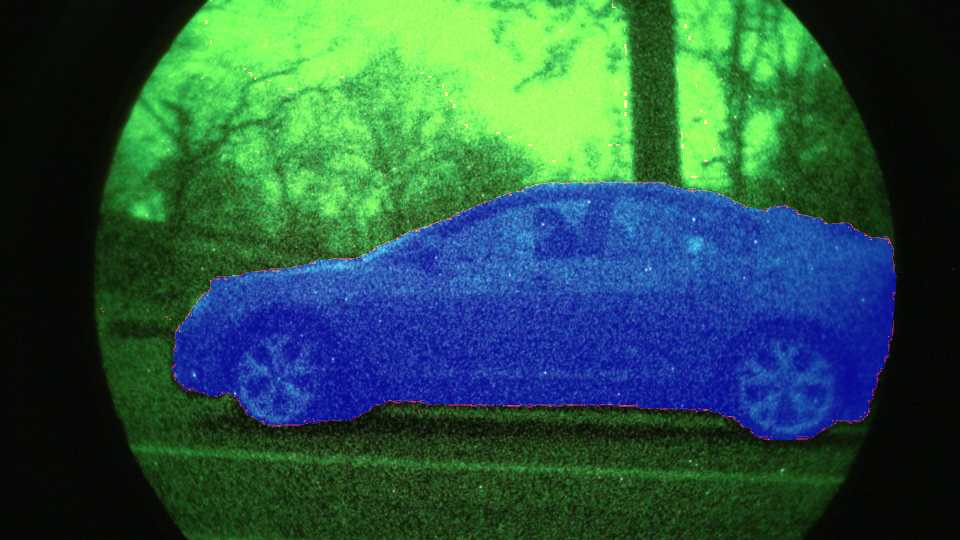}
    \end{subfigure} 
    \hfill
    \begin{subfigure}[b]{0.3\textwidth}
        \includegraphics[width=\linewidth]{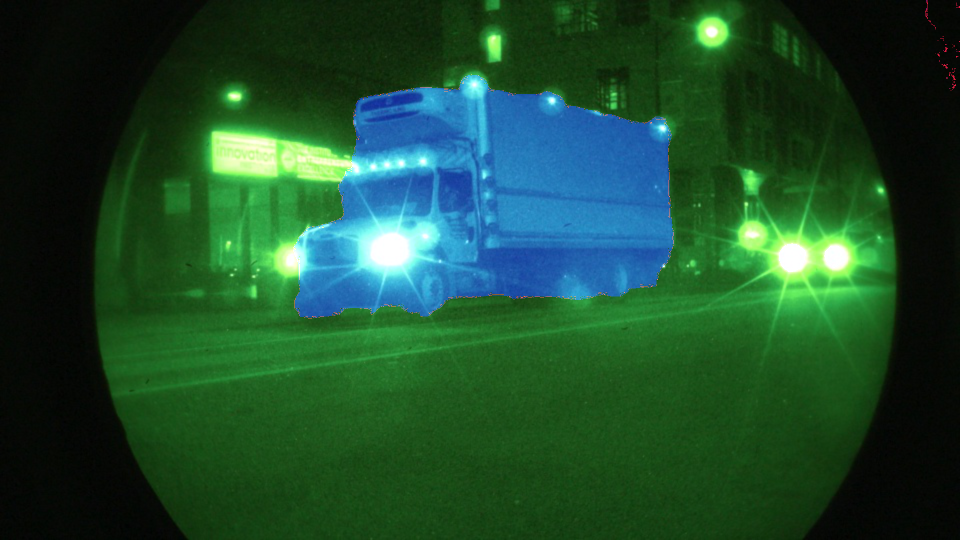}
    \end{subfigure}

    \begin{subfigure}[b]{0.3\textwidth}
        \includegraphics[width=\linewidth]{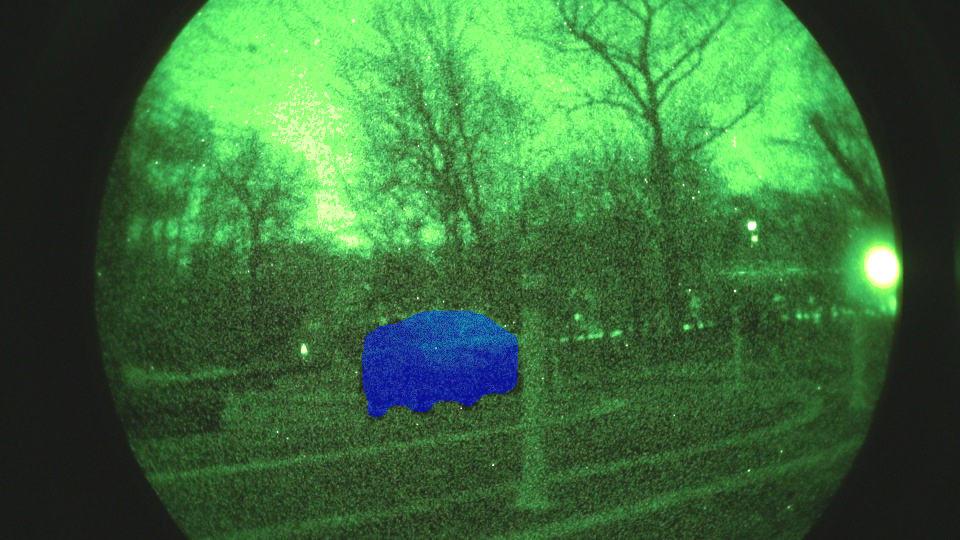}
    \end{subfigure}
    \hfill
    \begin{subfigure}[b]{0.3\textwidth}\includegraphics[width=\linewidth]{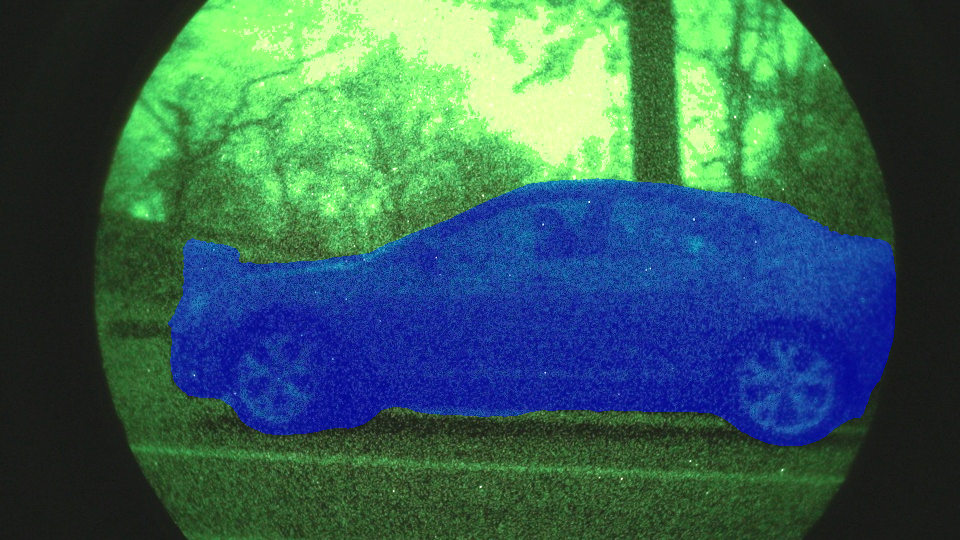}
    \end{subfigure} 
    \hfill
    \begin{subfigure}[b]{0.3\textwidth}
        \includegraphics[width=\linewidth]{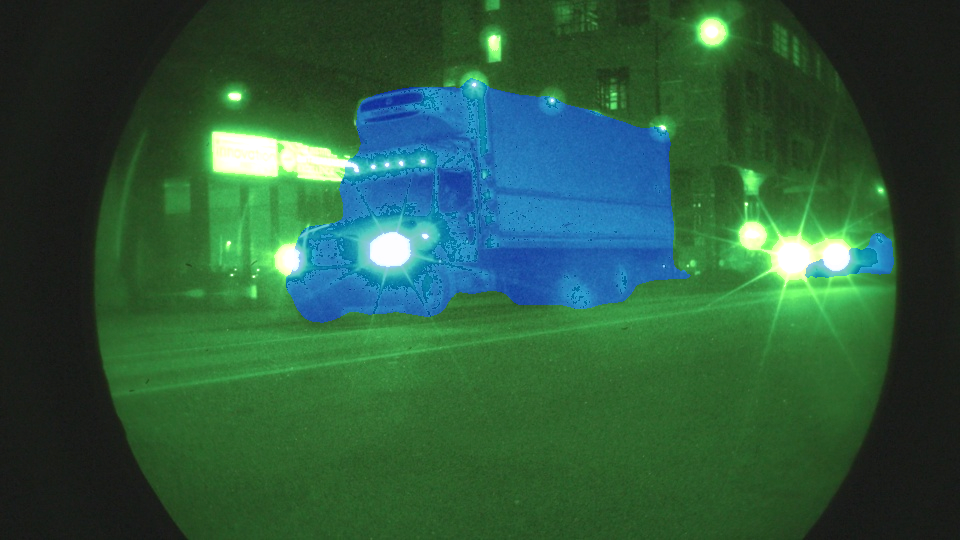}
    \end{subfigure}

    \begin{subfigure}[b]{0.3\textwidth}
        \includegraphics[width=\linewidth]{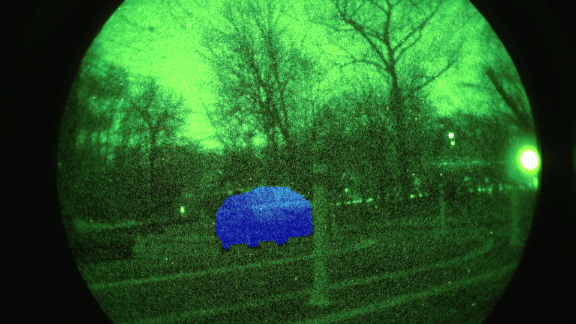}
    \end{subfigure}
    \hfill
    \begin{subfigure}[b]{0.3\textwidth}\includegraphics[width=\linewidth]{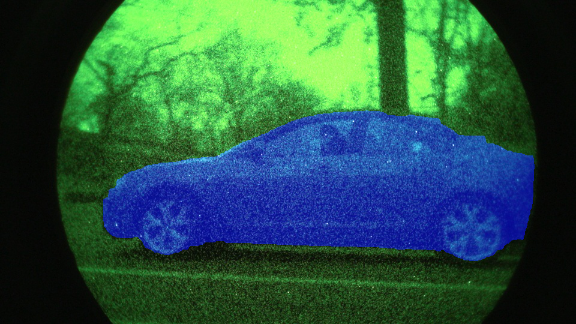}
    \end{subfigure} 
    \hfill
    \begin{subfigure}[b]{0.3\textwidth}
        \includegraphics[width=\linewidth]{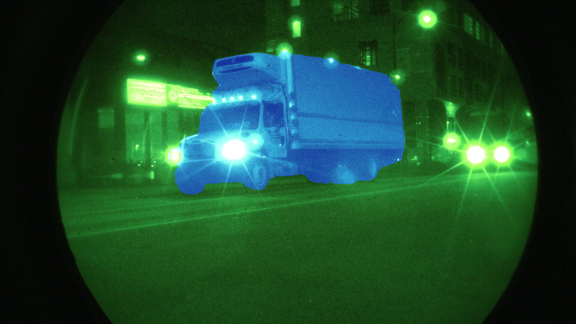}
    \end{subfigure}
    
    \caption{Stage-wise qualitative results, with first row corresponding to query image, second to baseline \texttt{UDA} results, third to \texttt{UDA} with \texttt{SP}, fourth with \texttt{UDA}, \texttt{SP}, and \texttt{SA}, i.e. \texttt{AUDA}, and the last corresponding to ground truth labeling. It is clear that each step of \texttt{AUDA}, critically \texttt{SP}, improves our target domain outputs.}
    \label{fig:qualitative-results2}
\end{figure}
\begin{figure}[h]
    \centering
    \begin{subfigure}[b]{0.3\textwidth}
        \includegraphics[width=\linewidth]{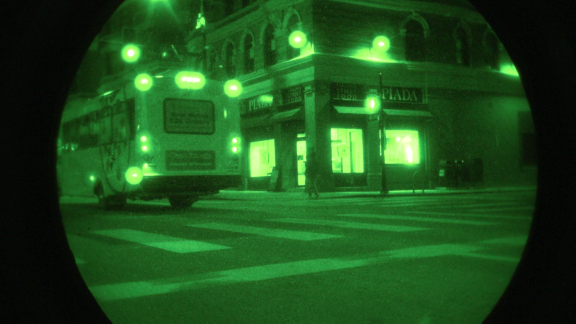}
    \end{subfigure}
    \hfill
    \begin{subfigure}[b]{0.3\textwidth}\includegraphics[width=\linewidth]{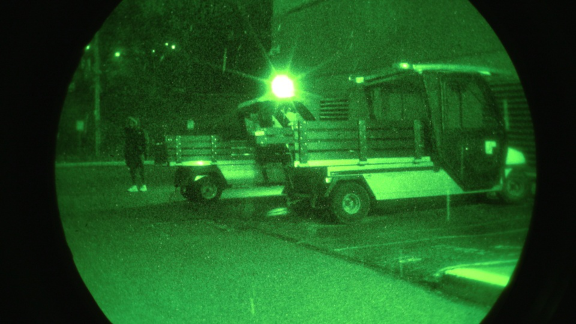}
    \end{subfigure} 
    \hfill
    \begin{subfigure}[b]{0.3\textwidth}
        \includegraphics[width=\linewidth]{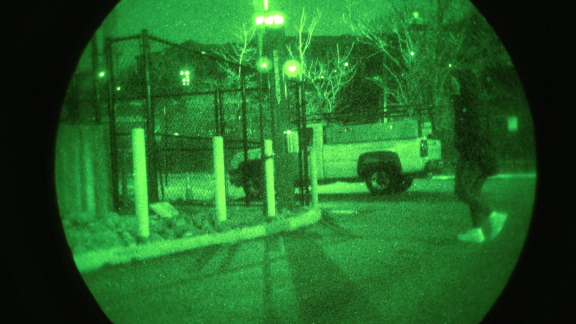}
    \end{subfigure}
    \begin{subfigure}[b]{0.3\textwidth}
        \includegraphics[width=\linewidth]{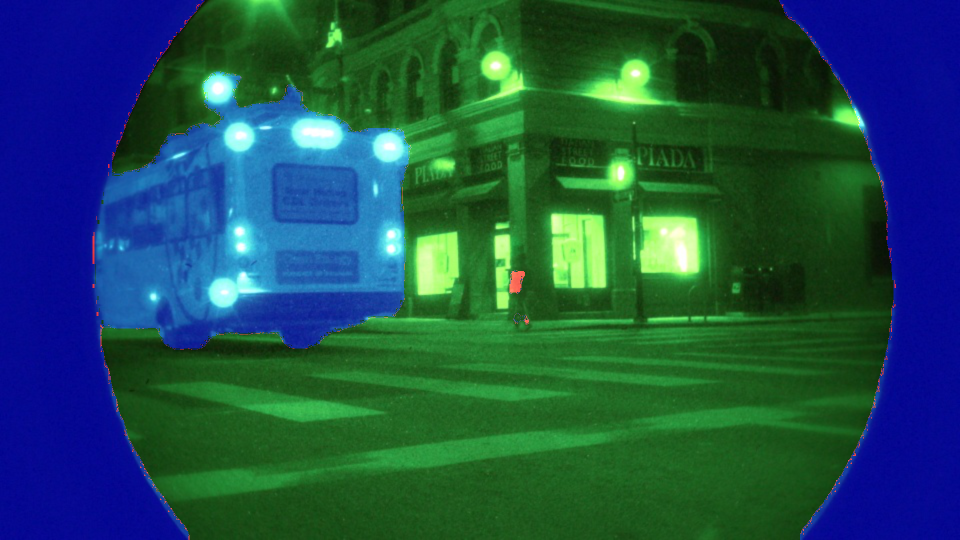}
    \end{subfigure}
    \hfill
    \begin{subfigure}[b]{0.3\textwidth}\includegraphics[width=\linewidth]{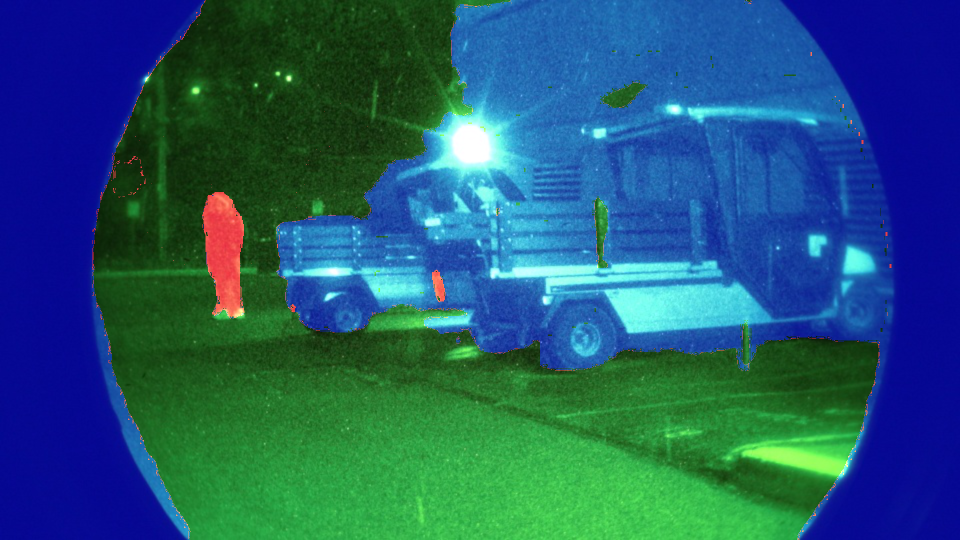}
    \end{subfigure} 
    \hfill
    \begin{subfigure}[b]{0.3\textwidth}
        \includegraphics[width=\linewidth]{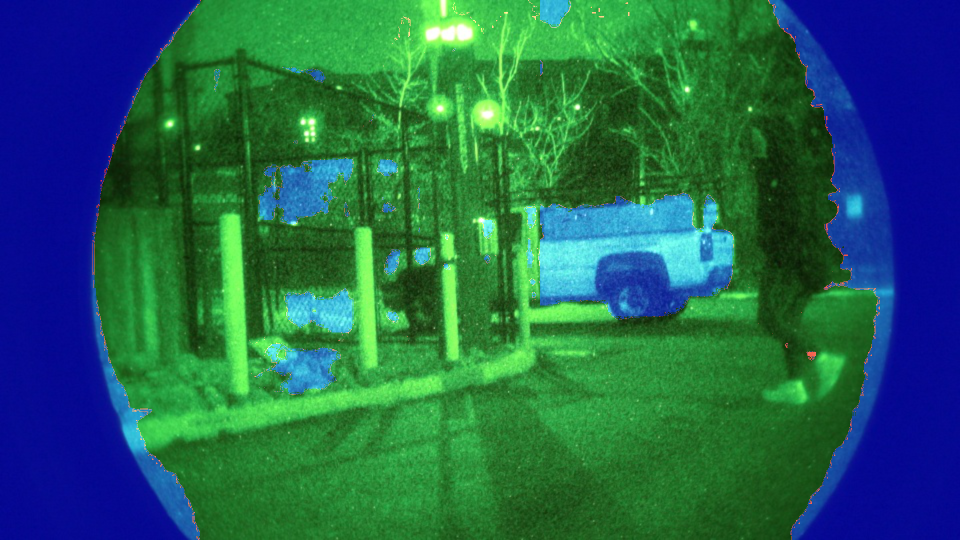}
    \end{subfigure}

    \begin{subfigure}[b]{0.3\textwidth}
        \includegraphics[width=\linewidth]{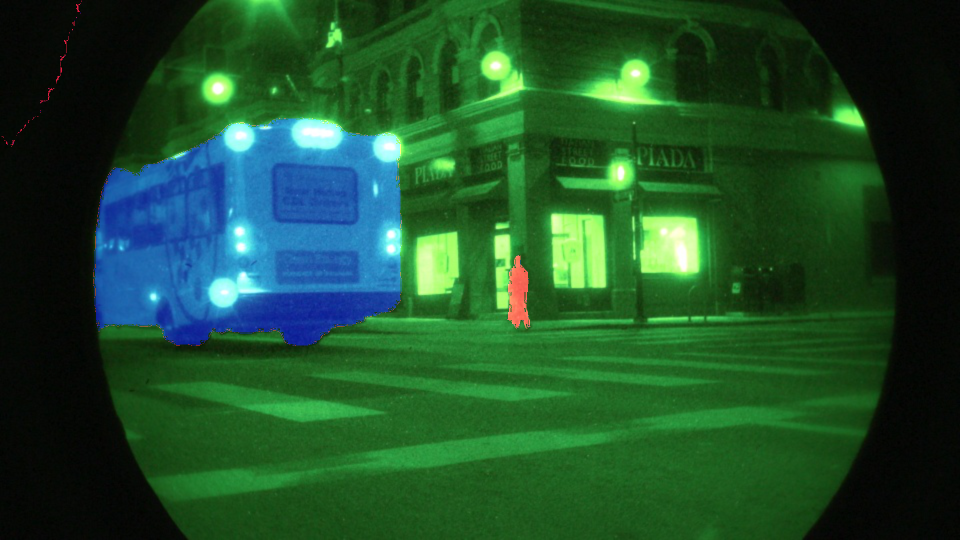}
    \end{subfigure}
    \hfill
    \begin{subfigure}[b]{0.3\textwidth}\includegraphics[width=\linewidth]{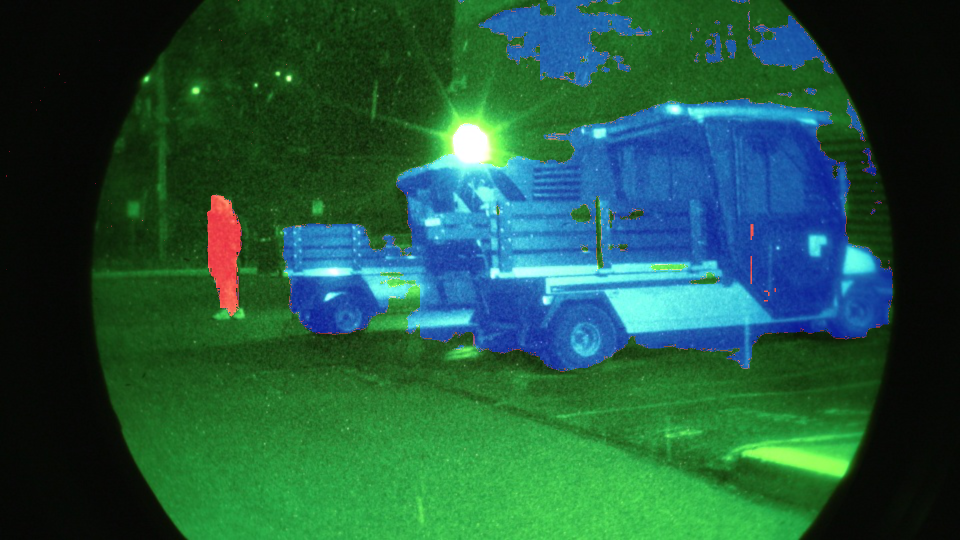}
    \end{subfigure} 
    \hfill
    \begin{subfigure}[b]{0.3\textwidth}
        \includegraphics[width=\linewidth]{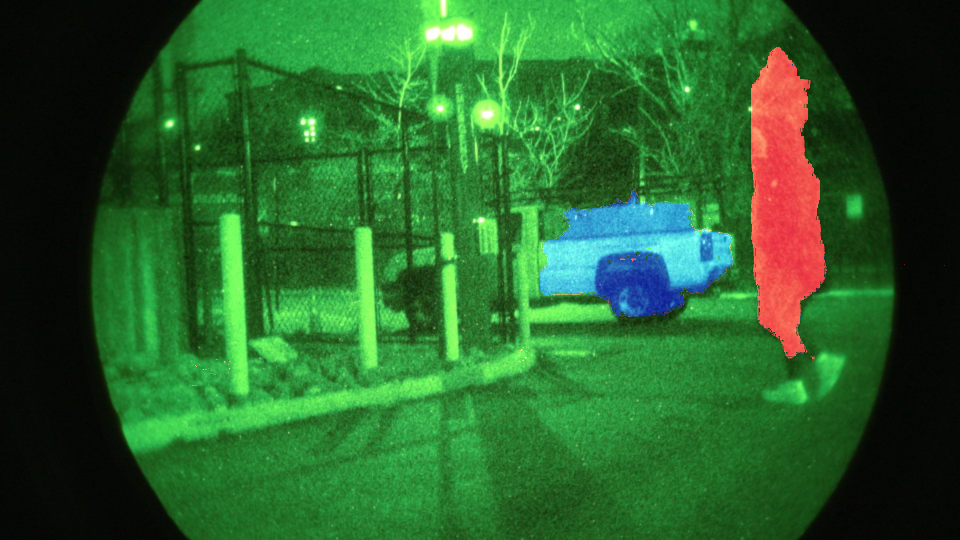}
    \end{subfigure}

    \begin{subfigure}[b]{0.3\textwidth}
        \includegraphics[width=\linewidth]{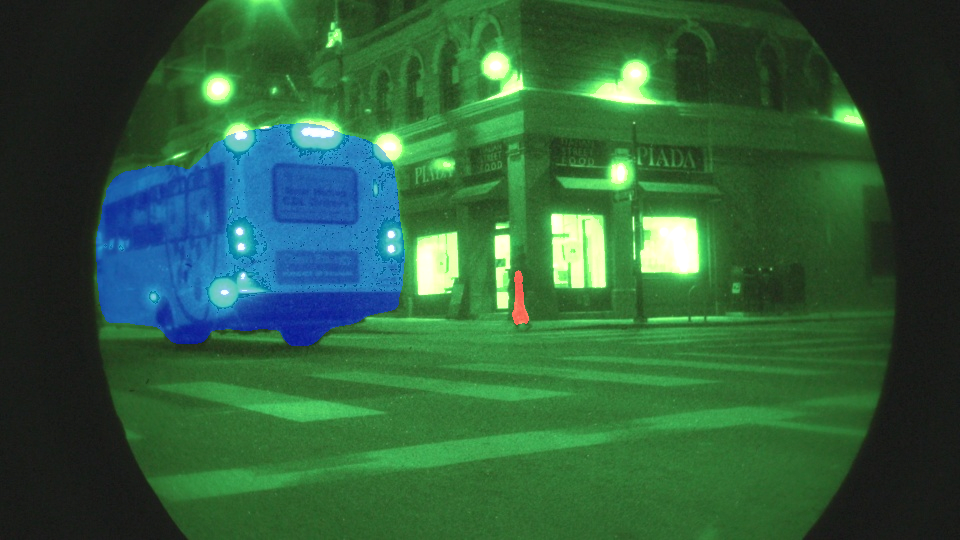}
    \end{subfigure}
    \hfill
    \begin{subfigure}[b]{0.3\textwidth}\includegraphics[width=\linewidth]{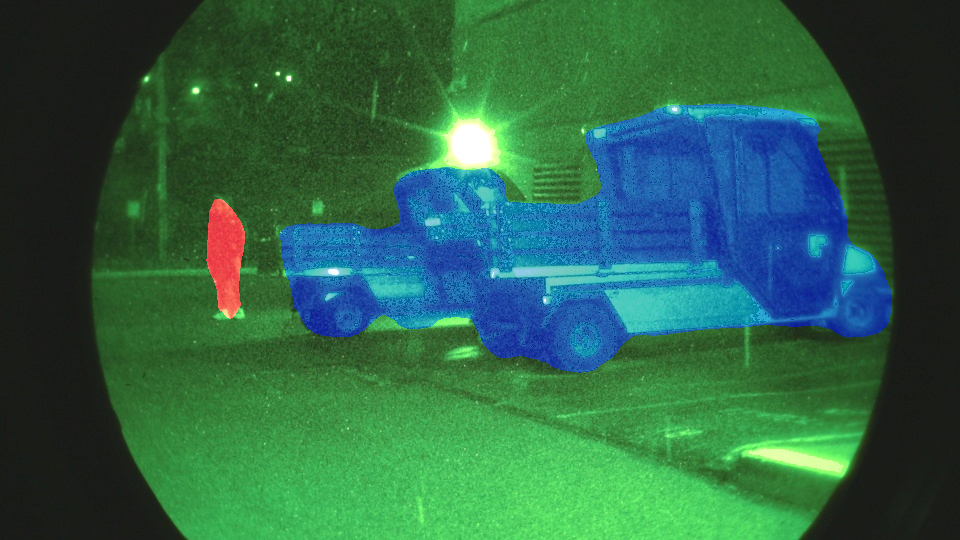}
    \end{subfigure} 
    \hfill
    \begin{subfigure}[b]{0.3\textwidth}
        \includegraphics[width=\linewidth]{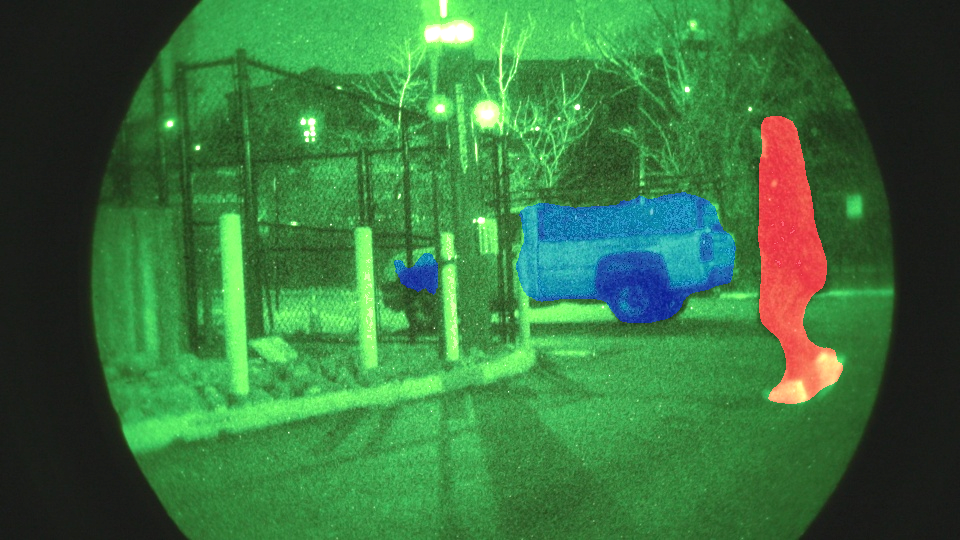}
    \end{subfigure}

    \begin{subfigure}[b]{0.3\textwidth}
        \includegraphics[width=\linewidth]{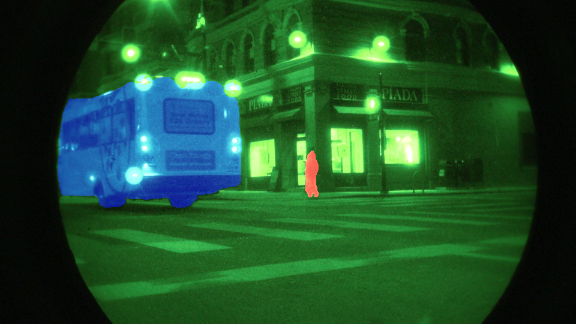}
    \end{subfigure}
    \hfill
    \begin{subfigure}[b]{0.3\textwidth}\includegraphics[width=\linewidth]{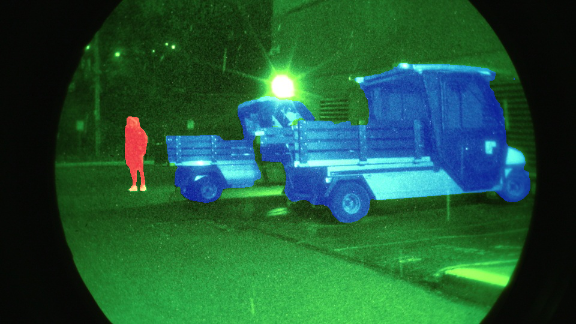}
    \end{subfigure} 
    \hfill
    \begin{subfigure}[b]{0.3\textwidth}
        \includegraphics[width=\linewidth]{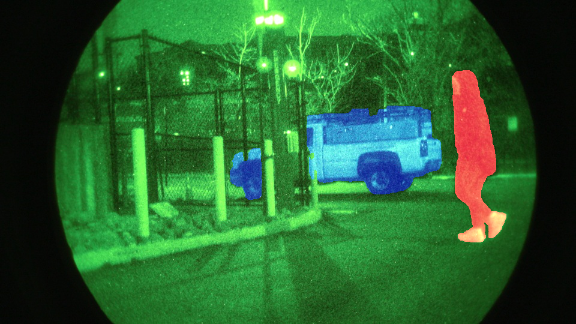}
    \end{subfigure}
    
    \caption{Stage-wise qualitative results, with first row corresponding to query image, second to baseline \texttt{UDA} results, third to \texttt{UDA} with \texttt{SP}, fourth with \texttt{UDA}, \texttt{SP}, and \texttt{SA}, i.e. \texttt{AUDA}, and the last corresponding to ground truth labeling. It is clear that each step of \texttt{AUDA}, critically \texttt{SP}, improves our target domain outputs.}
    \label{fig:qualitative-results3}
\end{figure}

Figures \ref{fig:qualitative-results1}, \ref{fig:qualitative-results2} and \ref{fig:qualitative-results3} show qualitative results after different stages of \texttt{AUDA} on \emph{CI}, along with results after just \texttt{UDA} without \texttt{SP} to understand and show the efficacy of \texttt{SP}. In all figures, all results corresponding to a particular image are added to the same column as the image. The first row corresponds to the query image, the second to the output we obtain after \texttt{UDA} without \texttt{SP}, the third to the output we obtain after \texttt{UDA} with \texttt{SP}, the fourth after \texttt{SP}, \texttt{UDA}, and \texttt{SA}, i.e. complete \texttt{AUDA}, and the last corresponds to the ground truth labeling. While we show predictions that go into region of the image blocked by the frame of the intensifier module, these regions are marked to belong to the `invalid' class, and so don't affect any quantitative metrics. 

From these figures, we can see that \texttt{UDA} with \texttt{SP} greatly helps with the reduction of both false positives and false negatives. This happens particularly in images, or regions of images with high noise, such as the images captured in a dark park, which has a lot of low-light noise, or in regions with bright lights on streets. Both of these have high-frequency components. With \texttt{SP} with our blur-based scheme, we make our source model more robust to variations in such features, which leads to enhanced domain adaptation as we had hypothesized. 

We can also see that \texttt{SA} generally refines, and further improves the outputs we obtain after \texttt{UDA} and \texttt{SP}, and gives us the results closest to what we observe in the last row, i.e. ground truth.   

Our qualitative results thus support our hypotheses of making source models more adaptable to enhance domain adaptation with \texttt{SP}, and of using limited \texttt{SA} to improve the models we can train in limited target label settings. 

\FloatBarrier

\section{CityIntensified Dataset: Extended Analysis}

\begin{figure}[h]
    \centering
    \begin{subfigure}[b]{0.3\textwidth}
        \includegraphics[width=\linewidth]{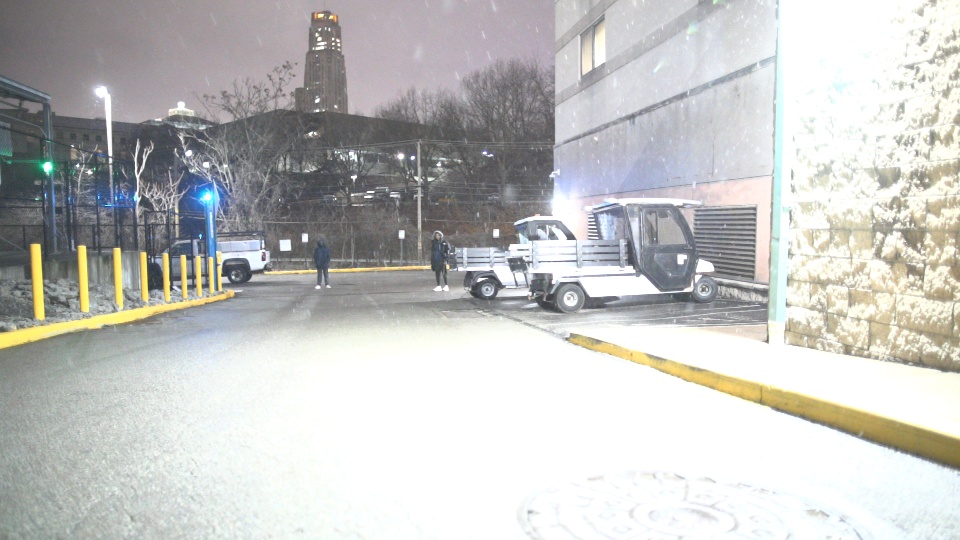}
    \end{subfigure}
    \hfill
    \begin{subfigure}[b]{0.3\textwidth}
        \includegraphics[width=\linewidth]
        {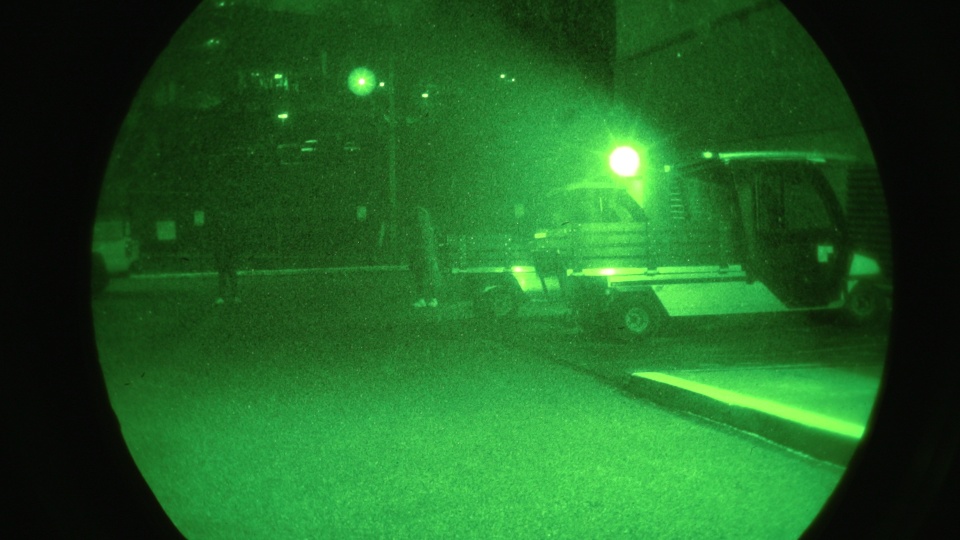}
    \end{subfigure}
    \hfill
    \begin{subfigure}[b]{0.3\textwidth}
        \includegraphics[width=\linewidth]{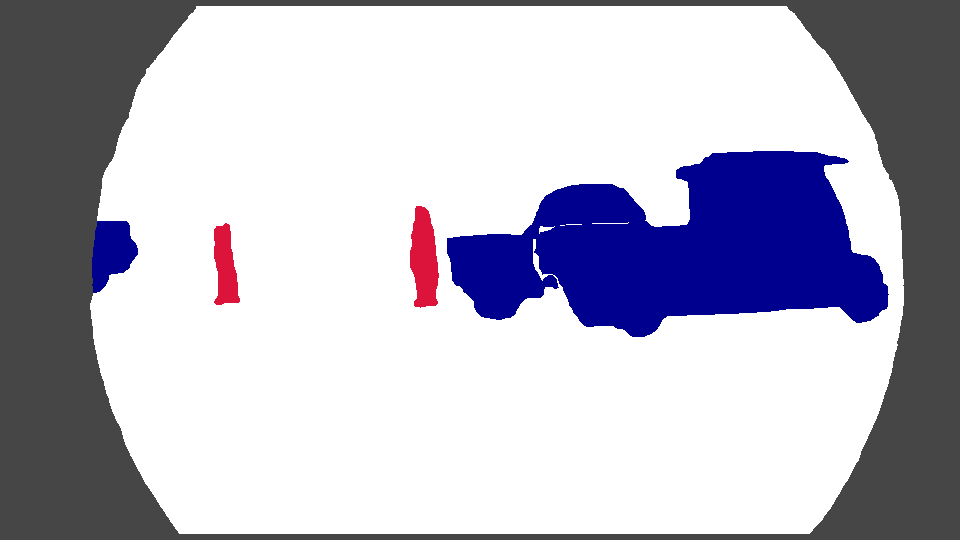}
    \end{subfigure}

    \begin{subfigure}[b]{0.3\textwidth}
        \includegraphics[width=\linewidth]{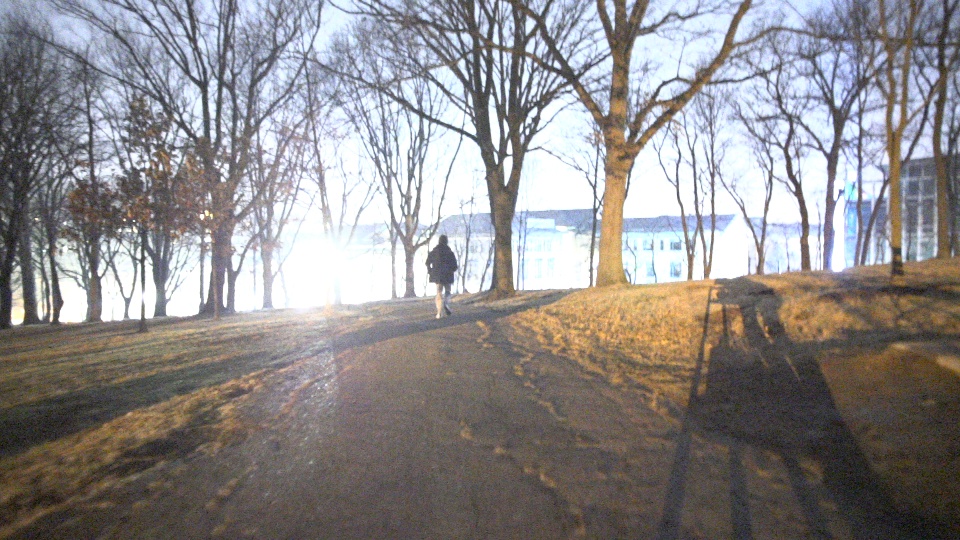}
    \end{subfigure}
    \hfill
    \begin{subfigure}[b]{0.3\textwidth}
        \includegraphics[width=\linewidth]{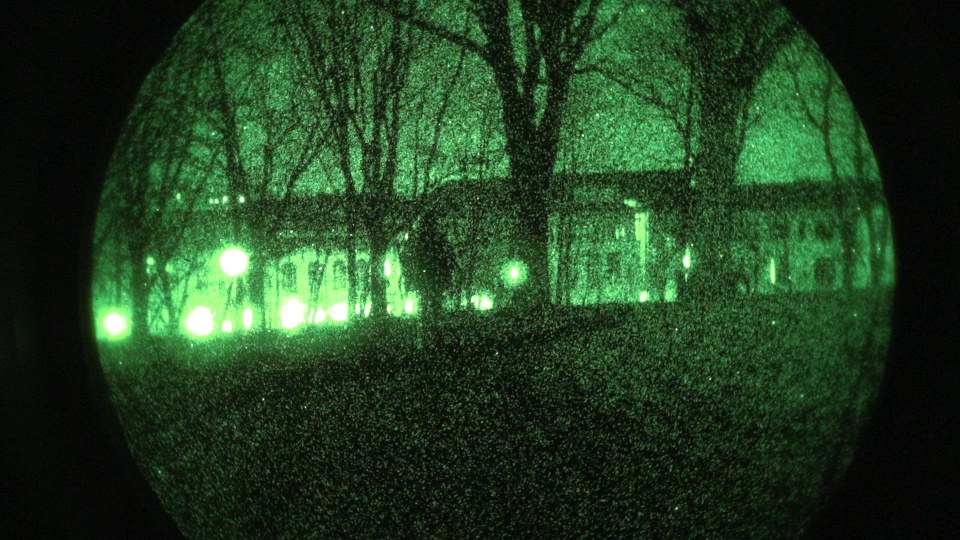}
    \end{subfigure}
    \hfill
    \begin{subfigure}[b]{0.3\textwidth}
\includegraphics[width=\linewidth]{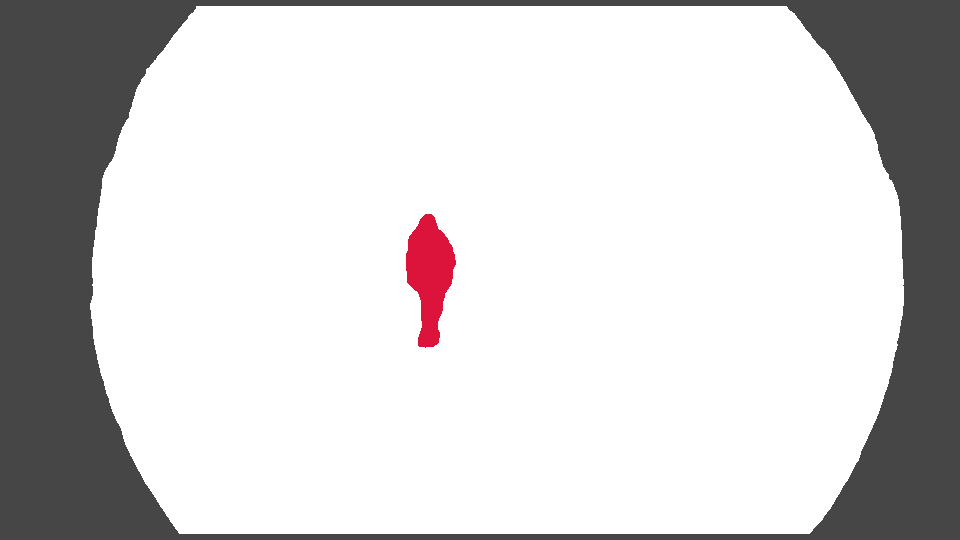}
    \end{subfigure}

    \begin{subfigure}[b]{0.3\textwidth}
        \includegraphics[width=\linewidth]{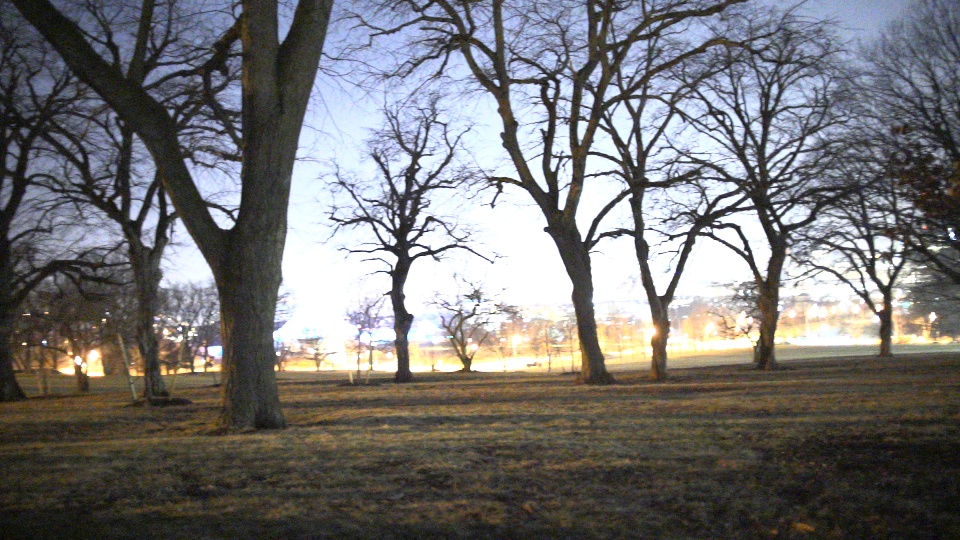}
    \end{subfigure}
    \hfill
    \begin{subfigure}[b]{0.3\textwidth}
        \includegraphics[width=\linewidth]{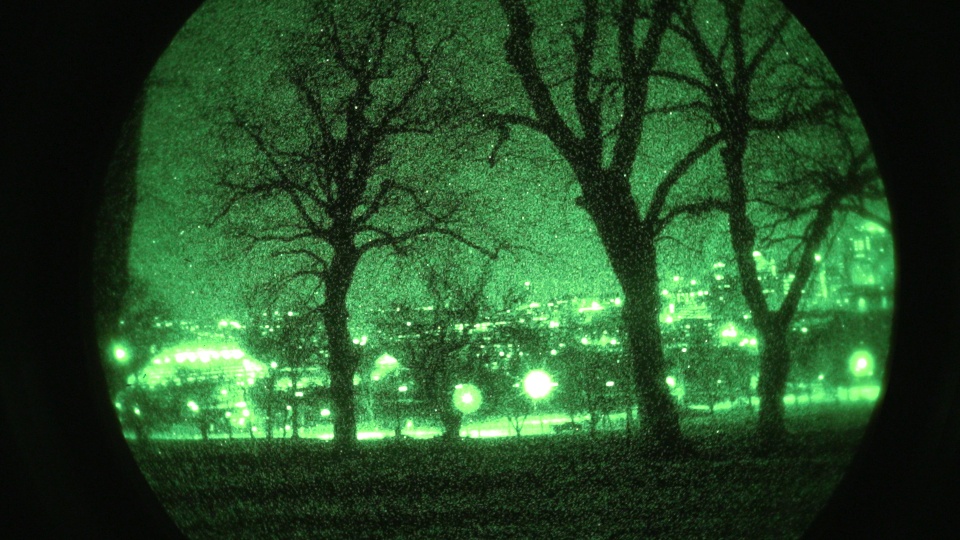}
    \end{subfigure}
    \hfill
    \begin{subfigure}[b]{0.3\textwidth}
\includegraphics[width=\linewidth]{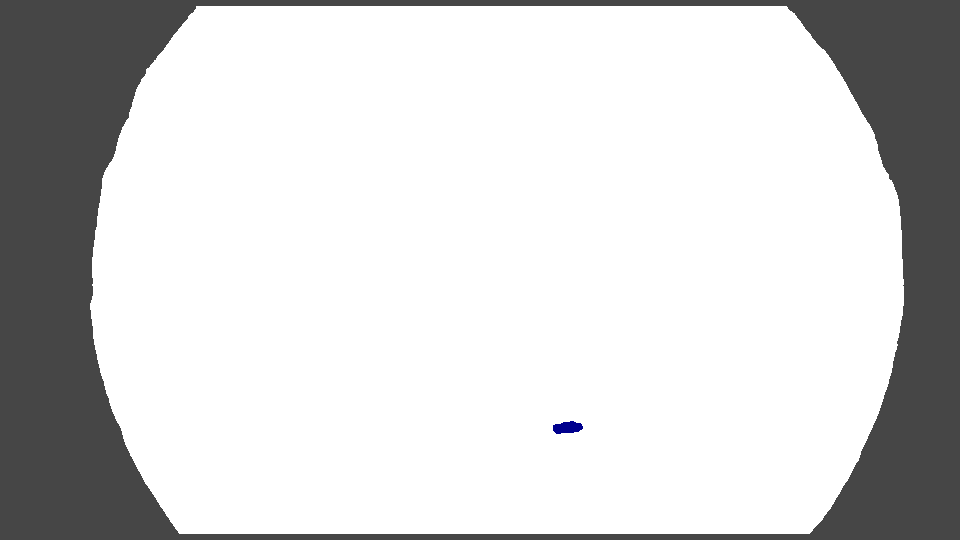}
    \end{subfigure}


    \begin{subfigure}[b]{0.3\textwidth}
        \includegraphics[width=\linewidth]{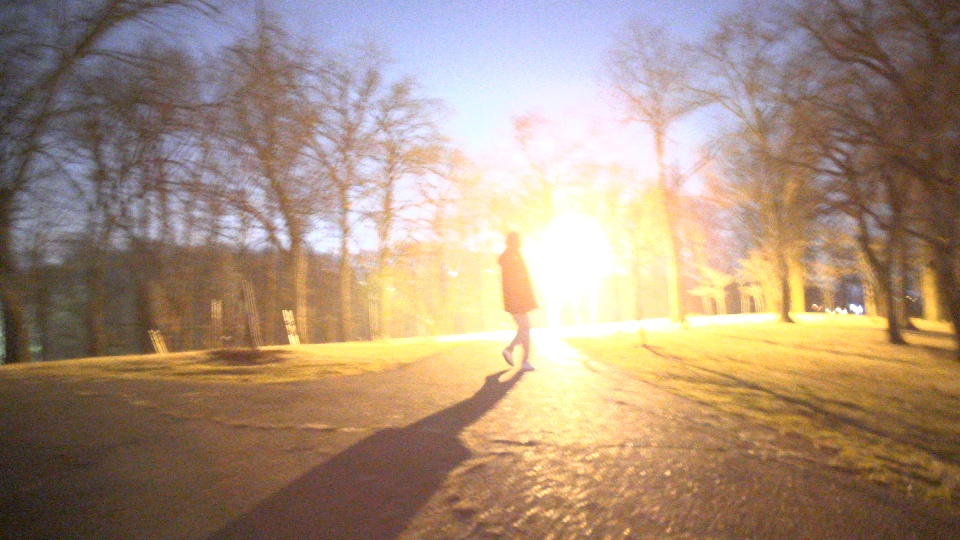}
    \end{subfigure}
    \hfill
    \begin{subfigure}[b]{0.3\textwidth}
        \includegraphics[width=\linewidth]{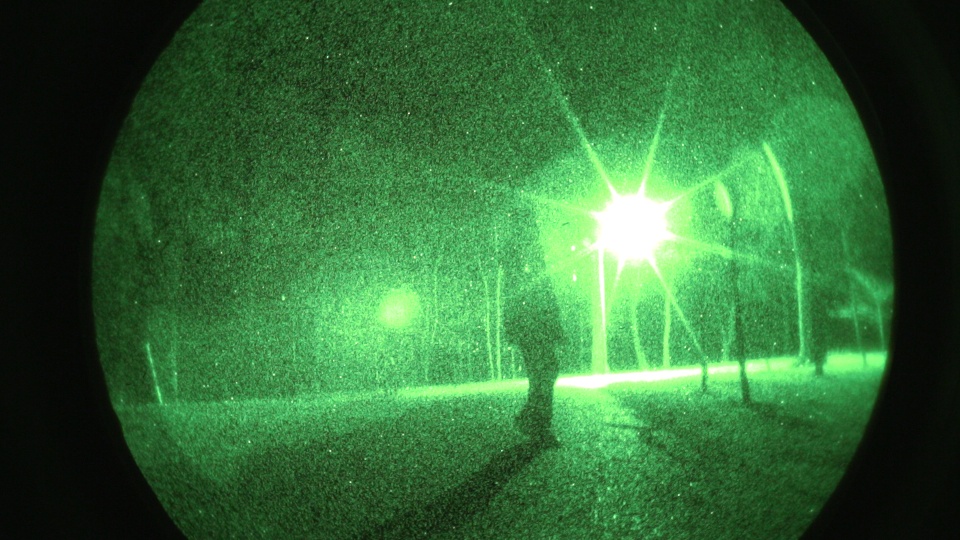}
    \end{subfigure}
    \hfill
    \begin{subfigure}[b]{0.3\textwidth}
\includegraphics[width=\linewidth]{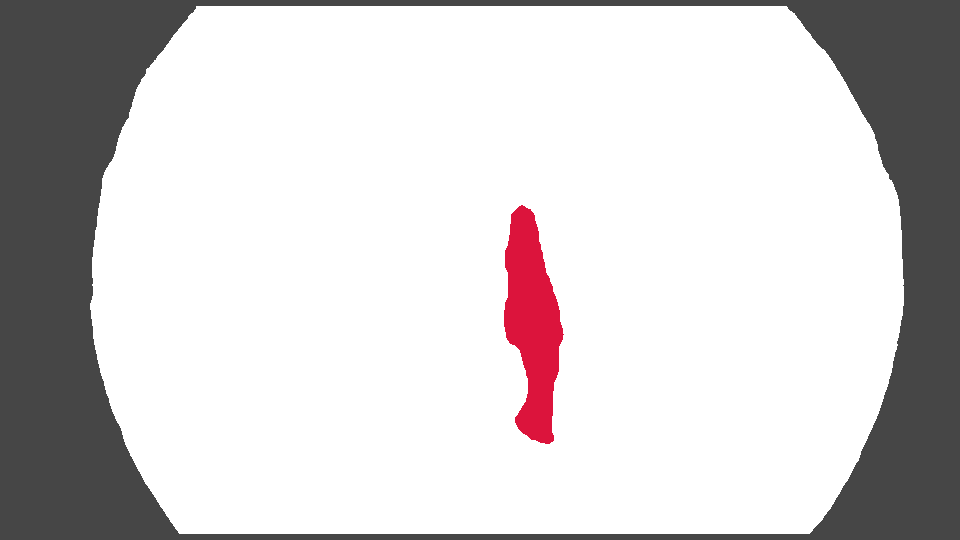}
    \end{subfigure}

    \begin{subfigure}[b]{0.3\textwidth}
        \includegraphics[width=\linewidth]{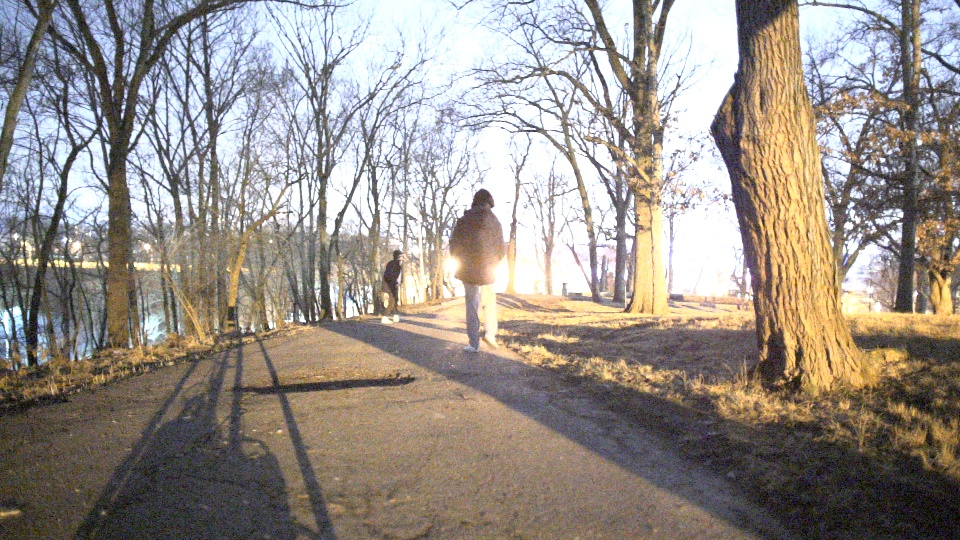}
    \end{subfigure}
    \hfill
    \begin{subfigure}[b]{0.3\textwidth}
        \includegraphics[width=\linewidth]{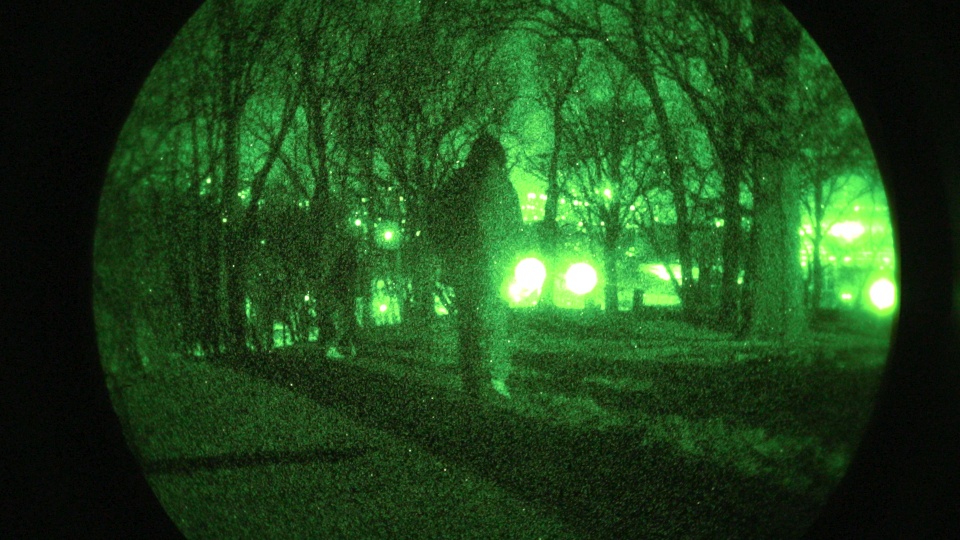}
    \end{subfigure}
    \hfill
    \begin{subfigure}[b]{0.3\textwidth}
\includegraphics[width=\linewidth]{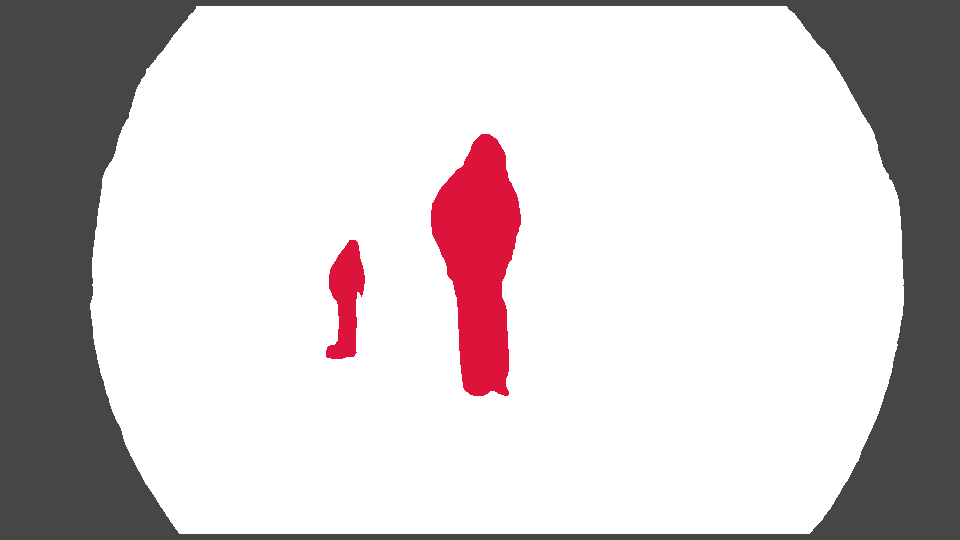}
    \end{subfigure}

    \begin{subfigure}[b]{0.3\textwidth}
        \includegraphics[width=\linewidth]{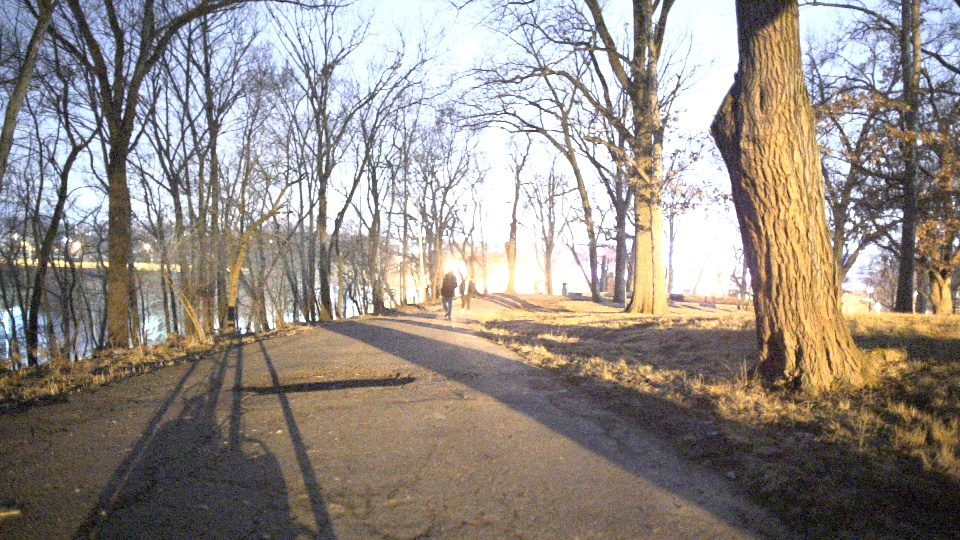}
    \end{subfigure}
    \hfill
    \begin{subfigure}[b]{0.3\textwidth}
        \includegraphics[width=\linewidth]{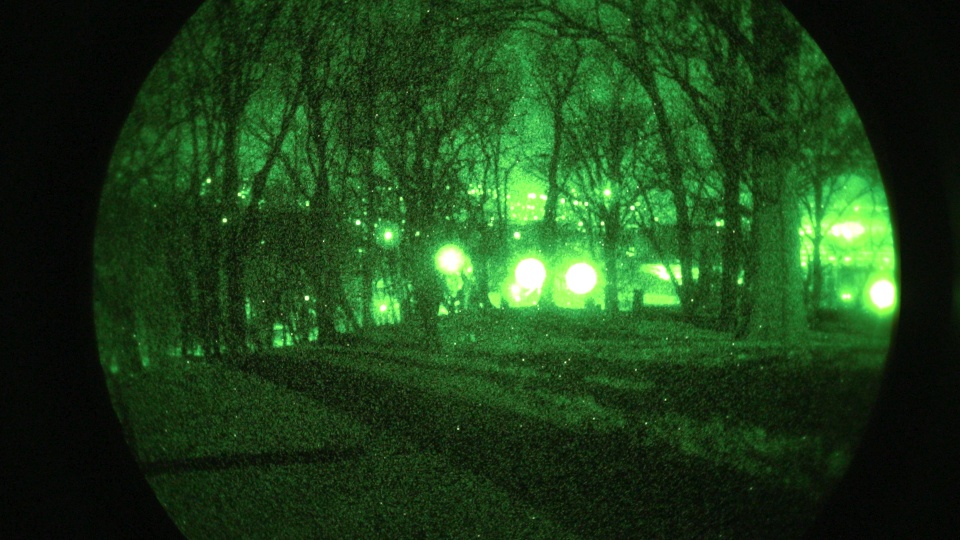}
    \end{subfigure}
    \hfill
    \begin{subfigure}[b]{0.3\textwidth}
\includegraphics[width=\linewidth]{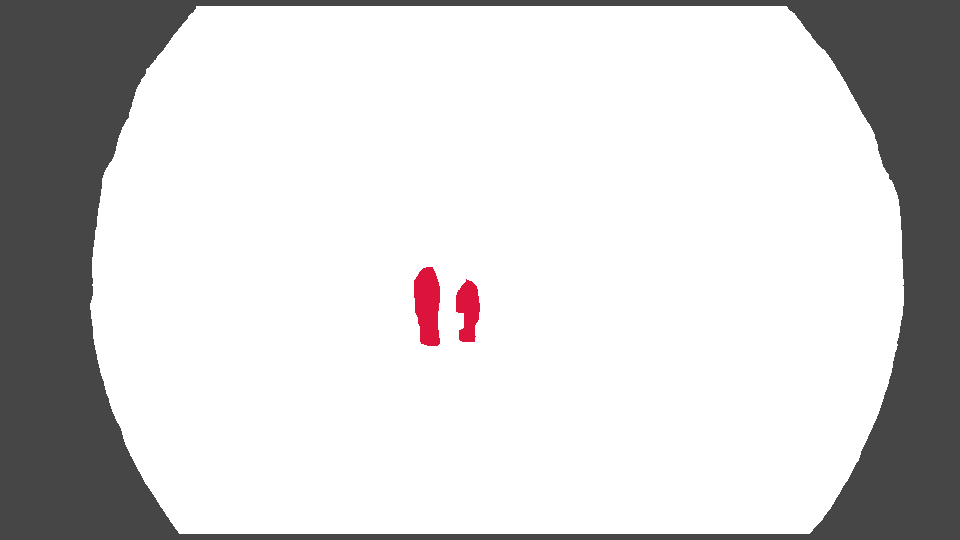}
    \end{subfigure}

    \begin{subfigure}[b]{0.3\textwidth}
        \includegraphics[width=\linewidth]{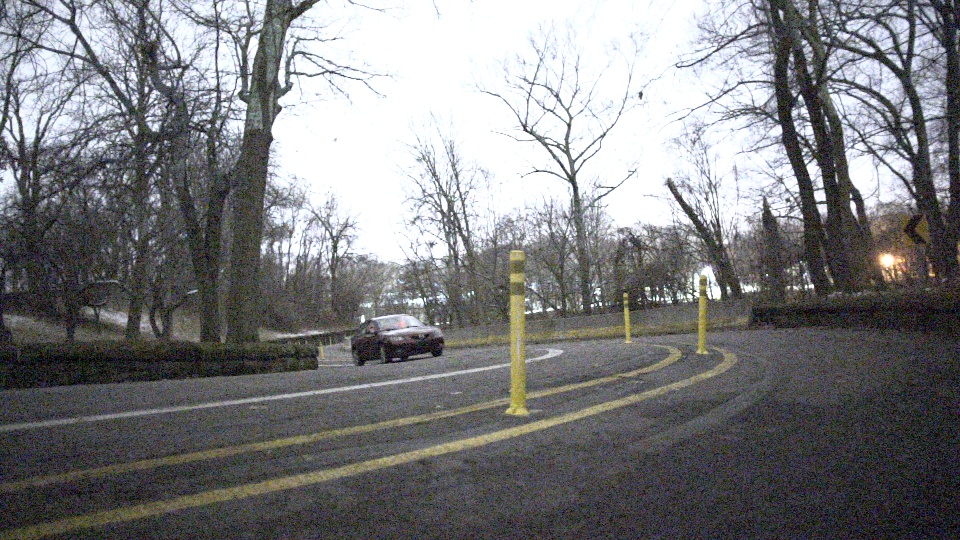}
    \end{subfigure}
    \hfill
    \begin{subfigure}[b]{0.3\textwidth}
        \includegraphics[width=\linewidth]{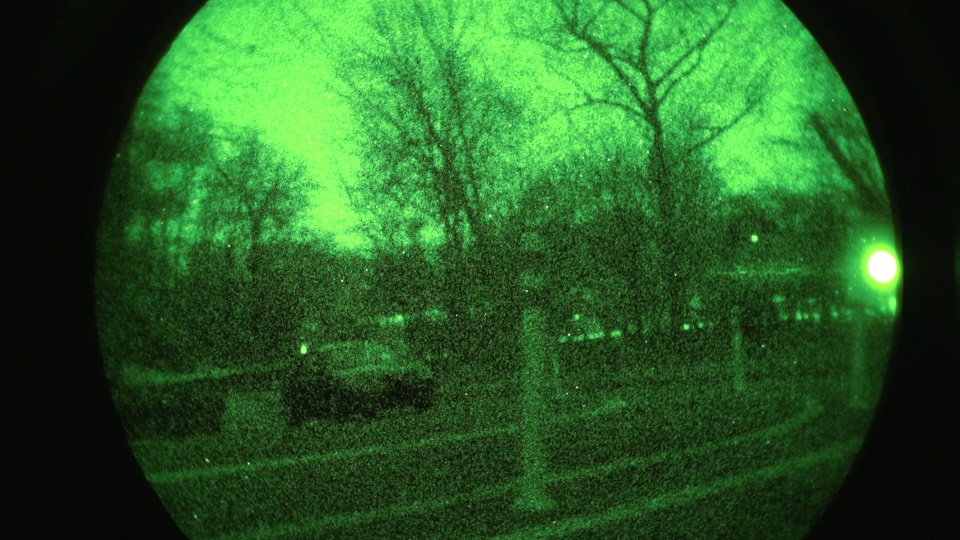}
    \end{subfigure}
    \hfill
    \begin{subfigure}[b]{0.3\textwidth}
\includegraphics[width=\linewidth]{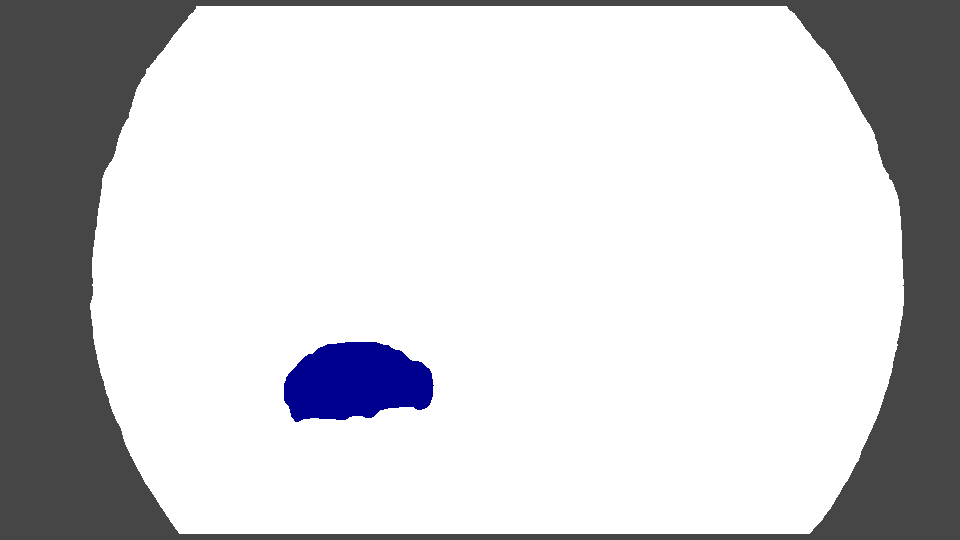}
    \end{subfigure}
    \caption{Additional representative examples from CityIntensified, with corresponding segmentation labels.}
    \label{fig:example-CI-supplemental}
\end{figure}
\begin{figure}[h]
    \centering
    \begin{subfigure}[b]{0.3\textwidth}
        \includegraphics[width=\linewidth]{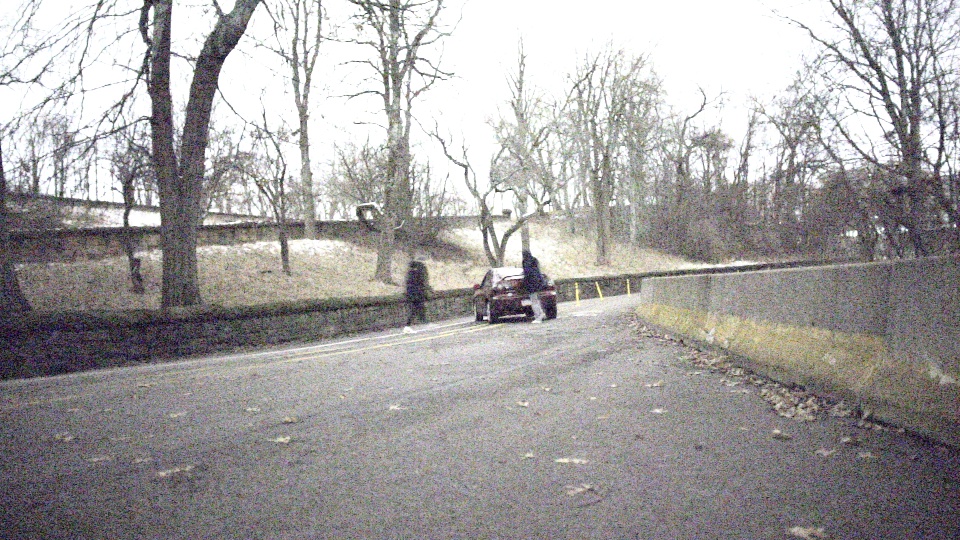}
    \end{subfigure}
    \hfill
    \begin{subfigure}[b]{0.3\textwidth}\includegraphics[width=\linewidth]{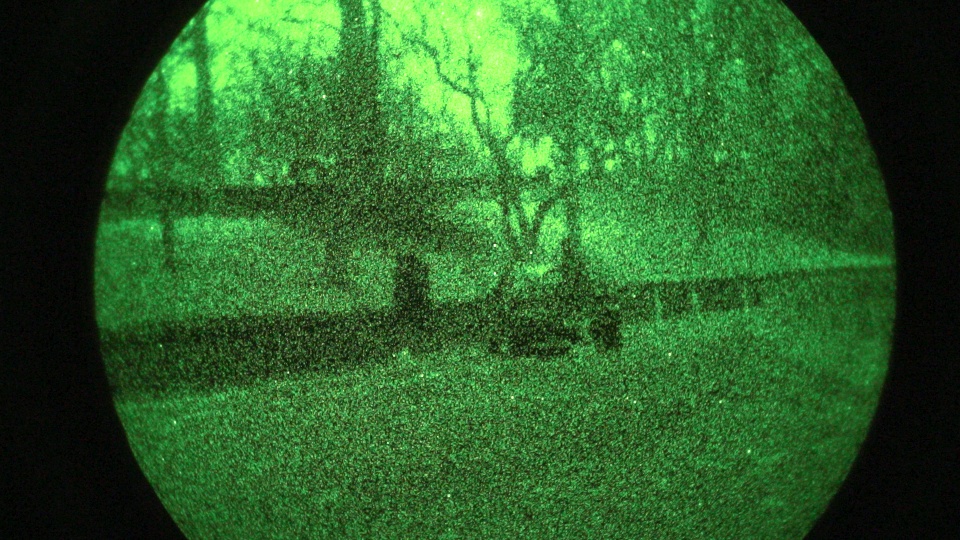}
    \end{subfigure}
    \hfill
    \begin{subfigure}[b]{0.3\textwidth}
        \includegraphics[width=\linewidth]{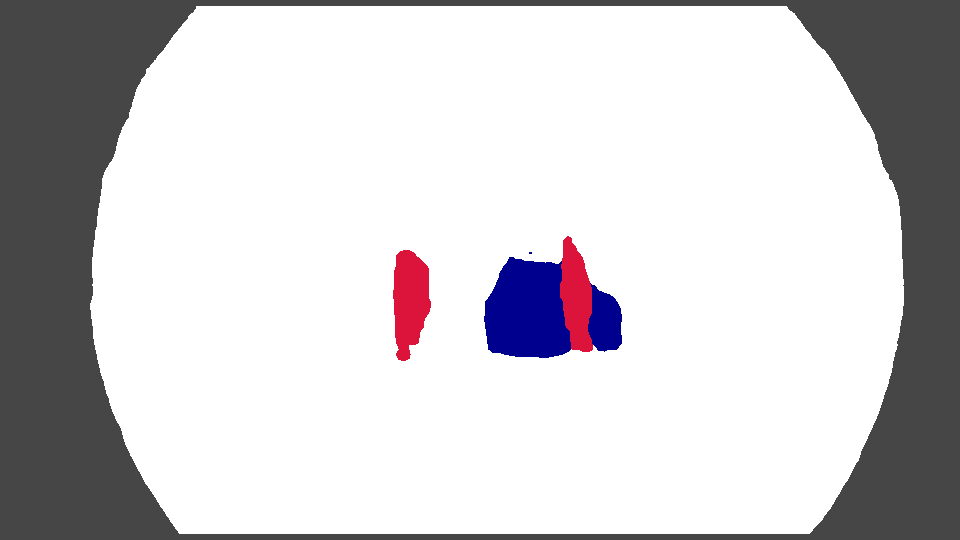}
    \end{subfigure}

    \begin{subfigure}[b]{0.3\textwidth}
        \includegraphics[width=\linewidth]{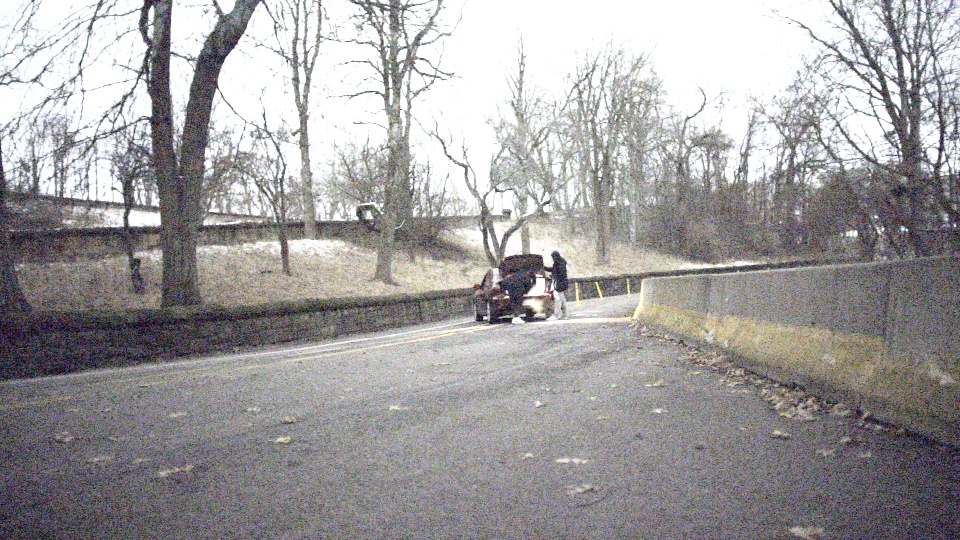}
    \end{subfigure}
    \hfill
    \begin{subfigure}[b]{0.3\textwidth}
        \includegraphics[width=\linewidth]{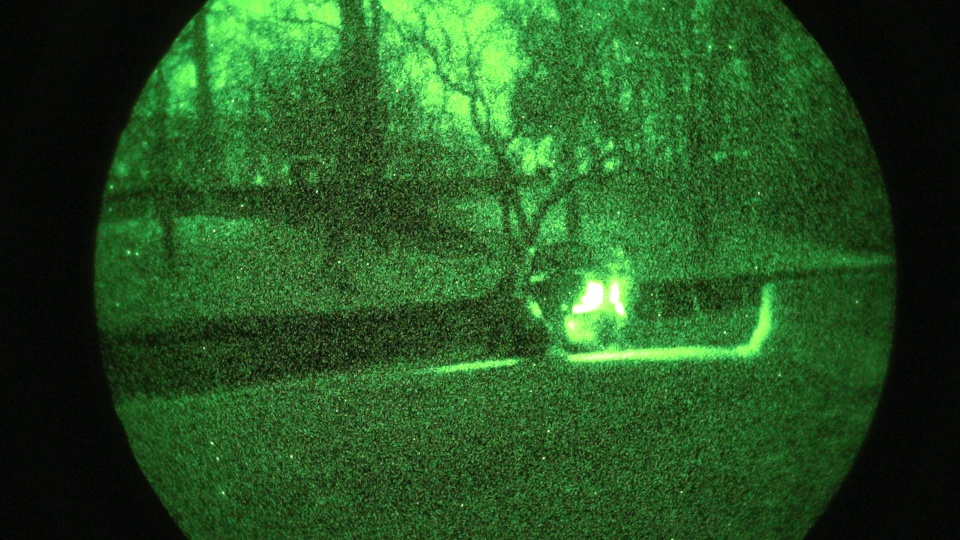}
    \end{subfigure}
    \hfill
    \begin{subfigure}[b]{0.3\textwidth}
\includegraphics[width=\linewidth]{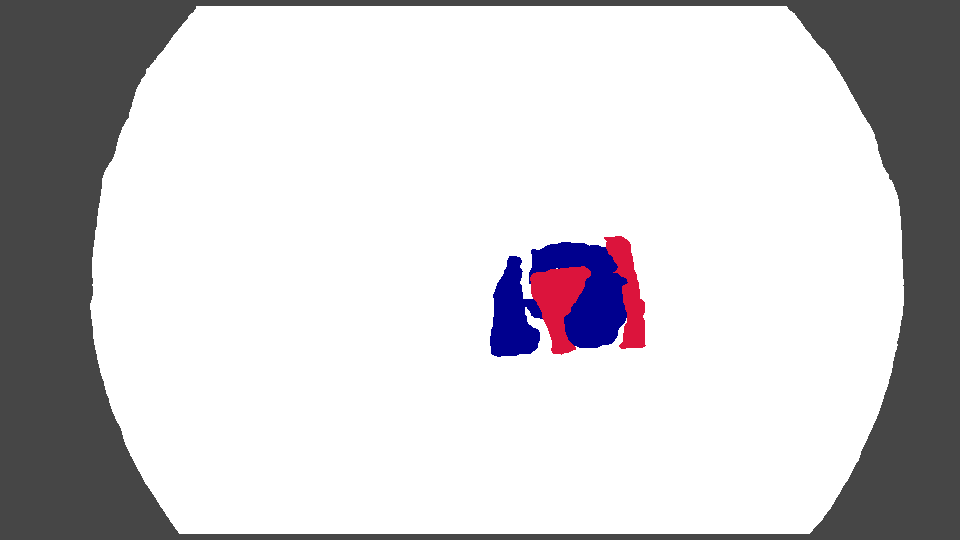}
    \end{subfigure}

    \begin{subfigure}[b]{0.3\textwidth}
        \includegraphics[width=\linewidth]{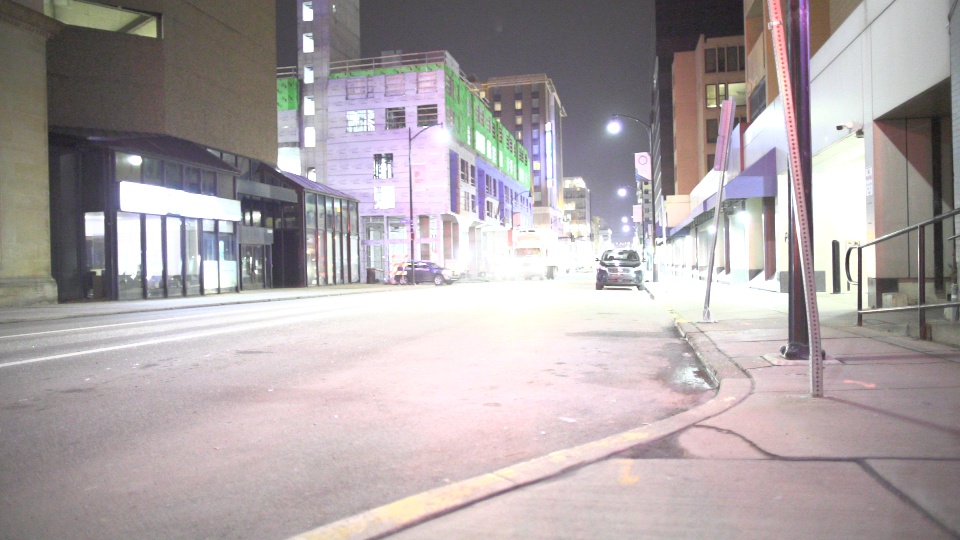}
    \end{subfigure}
    \hfill
    \begin{subfigure}[b]{0.3\textwidth}
        \includegraphics[width=\linewidth]{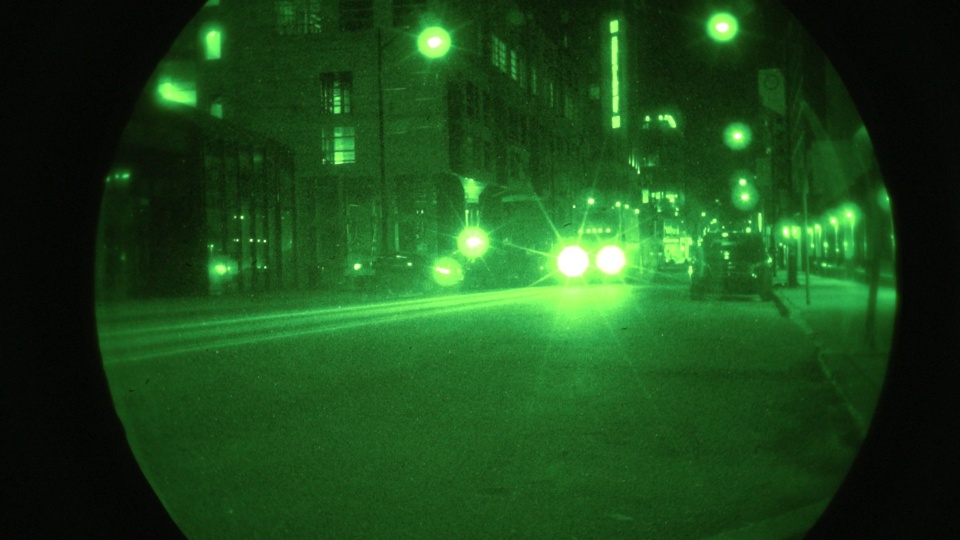}
    \end{subfigure}
    \hfill
    \begin{subfigure}[b]{0.3\textwidth}
\includegraphics[width=\linewidth]{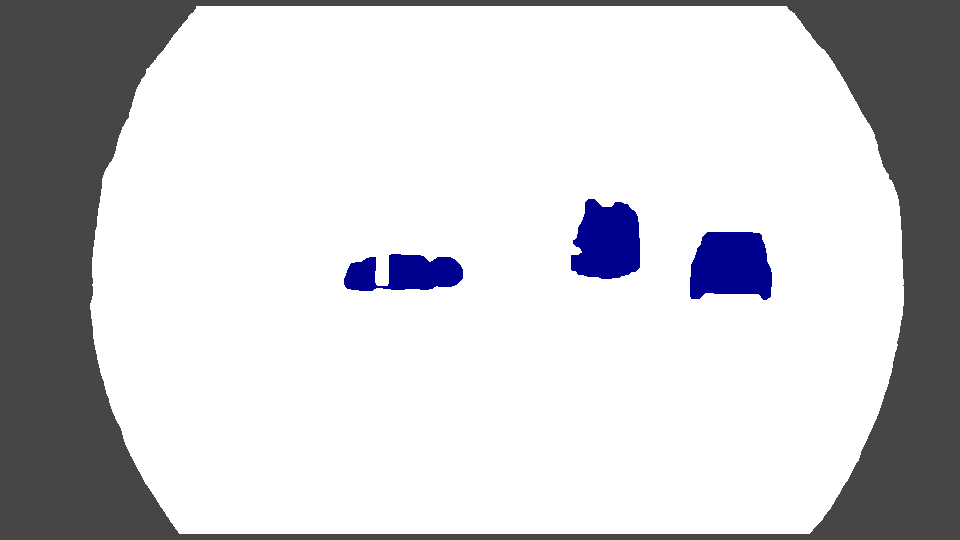}
    \end{subfigure}

    \begin{subfigure}[b]{0.3\textwidth}
        \includegraphics[width=\linewidth]{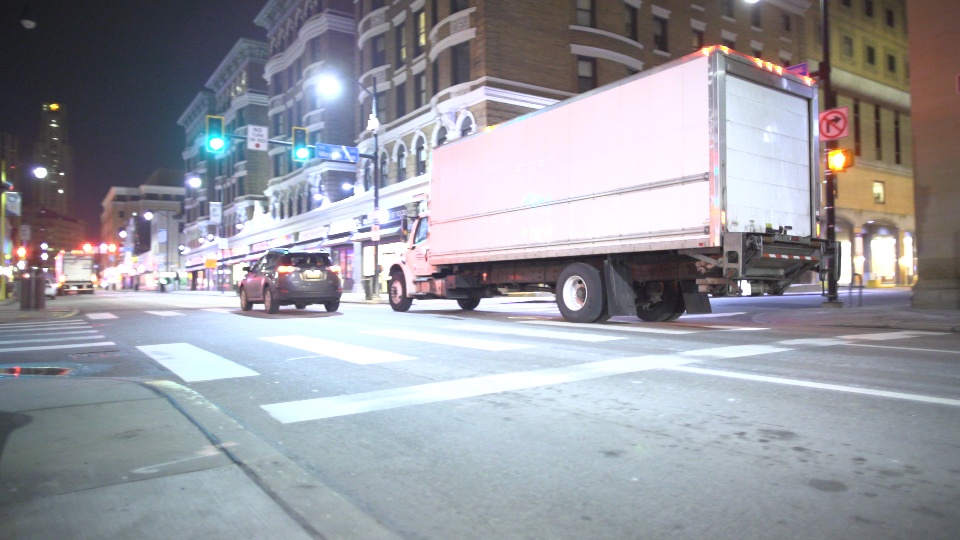}
    \end{subfigure}
    \hfill
    \begin{subfigure}[b]{0.3\textwidth}
        \includegraphics[width=\linewidth]{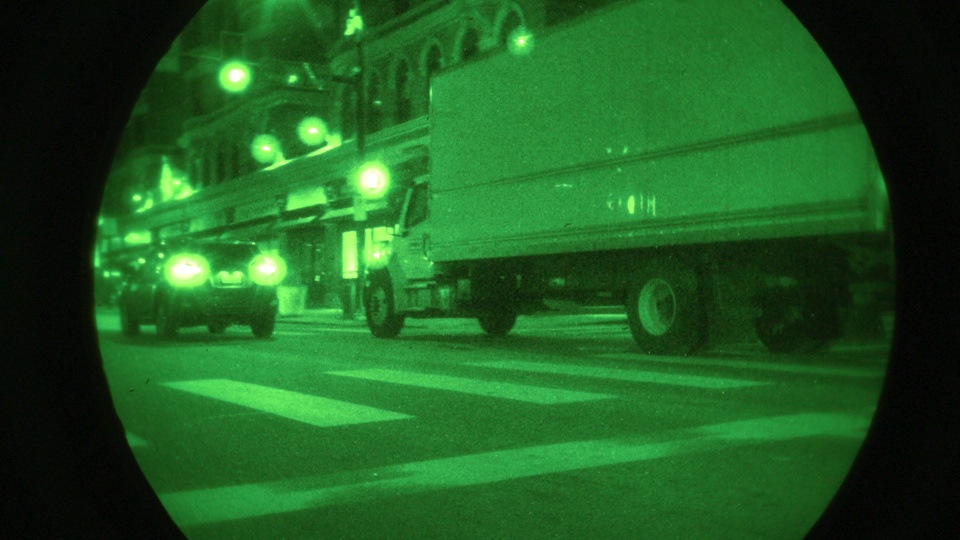}
    \end{subfigure}
    \hfill
    \begin{subfigure}[b]{0.3\textwidth}
\includegraphics[width=\linewidth]{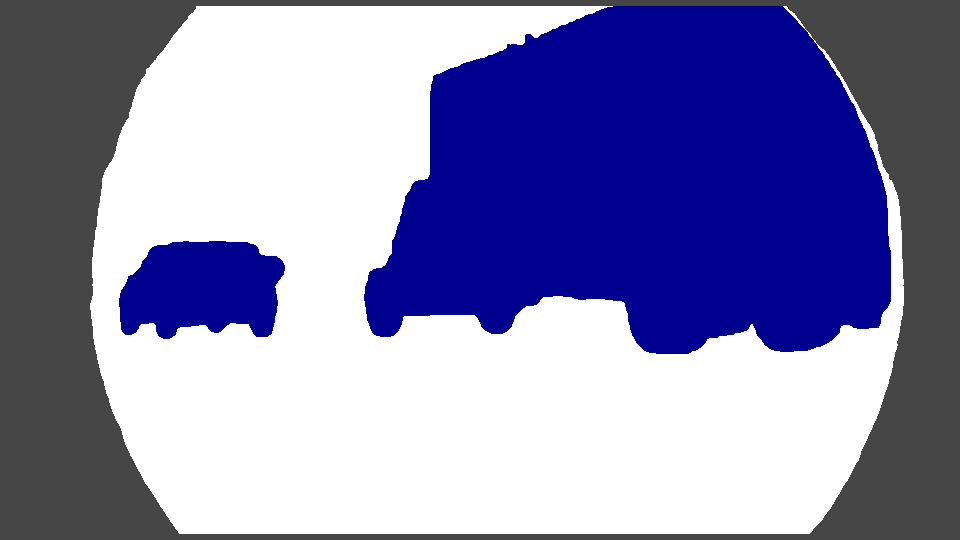}
    \end{subfigure}

    \begin{subfigure}[b]{0.3\textwidth}
        \includegraphics[width=\linewidth]{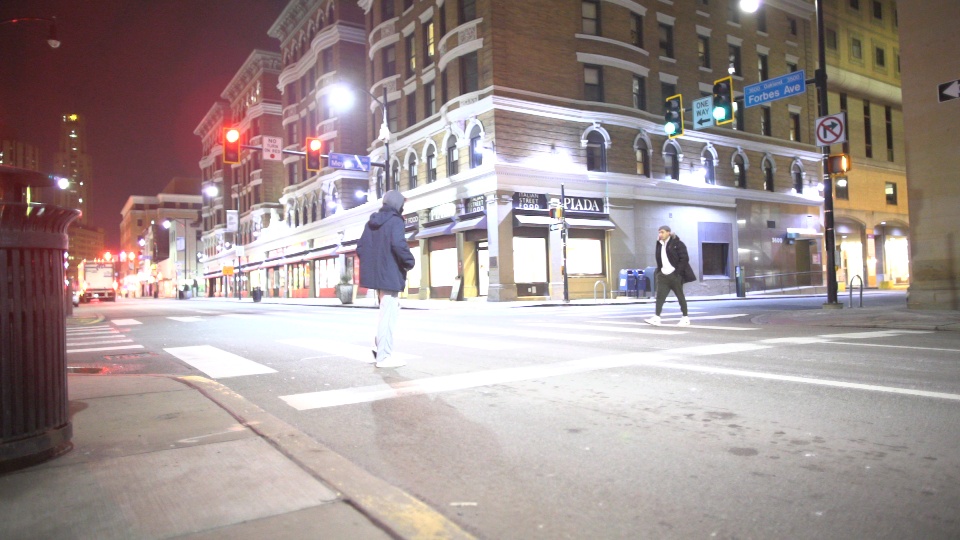}
    \end{subfigure}
    \hfill
    \begin{subfigure}[b]{0.3\textwidth}
        \includegraphics[width=\linewidth]{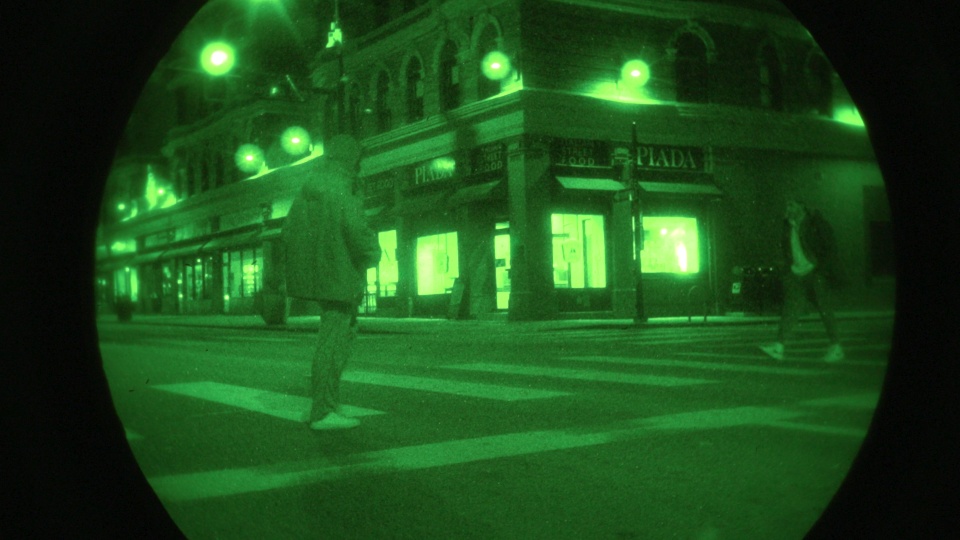}
    \end{subfigure}
    \hfill
    \begin{subfigure}[b]{0.3\textwidth}
\includegraphics[width=\linewidth]{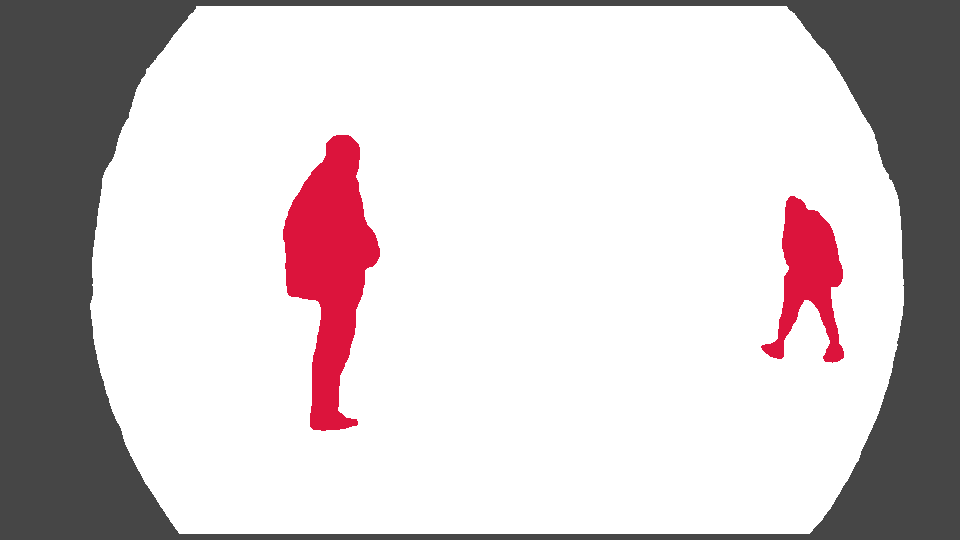}
    \end{subfigure}
    \caption{Additional representative examples from CityIntensified, with corresponding segmentation labels.}
    \label{fig:example-CI-supplemental2}
\end{figure}
\begin{figure}
    \centering
    \includegraphics[scale=0.15]{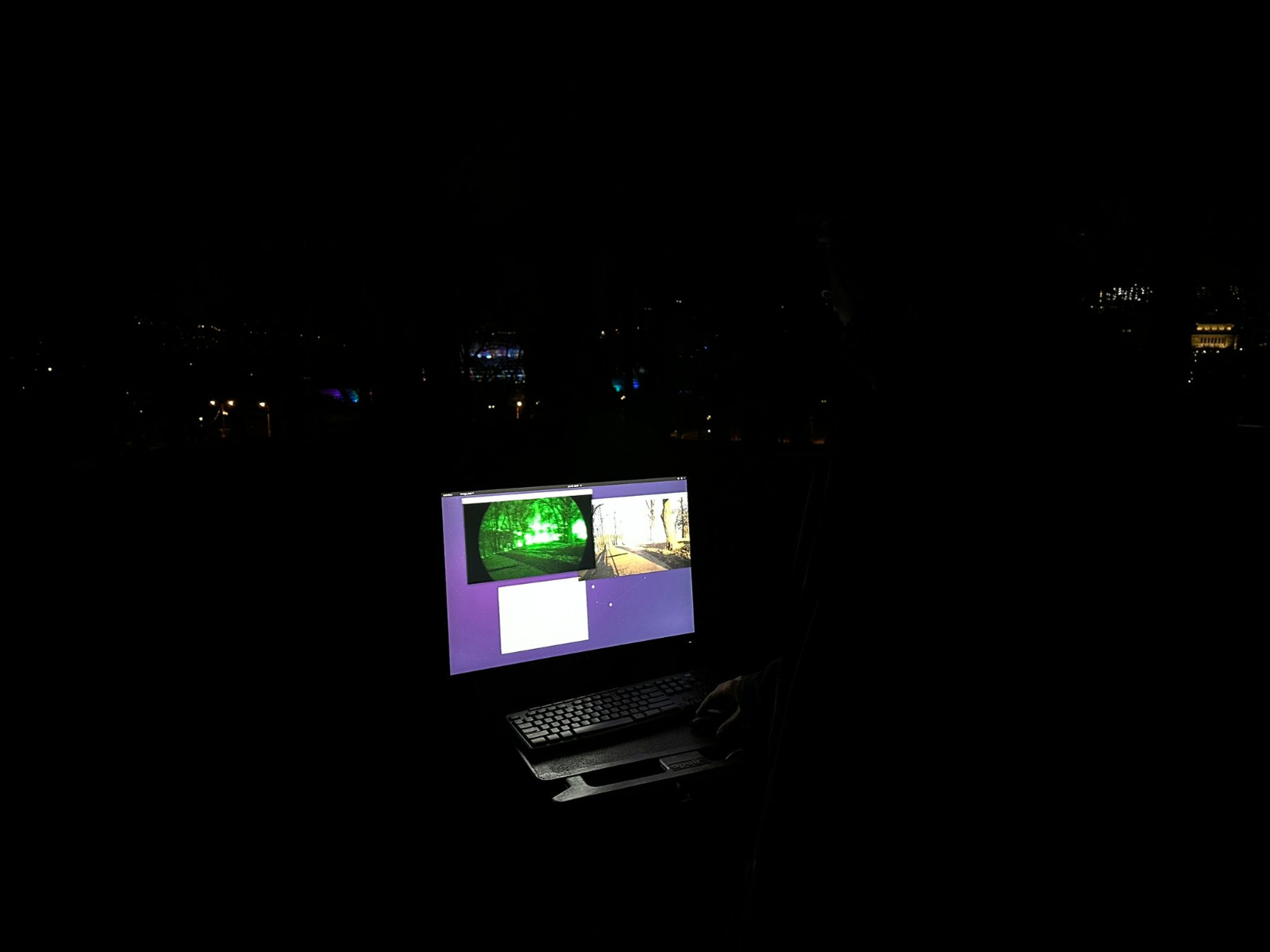}
\caption{Representation of how these scenes appear with a regular camera.}
    \label{fig:capture-setup}
\end{figure}

We collected the data that comprises CityIntensified on two separate nights in a city in the United States, with all images of size $960\times540$. It has a total of 11 sequences, 5 of which are taken within a park to capture scenes with minimal city light from sources such as buildings and street lights, 5 on-road, and 1 in a parking lot. Figure \ref{fig:capture-setup} captures our recording set-up with a regular phone camera and gives an idea of how dark these scenes appear before intensification. More examples from CityIntensified can be found in Figures \ref{fig:example-CI-supplemental}, \ref{fig:example-CI-supplemental2}, where the first column corresponds to images from the high-sensitivity camera, the second from the intensifier camera, and the third corresponding to their ground truth segmentation labels. In these labels, \blue{blue} is used to represent the `vehicle' class, \red{red} to represent the `people' class, \emph{white} to represent the background class, and \textcolor{gray}{gray} to represent the ignore label, which corresponds to a fixed area in the image blocked by the intensifier module, and predictions here can be accounted for trivially in a robotic set-up. 

\begin{figure}[h!]
    \centering\includegraphics[scale=0.3]{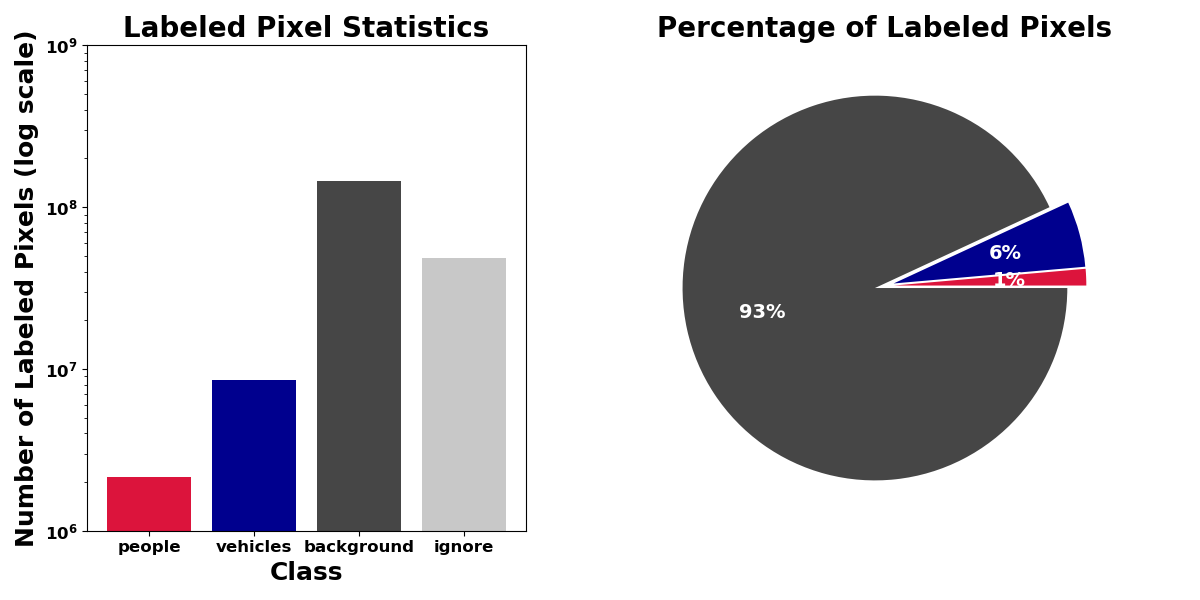}
    \caption{Number of annotated pixels in each labeled class in CityIntensified. }\label{fig:CIPixelLabelingStaistics}
\end{figure}

We manually refine the coarse annotations generated by Segment Anything \cite{kirillov2023segany} to provide semantic and instance-level labels for a subsampled set of 393 images from the intensifier camera. Figure \ref{fig:CIPixelLabelingStaistics} illustrates the number of pixels annotated per class, and the percentage of valid pixels belonging to each class. 

\begin{figure}[h!]
    \centering
    \includegraphics[scale=0.2]{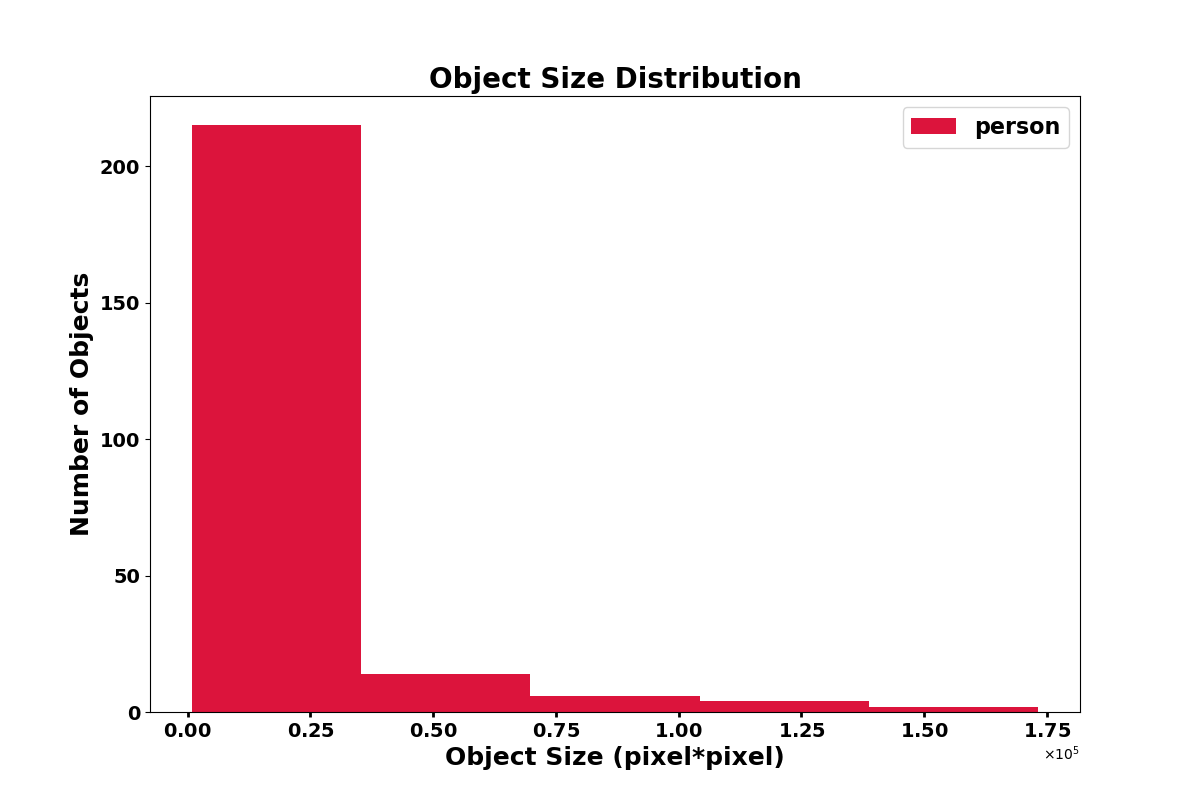}
\includegraphics[scale=0.2]{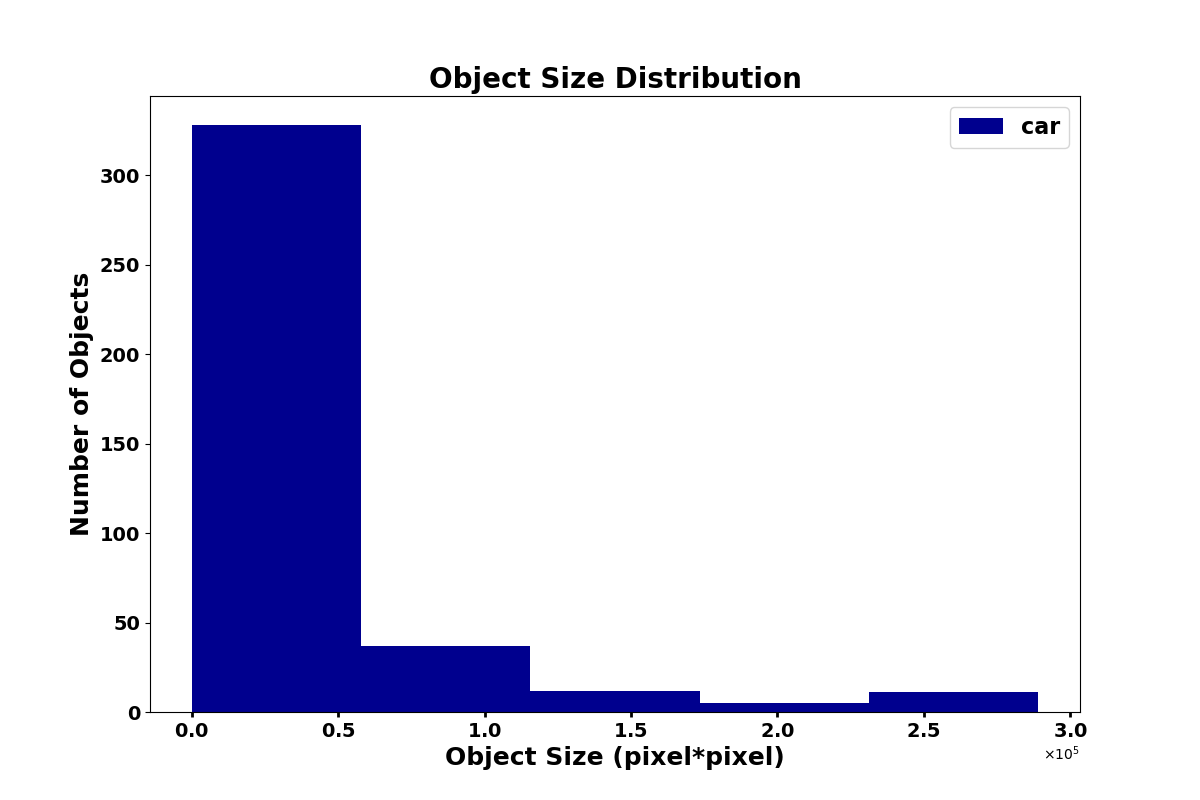}
    \caption{Number and distribution over sizes of annotated labels of objects with detection (bbox) and instance segmentation labels for people and vehicles. There are a total of 241 instances of the 'people' and 393 instances of the 'vehicle' class in the 393 labeled images of CityIntensified.}
    \label{fig:object-detection-label-statistics}
\end{figure}

As a part of our instance-level labels, we provide bounding-box annotations for people, and vehicles. We show their distribution over different sizes in Figure \ref{fig:object-detection-label-statistics}. Across all images, we have 241 bounding boxes corresponding to `people' and 393 corresponding to `vehicle'. 

\FloatBarrier
\section{Additional Details on Methods, Experimental Set-up, and Compute Use}



\textbf{Selection of SP strategy for blur.}
While we experiment over a different range of kernel sizes while formulating our gaussian blur based \texttt{SP} scheme, our key decision choice was the nature of the random sampling over our range of possible kernels of sizes (5,5) to (19,19). We compared sampling uniformly over this range against sampling with a normal distribution centered around 12, with a standard deviation of 3.5. After \texttt{UDA} with \texttt{Refign-HRDA*}, the mIoU with a uniformly sampled Gaussian blur, $73.14\%$ was more than what we observed with the normally sampled counterpart, $68.2\%$. This suggested that sampling more evenly from a range of sizes among blur kernels provided a more useful signal for training, though the difference may be small. This however indicates that the use of a more complex and varied blurring scheme may further improve our \texttt{SP} scheme.

\textbf{Selection of Classes used in Evaluation for different datasets.}
Since the algorithms we use for \texttt{UDA} maps across domains while assuming a common set of labels in both domains, we evaluate our algorithm based on performance across only common classes, i.e. to obtain mIoU we take the average of IoU over these select classes. In the case of \emph{Cityscapes$\rightarrow$DarkZurich}~\cite{DarkZurich}, this includes all 19 classes used for Cityscapes evaluation. For \emph{CS$\rightarrow$MFNT}, we evaluate over cars, person, and bike classes of the \emph{MFNT} dataset. To account for similar classes in Cityscapes, we remap predictions for both motorcycle and bicycle to \emph{MFNT} class bike, and person and rider to \emph{MFNT} class person. Similarly, since \emph{CI} contains classes for `people' and `vehicles', the latter of which comprises cars, buses, and trucks, we remap predictions for person and rider to \emph{CI} class people, and car, truck, and bus to \emph{CI} class vehicle. For consistency, we compute mIoU across these select classes in all aforementioned experiments. 

\textbf{Compute Costs.} All our experiments have been run in a single GPU setting, with NVIDIA A100 40GB \cite{a100gpu} GPUs. While training times would defer based on the choice of specific architectures and methods in our framework, a single complete training of \texttt{AUDA} using SegFormer and Refign with \texttt{SP}, \texttt{UDA}, and \texttt{SA} takes approximately 2.5 days on a single GPU. Once a source model is trained with \texttt{SP}, it can however be used to train \texttt{UDA} to different target domains without any additional training costs. \texttt{SA} takes a fraction of the time the other two components take since we run it for just a few iterations.

\clearpage
\bibliographystyle{unsrt}
{\small
\bibliography{neurips_2023}
}


\end{document}